\documentclass{article}

\PassOptionsToPackage{numbers, sort&compress}{natbib}

\usepackage[main, final]{neurips_2026}

\usepackage{graphicx}
\usepackage{subcaption}
\usepackage{float}
\usepackage[justification=justified]{caption}	
\usepackage{lscape}                                         
\usepackage{wrapfig}

\usepackage[lined,ruled,linesnumbered]{algorithm2e}
\usepackage{animate} 

\usepackage{booktabs}                   
\usepackage{multirow}
\usepackage{ragged2e}
\usepackage{makecell}

\usepackage{paralist}
\usepackage{enumitem}

\usepackage{bm}                          
\usepackage{epsfig}                      
\usepackage{graphicx}                  
\usepackage{times}
\usepackage{mathptmx}
\usepackage{mathtools}
\usepackage{amssymb,amsmath}   

\usepackage{units}
\usepackage{color}
\usepackage[T1]{fontenc}    
\usepackage{amsfonts}       
\usepackage[utf8]{inputenc} 
\usepackage{nicefrac}       
\usepackage{comment}

\usepackage{url}  
\usepackage[pagebackref=true,breaklinks=true,colorlinks,bookmarks=false]{hyperref}
\usepackage{xspace}
\usepackage[table]{xcolor}
\usepackage{setspace}
\usepackage{grfext}
\PrependGraphicsExtensions*{.jpg,.png,.PNG}

\usepackage{kotex} 

\def\eg{e.g.,~}               
\def\ie{i.e.,~}               

\newlength\paramargin
\newlength\figmargin

\newlength\secmargin
\newlength\figcapmargin
\newlength\tabcapmargin
\newlength\tabmargin

\newcommand{\mpage}[2]
{
\begin{minipage}{#1\linewidth}\centering
#2
\end{minipage}
}

\newcommand{\secref}[1]{Section~\ref{sec:#1}}
\newcommand{\figref}[1]{Figure~\ref{fig:#1}} 
\newcommand{\tabref}[1]{Table~\ref{tab:#1}}

\newcommand{\apref}[1]{Appendix~\ref{appen:#1}}
\newcommand{\apreftab}[1]{Table~\ref{appentab:#1}}

\long\def\ignorethis#1{}

\newbox\jsavebox%

\newcommand{\ours}{\texttt{DeltaDirect}}
\newcommand{\md}{\textsc{MoDirect}}
\newcommand{\mdi}{\textsc{MoDirect-Inst}}
\newcommand{\mds}{\textsc{MoDirect-SynBench}}
\newcommand{\mdr}{\textsc{MoDirect-RealBench}}

\newcommand{\ps}{Primitive-on-Syn}
\newcommand{\pp}{Primitive-on-Real}
\newcommand{\cs}{Cutout-on-Syn}
\newcommand{\cp}{Cutout-on-Real}

\newcommand{\psyn}{P-Syn}
\newcommand{\preal}{P-Real}
\newcommand{\csyn}{C-Syn}
\newcommand{\creal}{C-Real}

\newcommand{\tocsection}[3]{\vspace{0.8em} \noindent\textbf{#1 \quad #2} \hfill \pageref{#3} \par}
\newcommand{\tocsubsection}[3]{\noindent\hspace{1.5em} #1 \quad #2 \dotfill \pageref{#3} \par}

\def\xi{\mathbf{x}_i}

\graphicspath{{figure}, {example}}

\title{Which Way Did It Move?\\
Diagnosing and Overcoming\\
Directional Motion Blindness in Video-LLMs}

\author{
Jongseo Lee$^{1*}$\hspace{0.5em}%
Hyuntak Lee$^{1*}$\hspace{0.5em}%
Sunghun Kim$^{1}$\hspace{0.5em}%
Sooa Kim$^{1}$\hspace{0.5em}%
Jihoon Chung$^{2}$\hspace{0.5em}%
Jinwoo Choi$^{1\dagger}$\\[0.3em]
$^{1}$Kyung Hee University, Republic of Korea \qquad
$^{2}$Princeton University, USA \\
{\small\ttfamily \{jong980812, takhyun03, rlatjdgns0816, suhakim12, jinwoochoi\}@khu.ac.kr} \\
{\small\ttfamily jc5933@princeton.edu}
}

\begin{document}

\maketitle

{\renewcommand{\thefootnote}{*}\footnotetext{Equally contributed first authors.}}
{\renewcommand{\thefootnote}{\textdagger}\footnotetext{Corresponding author.}}

\begin{abstract}

Video Large Language Models (Video-LLMs) have made rapid progress on temporal video understanding, yet many fail at a basic perceptual primitive: signed image-plane motion direction.
%
On simple videos of a single object moving left, right, up, or down, most Video-LLMs perform near chance, with above-chance cases largely attributable to prediction biases rather than genuine direction understanding.
We call this failure \emph{directional motion blindness}.
We localize the failure by tracing motion direction information through the Video-LLM pipeline. 
Motion direction remains linearly accessible from the vision encoder, projector, and LLM hidden states, but the readout fails to bind this signal to the correct verbal answer option, revealing a \emph{direction binding gap}.
Although synthetic motion direction instruction tuning reduces this gap on the source domain, motion direction concept vector analysis shows that visual complexity weakens the signal magnitude and limits out-of-domain generalization.
We introduce \md{}, a dataset family for motion direction instruction tuning and evaluation, and \ours{}, a diagnosis-driven, projector-level objective that predicts normalized 2-D motion vectors from adjacent-frame feature deltas.
%
On \mds{}, instruction tuning with \ours{} improves motion direction accuracy from 25.9\% to 85.9\%. 
%
On \mdr{}, \ours{} improves real-world motion direction accuracy by 21.4 points over the vanilla baseline without real-world tuning data, while preserving standard video-understanding performance.
Our project page is available at \url{https://jong980812.github.io/which-way-did-it-move/}.
\end{abstract}

%
%
%



\begin{figure}[ht]
\centering
\includegraphics[width=\linewidth]{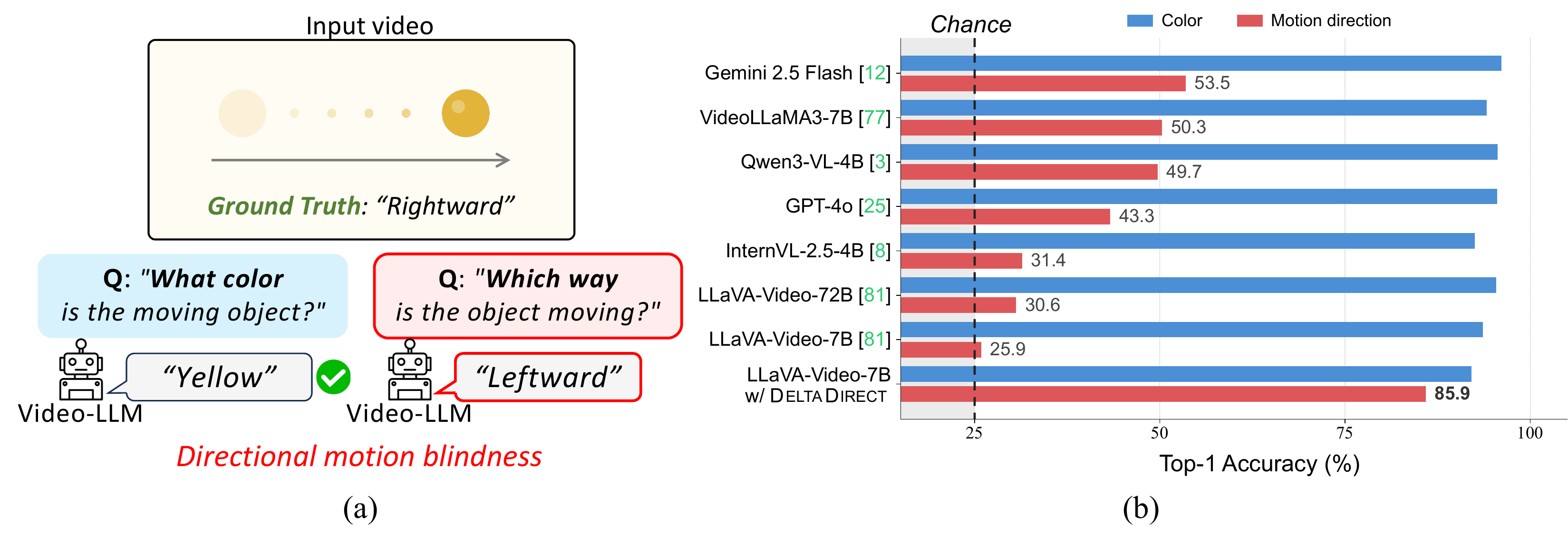}

\vspace{-0.7em}
\caption{
    \textbf{Directional motion blindness in Video-LLMs.
    }
    (a) Given a simple synthetic video of a yellow circle moving from left to right, recent Video-LLMs correctly identify the object's color but answer the wrong motion direction.
    (b) Across Video-LLMs, appearance recognition is high, yet signed motion direction accuracy remains much lower, often near chance.
}
\label{fig:teaser}
\vspace{-1em}
\end{figure}
\section{Introduction}
\label{sec:intro}

\vspace{-0.1em}
Recent work on Video-LLMs has made rapid progress in temporal understanding~\cite{tomato,tempcompass,hong2025motionbench,litemporal,backbone_temporal_reasoning} with different benchmarks analyzing a model's temporal capability by testing its memory~\cite{perceptiontest}, ability to identify complex human action or object movements~\cite{actionatlas, tempcompass,tomato}, identifying the order of stitched videos~\cite{tempcompass,vniah}, or understanding long-form videos~\cite{zhou2025mlvu,videomme,mangalam2023egoschema}.
Despite this progress, we find that many Video-LLMs~\cite{videollama2,llama-vid,llava_onevision,llava_next_video,videollava, internvideo2,mash-vlm,mvbench} fail at a much simpler temporal primitive: motion direction understanding.
Consider a seemingly trivial synthetic video in which a yellow circle moves from left to right, as shown in \figref{teaser}. 
Can a Video-LLM correctly tell which way it moves: left$\rightarrow$right, right$\rightarrow$left, bottom$\rightarrow$top, or top$\rightarrow$bottom?
To our surprise, most Video-LLMs show accuracy close to random chance (25\%), with Gemini2.5-Flash~\cite{gemini2.5flash} only reaching 53.5\% accuracy.
%
Humans identify such motion directions instantly and effortlessly, reflecting the fundamental role of motion direction in visual perception and physical navigation~\cite{born2005structure,gibson2014ecological,nakayama1985}. 
We refer to this systematic failure to distinguish signed motion directions as \emph{directional motion blindness}.

\vspace{-0.1em}
To understand where this failure arises, we trace motion direction information across the Video-LLM pipeline.
We find that the signal remains linearly accessible throughout the pipeline: motion direction is decodable from the visual encoder, the projector output, and intermediate LLM states.
The breakdown emerges at the readout position, \ie the final token representation used to predict the answer: although a motion direction signal is still linearly accessible, the model does not reliably connect it to the correct verbal response~\cite{orgadllms2025,marks2024the}.
We call this mismatch a \emph{direction binding gap}, and observe the same pattern across multiple Video-LLMs (\figref{cross_model_binding_gap}).
This suggests that directional motion blindness is not simply a failure of visual perception, but a failure to make an available motion signal usable for language-level readout.



\vspace{-0.1em}
We therefore introduce \md{}, a dataset family for signed image-plane motion direction, with three subsets: \mdi{}, \mds{}, and \mdr{}.
Its synthetic part contains four domains under a controlled 2$\times$2 design over foreground type and background type: \ps{}, \cs{}, \pp{}, and \cp{}.
Primitive videos contain rendered geometric primitives such as circles, triangles, and squares; Cutout videos contain segmented real-world object images~\citep{coco}.
Synthetic videos use uniform-color backgrounds, whereas Real videos use natural scene backgrounds~\citep{place365}.
From the simplest synthetic domain, \ps{}, we construct \mdi{} for instruction tuning.
We use all four synthetic domains as \mds{} for controlled synthetic evaluation, and use \mdr{} for real-video evaluation.

\vspace{-0.1em}
Instruction tuning on \mdi{} substantially improves on the source domain motion direction recognition. 
However, models that perform well on \ps{} still degrade on more complex domains, especially \cp{}.
This indicates that next-token prediction on motion direction instructions can improve answer binding on the source domain, \ps{}, but does not necessarily produce a strong domain-invariant motion direction signal.

\vspace{-0.1em}
We next ask why this generalization gap persists. 
%
Using difference-in-means~\cite{arditi2024refusal, marks2024the, li2023inference, rimsky-etal-2024-steering, tigges2024language}, we extract motion direction concept vectors from the readout position of each LLM layer for each domain and motion direction.
After instruction tuning, motion direction concept vectors become well aligned across domains, despite being trained only on \ps{}; for example, at certain layers, their cross-domain cosine similarity is much higher than in the baseline without motion direction instruction tuning.
Yet their magnitude drops sharply on complex domains, mirroring the observed accuracy drop. 
Moreover, restoring only the magnitude of the motion direction concept vector recovers much of the lost accuracy. 
These results suggest that the out-of-domain failure is not due to a missing motion direction geometry, but to a \emph{magnitude deficit}: 
the model learns the motion direction structure, but the signal is too weak to be reliably read out across domains.
Please refer to \secref{inst_tuning} for more details.

\vspace{-0.1em}
This diagnosis motivates \ours{}, a training-only projector-level objective that strengthens signed displacement cues at the visual-language interface.
Rather than adding motion tokens or a motion-specific encoder at inference time, \ours{} uses analytically available 2-D motion vectors from \mdi{} as supervision during training.
Specifically, \ours{} predicts normalized 2-D motion vectors from adjacent-frame projector-feature deltas, encouraging the projector output to carry a stronger signed displacement signal before it enters the LLM.
The auxiliary branch is discarded after training, leaving the test-time input format, token sequence, model architecture, and decoding procedure unchanged.

\vspace{-0.1em}
We evaluate \ours{} across synthetic, real-world, and general video understanding benchmarks. 
On \mds{}, instruction tuning on \mdi{} improves over the zero-shot model, and adding \ours{} further improves average motion direction accuracy by 6.6 points, with the largest gain of 11.5 points on \cp{}.
To test real-world transfer, we construct \mdr{}, a motion direction benchmark curated from Something-Something-V2 (SSv2)~\cite{ssv2}, TOMATO~\cite{tomato}, and KTH~\cite{kth}.
Without seeing real-world videos during tuning, \ours{} improves accuracy by 21.4 points over the vanilla baseline. 
Beyond motion direction-specific benchmarks, \ours{} preserves or improves performance on standard video understanding benchmarks~\cite{mvbench,tempcompass,perceptiontest,hong2025motionbench,tufavor,mangalam2023egoschema,zhang2024vinoground}, suggesting positive transfer rather than overfitting to synthetic motion. 

Finally, when applied during full fine-tuning, \ours{} alone yields strong improvements on both motion direction and general video benchmarks, indicating that motion-vector supervision can serve as a useful training signal for future Video-LLMs.

\vspace{-0.1em}
In this work, our major contributions are threefold:
\vspace{-0.4em}
\begin{itemize}
    \item We identify \emph{directional motion blindness} in Video-LLMs: despite strong appearance recognition, recent models fail to distinguish basic signed motion directions such as left, right, up, and down. 
    We diagnose this failure as a \emph{direction binding gap}, where motion direction information is linearly decodable but not reliably bound to the correct verbal response.

    \item We show that instruction tuning closes the direction binding gap on the source domain but leaves an out-of-domain generalization gap. 
    Through concept vector analysis, we trace this gap to a magnitude deficit in shared motion direction representations across domains.

    \item We introduce \md{}, a controlled dataset family for motion direction instruction tuning and evaluation, and propose \ours{}, a training-only auxiliary objective that predicts motion vectors from adjacent-frame feature deltas. 
    \ours{} improves motion direction understanding while preserving or improving general video understanding.
\end{itemize}

\section{Related Work}
\label{sec:related}
\vspace{-0.3em}

\vspace{\secmargin}

\paragraph{Video-LLMs and temporal-motion benchmarks.}
Video-LLMs extend image-based MLLMs to temporal understanding through video instruction tuning, unified image-video architectures, and long-video modeling~\cite{videochatgpt,videollava,llama-vid,internvideo2,st_llm,mash-vlm,videollama2,videollama3,merv,llava_onevision,qwen25vl,qwen3vl,nvila,gpt4o,
gemini2.5flash}. 
Existing benchmarks evaluate temporal memory, action and event understanding, temporal ordering, and long-form video reasoning ~\cite{mvbench,tempcompass,tomato,tvbench,mangalam2023egoschema}.
Motion direction appears in several temporal or motion-centric benchmarks, often alongside speed, order, rotation, and broader motion reasoning ~\cite{li2024vitatecs,zhang2024vinoground,hong2025motionbench,tufavor,vlm4d,
backbone_arrow_of_time,du2026motionsight}. 
These benchmarks are valuable for real-world evaluation, but real videos entangle apparent direction with camera motion, occlusion, object identity, scene layout, scale change, and event semantics.
We instead separate diagnosis from transfer: controlled synthetic videos \emph{isolate signed image-plane direction}, and real-world benchmarks test whether the intervention transfers.

\vspace{\paramargin}
\paragraph{Mechanistic analysis of LLMs.}
%
A growing line of work analyzes internal representations rather than final outputs alone. 
Probing and difference-in-means analyses show that model states can encode information that is not reliably verbalized~\cite{orgadllms2025,marks2024the}. 
For Video-LLMs, recent work begins to trace layer-wise information flow and temporal bottlenecks inside the model~\cite{mapflow,litemporal}. 
Our analysis is complementary: we isolate motion direction as a single perceptual primitive and ask \emph{whether the signal is present, where it weakens, and how it couples to verbal readout}.

\vspace{\paramargin}
\paragraph{Motion representations and supervision.}

Video models encode motion explicitly through frame differences, pose sequence, optical flow, or trajectories for action recognition~\cite{tdn,tea,motionsqueeze,dance}, flow distillation~\cite{mars,hidden_two_stream,d3d}, and self-supervised learning~\cite{motionmae,mosi}.
Recent Video-LLMs use motion-vector tokenization~\cite{video_lavit}, compressed-video cues~\cite{ema}, optical flow~\cite{flow4agent,moose}, or motion-appearance disentanglement~\cite{phyvllm}.
\ours{} instead uses motion vectors only as training-time supervision on projector-feature deltas, without changing the test-time input or decoding pipeline.

\vspace{\secmargin}
\vspace{-0.5em}


\section{Where Does Directional Motion Blindness Arise in Video-LLMs?}
\label{sec:diagnosis}


\vspace{\paramargin}
%
We investigate a Video-LLM, LLaVA-Video~\cite{llava_video}, from visual representations to the answer token, and test whether the failure is due to missing supervision, insufficient prompting, visual encoding, projection, or the final binding between a perceived direction and the corresponding answer option.
We also investigate several Video-LLMs and draw the same conclusions.
For more results and details, refer to Appendix~\ref{appen:additional_anal}.


We organize the analysis around three research questions.
\textbf{(i)} Can the failure be explained by missing direction supervision or insufficient input-side scaffolding? (\secref{data_analysis})
\textbf{(ii)} Is the signed direction erased by the vision encoder or the
vision-language projector, or the intermediate LLM layers? (\secref{perception})
\textbf{(iii)} If direction remains accessible, can the final readout token bind it to
    the prompt-specific answer option? (\secref{readout_binding})
We find that motion direction remains linearly accessible across the visual encoder, projector, and LLM hidden states, but the final readout token fails to encode the mapping from direction to answer option. 
We call this mismatch a direction binding gap.

\vspace{\secmargin}
\subsection{Experimental Setup}
\begin{figure}[t]
\centering
\includegraphics[width=\linewidth]{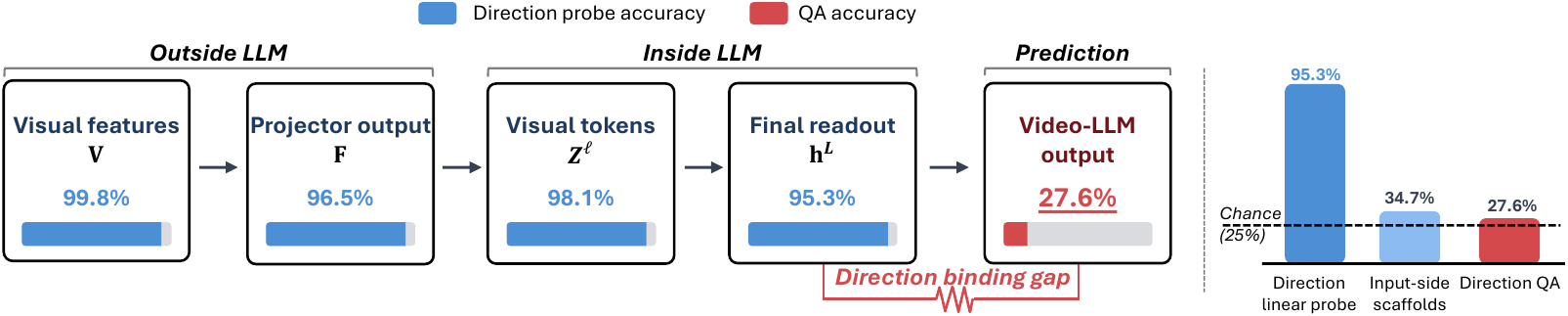}
\vspace{-0.5em}
\caption{\justifying
\textbf{Direction is decodable, but not converted into the answer.}
On \ps{} in \mds{}, motion direction remains linearly decodable throughout LLaVA-Video-7B~\cite{llava_video},
yet QA accuracy stays near chance, exposing the \emph{direction binding gap}.
}
\label{fig:direction_binding_gap}
\vspace{-1em}
\end{figure}

\vspace{-0.3em}
\label{sec:anal_setup}


\paragraph{Datasets.}
%

We introduce \mds{}, a controlled synthetic benchmark for signed
image-plane motion direction.
\mds{} follows a 2$\times$2 design over foreground type and background type: \ps{}, \cs{}, \pp{}, and \cp{}.
Each video contains a single foreground object moving in one of four signed directions: left, right, up, down.
%
We randomize start and end positions within the valid frame region and use held-out start--end trajectories for evaluation, reducing static-position shortcuts.
%
In every analysis on \mds{}, we train a linear probe on the training split and evaluate it on held-out videos.
%

\vspace{\paramargin}


\paragraph{Task.}
We use a four-way multiple-choice question (MCQ) format as the primary evaluation protocol. 
The four options correspond to left$\rightarrow$right, right$\rightarrow$left, bottom$\rightarrow$top, and top$\rightarrow$bottom.
Crucially, we randomize the order of answer options for each example.
Therefore, the correct answer letter is not a fixed alias of the motion direction; the model must bind the perceived direction to the prompt-specific option text.
%
This design enables single-token, logit-based evaluation and reduces free-form generation confounds, while directly testing whether the model can bind the perceived direction to the correct prompt-specific option~\cite{mvbench,robinson2022leveraging,mapflow}.
\vspace{\paramargin}

\paragraph{Video-LLM pipeline.}
A Video-LLM processes a video through three stages.
%
%
Given $T$ sampled frames, a vision encoder produces visual features $\mathbf{V}\in\mathbb{R}^{T\times M\times D_v}$, where $M$ denotes the number of patch tokens per frame and $D_v$ denotes the embedding dimension.
A projector maps these features into the LLM embedding space, producing visual tokens $\mathbf{F}\in\mathbb{R}^{T\times N\times D}$, where $N\leq M$ reflects spatial pooling and $D$ is the LLM hidden dimension.
An $L$-layer LLM consumes the visual tokens with text tokens and produces an answer.
Inside the LLM, we track the hidden states at the visual-token positions across layers.
Let $\mathbf{z}^{\ell}_{t} \in \mathbb{R}^{D}$ denote the spatially pooled hidden state of the visual-token positions at frame $t$ and layer $\ell \in \{1,\ldots,L\}$.
%

We refer to the hidden state at the last token position used for next-token prediction as the \emph{readout token}, and denote $\mathbf{h}^{\ell} \in \mathbb{R}^{D}$ for the readout token at layer $\ell$.
We call $\mathbf{h}^{L}$ the \emph{final readout token}; this is the representation the Language Model (LM) head maps to output vocabulary logits.

\vspace{\secmargin}
\subsection{Can Data or Prompting Explain the Failure?}
\label{sec:data_analysis}

Before inspecting model internals, we rule out two external explanations.
First, motion-direction supervision is scarce in existing video instruction data.
Using keyword pre-filtering followed by semantic classification, we estimate that only $0.91\%$ of LLaVA-Video-178K~\cite{llava_video} examples are direction-dependent; 
human verification further indicates that these estimates are upper bounds (\apref{videollms_instruction_dataset_anal}).
This scarcity motivates controlled motion-direction supervision, but it does not localize where the failure occurs.
Second, simple input-side scaffolds do not solve the task. 
We test visual boundary cues, step-by-step location reasoning~\cite{wei2022chain,kojima2022large}, and coordinate-grid prompts~\cite{mindcube,chen2024spatialvlm,cheng2024spatialrgpt};
the best combination reaches only $34.7\%$ on \ps{} as shown in~\figref{direction_binding_gap}.
Thus, external factors alone are insufficient, and we turn to representation-level diagnostics.

\vspace{\secmargin}

\subsection{Is Motion Direction Erased by the Vision Encoder or Projector?}
\label{sec:perception}



We next ask whether the failure is caused by visual processing.
We use linear probing as a diagnostic tool~\cite{orgadllms2025,li2023inference,gurnee2023language}: 
if a linear classifier can recover the signed direction from a frozen representation, then that representation contains direction information in a linearly accessible form.
This does not imply that the model can directly use the signal for generation; 
rather, it tests whether the visual-temporal evidence is present before the final readout.

\vspace{\paramargin}


\paragraph{Vision encoder preserves direction.}
We train a four-way linear probe on the frozen vision-encoder output $\mathbf{V}$. 
The probe recovers motion direction with 99.8\% accuracy (\figref{direction_binding_gap}), indicating that the visual encoder preserves a near-perfect direction signal. 
%

\vspace{\paramargin}

\vspace{-0.2em}


\paragraph{Projector also preserves direction.}
We repeat the same probe on the projector output $\mathbf{F}$.
Accuracy remains high at 96.5\%, indicating that the vision-language projection step does not erase the signed direction signal. 
Thus, directional motion blindness is not explained by a lack of linearly accessible direction information in either the vision encoder or the projector.
The failure must arise from \emph{how this signal is used or read out by the LLM}.

\vspace{\secmargin}


\subsection{Can the Readout Bind Motion Direction to the Answer Option?}
\label{sec:readout_binding}

\vspace{-0.2em}
Direction information reaches the LLM input in a linearly accessible form.
We now ask whether it remains accessible inside the LLM and whether the readout token binds this direction signal to the correct prompt-specific answer option.

\vspace{\paramargin}


\paragraph{Direction remains decodable inside the LLM.}
We probe two internal representations across LLM layers: the visual tokens $\mathbf{z}^{\ell}_{t}$ and the readout token $\mathbf{h}^{\ell}$.
As shown in \figref{direction_binding_gap}, visual-token probes achieve $98.1\%\pm1.3$, and readout token probes achieve $95.3\%\pm2.4$.
These results show that the LLM does not simply discard the direction signal.

\vspace{\paramargin}


\paragraph{The bottleneck is answer-option binding.}
The key question is whether the internally decodable direction signal reaches the correct \emph{answer option}.
Because the MCQ option order is randomized, the answer letter is not a fixed alias of the motion direction; the model must bind the perceived direction to the option text in the current prompt.
Yet the model achieves only $27.6\%$ MCQ accuracy, near the $25\%$ chance level (\figref{direction_binding_gap}).
This shows that the model carries the direction signal internally but fails to bind it to the prompt-specific answer option.

\vspace{\paramargin}
\paragraph{Direction binding gap.}
\begin{figure*}[t]
\centering
\begin{minipage}[t]{0.54\linewidth}
    \centering
    \includegraphics[width=\linewidth]{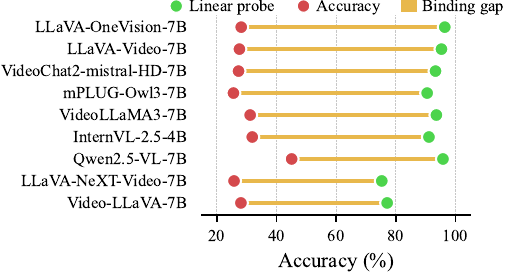}
    \vspace{-1.5em}
    \caption{\textbf{The direction binding gap is universal across Video-LLMs.}
    }
    \label{fig:cross_model_binding_gap}
\end{minipage}
\hfill
\begin{minipage}[t]{0.44\linewidth}
    \centering
    \includegraphics[width=\linewidth]{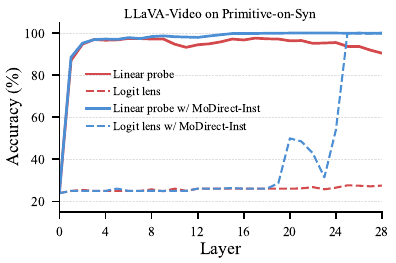}
    \vspace{-1.5em}
\caption{\justifying \textbf{Instruction tuning closes the direction binding gap (source domain).}
    }
    \label{fig:layerwise_binding_gap}
\end{minipage}
\vspace{-1em}
\end{figure*}

Taken together, these results show that the bottleneck lies not in representing motion direction, but in converting that representation into the correct answer option.
The direction signal remains accessible across the vision encoder, the projector, and the readout state, yet it is not reliably bound to the prompt-specific answer option~\cite{orgadllms2025,sun2025probing_bindinggap,park2025bridging_bindinggap}.
We call this mismatch the \emph{direction binding gap}.
As shown in~\figref{cross_model_binding_gap}, this gap persists across architectures and scales, indicating a shared structural limitation rather than a model-specific artifact.
We further verify in \apref{gap_beyond_mcq} that the same failure appears in open-ended direction-answer generation, confirming that the phenomenon is not an artifact of the MCQ format.
This diagnosis motivates \secref{inst_tuning}, where we test whether controlled motion direction instruction tuning can close the gap.

\vspace{\secmargin}

%
%
%

\section{What Instruction Tuning Solves and What It Does Not}
\label{sec:inst_tuning}
\vspace{-0.5em}


\secref{diagnosis} localizes directional motion blindness to answer-option binding: direction is linearly accessible inside the model, but the readout does not reliably link it to the correct answer option. 
We now test the most direct remedy: \emph{motion direction instruction tuning}. 
We fine-tune on the simplest synthetic domain and ask whether the learned binding transfers to more complex visual conditions.


%

\vspace{\secmargin}

\subsection{Setup: Training on \mdi{}, Testing on \mds{}}
\label{sec:finetune_setup}
\vspace{-0.2em}



\paragraph{Training data.}
We fine-tune a Video-LLM on \mdi{}, a synthetic instruction dataset built from the \ps{} domain. 
\mdi{} contains 100K training videos with diverse motion direction-related QA formats, including direction MCQ, open-ended direction, and static appearance questions (\eg object color and shape, to preserve general visual recognition).
%
%

\vspace{\paramargin}
\paragraph{Evaluation data.}
We use \mds{} as our testbed.
\mds{} includes \ps{}, the source domain used to construct \mdi{}, and three OOD domains: \pp{}, \cs{}, and \cp{}.
These OOD domains introduce natural scene backgrounds~\cite{place365}, realistic object cutouts~\cite{coco}, or both.
This split allows us to distinguish source domain binding from domain-invariant direction understanding.


\vspace{\paramargin}
\paragraph{Training.}
We use LLaVA-Video-7B~\cite{llava_video}  as the backbone, keep the vision encoder and original LLM weights frozen, and update only the projector and LoRA~\cite{lora} adapters in the LLM.

\vspace{-0.5em}
 We provide the comprehensive details of the analysis in Appendix ~\ref{appen:lora_finetune_setup}.

\vspace{\secmargin}
\subsection{Source Domain: Answer-Option Binding Emerges}
\label{sec:gap_closes}


We evaluate whether instruction tuning closes the direction binding gap on the held-out \ps{} test set.
We repeat the layer-wise probing analysis from \secref{diagnosis}.
%
%
To diagnose answer-option binding, we use the \emph{logit lens}~\cite{nostalgebraist2020logitlens,mapflow}, which projects each intermediate readout token $\mathbf{h}^{\ell}$ through the LM head without training an external classifier.
This measures whether the LM head can recover the correct answer letter at each layer.
As shown in \figref{layerwise_binding_gap}, in the vanilla model, direction is highly decodable from the readout token (\textcolor[HTML]{D4494C}{\textbf{---}}), but its logit lens accuracy stays near chance (\textcolor[HTML]{D4494C}{\textbf{- -}}), matching its low MCQ accuracy of 27.6\%.
%
After instruction tuning, logit lens accuracy (\textcolor[HTML]{4C8FD4}{\textbf{- -}}) rises in the late layers and converges to direction probing accuracy (\textcolor[HTML]{4C8FD4}{\textbf{---}}) in the final layers. 
MCQ accuracy correspondingly increases to 99.7\%. 
Thus, direction-focused instruction tuning establishes a source domain binding pathway from the internal direction signal to the correct answer option.

\begin{figure*}[t]
\centering
\includegraphics[width=\linewidth]{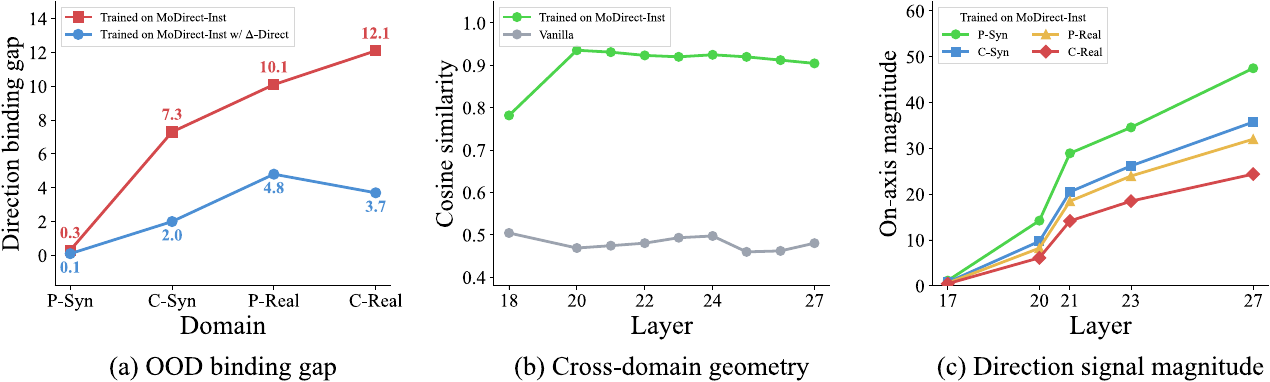}
\caption{
\textbf{Shared orientation, weak magnitude.}
\textbf{(a)} Instruction tuning closes the binding gap on \ps{}, but the gap reopens on OOD domains; \ours{} narrows it across domains.
\textbf{(b)} Direction concept vector orientations align across domains after instruction tuning, with late-layer cosine similarity exceeding $0.9$.
\textbf{(c)} Despite this alignment, concept-vector magnitude decreases with visual complexity, revealing the \emph{magnitude deficit} behind the OOD binding gap.
}
\label{fig:concept_vector_analysis}
\vspace{-0.7em}
\end{figure*}




\vspace{\secmargin}
\subsection{Out-of-Domain: The Binding Gap Reopens}
\label{sec:gap_reopens}


We next ask whether this learned binding transfers to more complex visual conditions. 
As shown in ~\figref{concept_vector_analysis} (a), instruction tuning nearly closes the gap on the source domain \ps{} test set (0.3 points), but the gap reopens on OOD domains, \eg 12.1 points on \cp{}. 
%
Although direction remains decodable from the final readout token, MCQ accuracy drops as object and background complexity increase.
This suggests that next-token supervision learns source domain answer binding, but does not produce a direction signal strong enough for reliable OOD readout.

\vspace{\secmargin}
\subsection{Why Does the Binding Gap Reopen? Shared Orientation but Insufficient Magnitude}
\label{sec:magnitude_deficit}


To understand the OOD failure, we analyze \emph{motion direction concept vectors} using difference-in-means~\cite{arditi2024refusal, marks2024the, li2023inference, rimsky-etal-2024-steering, tigges2024language} at the readout state after instruction tuning. 
For each domain $A$, direction class $d$, and layer $\ell$, we define 
$
\mathbf{v}_{d,A}^{\ell}
=
\mathbb{E}[\mathbf{h}^{\ell}\mid y=d,A]
-
\mathbb{E}[\mathbf{h}^{\ell}\mid A].
$
Here we compute the expectation over the \mds{} samples.
We decompose each vector into its unit orientation $\hat{\mathbf{v}}_{d,A}^{\ell} = \mathbf{v}_{d,A}^{\ell} / \|\mathbf{v}_{d,A}^{\ell}\|$ and magnitude $\|\mathbf{v}_{d,A}^{\ell}\|$. 
We use ``orientation'' here to distinguish the representation-space axis from the physical motion direction.
\vspace{-0.7em}
\paragraph{Orientation aligns across domains.}
To test whether the representation space orientation of each motion direction is shared across domains, we compare unit motion direction concept vectors across domains. 
For each layer $\ell$, direction $d \in \{\text{left}, \text{right}, \text{up}, \text{down}\}$, and unordered pair of domains $(A,B)$, we compute
$  \cos\!\left(\hat{\mathbf{v}}^{\ell}_{d,A}, \hat{\mathbf{v}}^{\ell}_{d,B}\right).$
Since \mds{} has four domains, this gives $\binom{4}{2}=6$ domain pairs and therefore $4 \times 6 = 24$ cosine similarity values per layer. 
The cosine similarity reported in \figref{concept_vector_analysis} (b) is the average over these values.
After instruction tuning, this average rises sharply in the late layers, indicating that the model learns a shared motion direction concept vector orientation for each motion direction across domains, despite being trained on \ps{}.

\vspace{-0.7em}
\paragraph{Magnitude drops under visual complexity.}
The aligned orientation alone is insufficient. 
As shown in \figref{concept_vector_analysis} (c), the same late layers exhibit a magnitude drop under increasing visual complexity; in particular, \cp{} shows a clear magnitude deficit relative to \ps{}.
This mirrors the OOD MCQ accuracy drop and suggests a \emph{magnitude deficit}: although domains share a motion direction concept vector orientation for each motion direction, the corresponding signal magnitude is too weak in complex domains to support reliable answer-option binding.



\vspace{-0.7em}


\paragraph{Restoring magnitude recovers accuracy.}
As a diagnostic intervention, for each OOD sample with ground-truth motion direction $d$, we rescale $\mathbf{h}^{\ell}$ along the corresponding unit concept-vector orientation $\hat{\mathbf{v}}^{\ell}_{d,A}$ to match the average \ps{} magnitude, leaving the orthogonal component unchanged.
We observe that this intervention improves OOD MCQ accuracy, with the largest gain of 15.5 points on \cp{}.
%
%
These results suggest that the OOD binding gap is not caused by a missing shared orientation across domains, but by \emph{insufficient signal magnitude}.

%
%

\vspace{-0.3em}
This motivates a training objective (\secref{deltadirect}) that does not merely teach the answer token, but makes the projector output carry a stronger signed displacement signal before it enters the LLM.

\vspace{\secmargin} 
%
%
%
%

\section{\ours{}: Strengthening Motion Signals at the VL Interface}
\label{sec:deltadirect}

\begin{figure}[t]
\centering
\includegraphics[width=\linewidth]{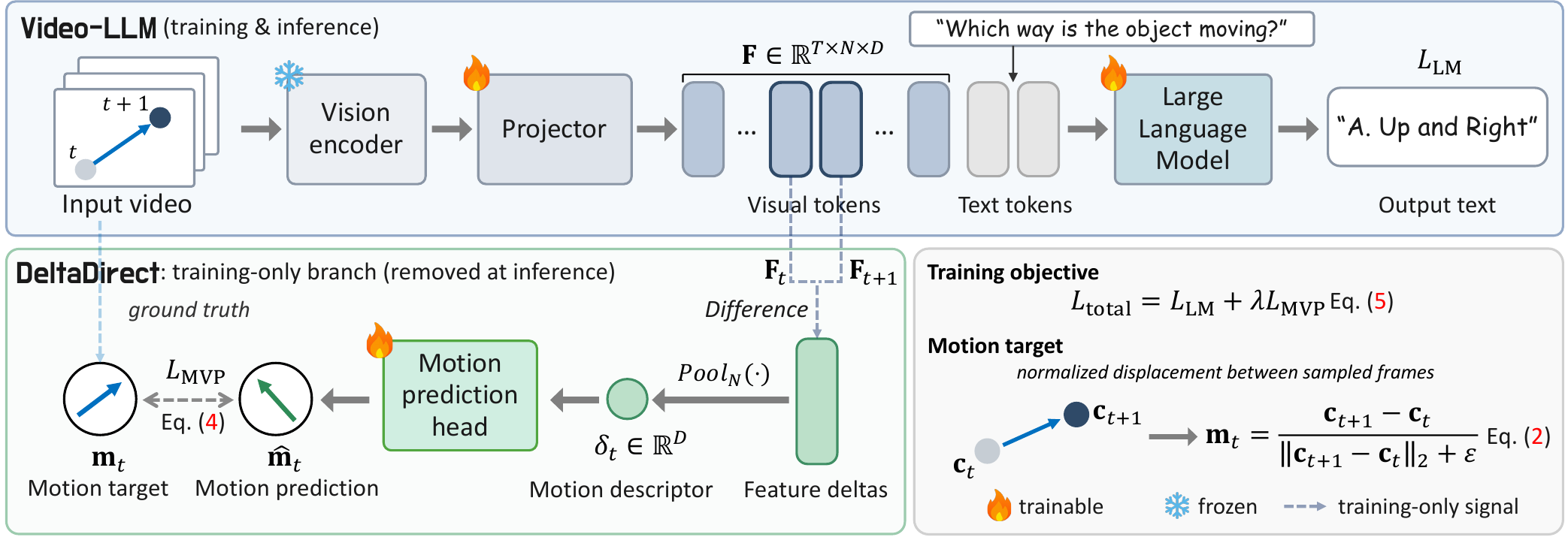}
\vspace{-0.7em}
\caption{\textbf{\ours{} Overview.}
Our diagnosis shows that visual complexity weakens the direction signal and reopens the binding gap.
This finding motivates \ours{}, which makes the projector output carry a stronger signed displacement signal before it enters the LLM.
We pool adjacent-frame feature deltas at the projector output into a motion descriptor $\boldsymbol{\delta}_t$, and a lightweight linear head predicts the normalized 2-D motion target $\mathbf{m}_t$ from it.
This auxiliary branch is used only during training and removed at inference, introducing no additional inference-time overhead.}
\label{fig:overview}
\vspace{-0.7em}
\end{figure}



The preceding sections suggest a specific intervention point.
Directional motion blindness is not caused by the complete absence of direction evidence: direction is linearly accessible from the vision encoder, projector output, and LLM hidden states (\secref{diagnosis}).
However, this signal is not reliably bound to the correct answer option. 
Moreover, motion direction instruction tuning closes the binding gap on the source domain, but visual complexity weakens the magnitude of the shared concept-vector orientation, reopening the gap in out-of-domains (\secref{inst_tuning}).


These observations motivate a simple design principle: make \emph{the projector output carry a stronger signed displacement signal} before it enters the LLM.
We therefore introduce \ours{}, a training-only auxiliary objective applied to the projector output. 
Instead of adding learned motion tokens, or a motion-specific encoder at inference time, \ours{} uses synthetic 2-D motion vectors as supervision during training. 
The auxiliary branch is discarded after training, so the test-time input format, token sequence, model architecture, and decoding procedure remain unchanged.
%


These observations lead to three design requirements. 
First, the supervision should act at the projector output, where visual tokens enter the LLM. 
Second, it should target motion-induced change rather than static appearance, so we supervise adjacent-frame feature deltas.
Third, it should supervise signed displacement directly, rather than only the answer word, so we predict normalized 2-D motion vectors. 
\ours{} instantiates these requirements with a training-only branch.




\subsection{Motion-Induced Feature Deltas}
\label{sec:delta_features}
\vspace{-0.5em}
\ours{} operates on the projector output $\mathbf{F}\in\mathbb{R}^{T\times N\times D}$. For each adjacent frame pair, we spatially pool the temporal feature difference to obtain a motion-change descriptor:
\vspace{-0em}
\begin{equation}
    \boldsymbol{\delta}_t =\frac{1}{N}\sum_{n=1}^{N}  \left(\mathbf{F}_{t+1}[n] - \mathbf{F}_{t}[n]\right) \in \mathbb{R}^{D}, 
                                                \qquad t=1,\dots,T-1.
\end{equation}
\vspace{-0em}
Feature differencing suppresses content that remains stable across adjacent frames and focuses the auxiliary branch on motion-induced change. 
We compute deltas at the projector output because these are the visual tokens directly passed to the LLM. 
This lets the auxiliary branch emphasize motion-induced changes in the same representation space used for language readout, without modifying the vision encoder or adding inference-time motion inputs.

\vspace{-0.5em}
\vspace{\secmargin}
\subsection{Training-Only Motion Vector Prediction}
\label{sec:mvp}


\vspace{-0.5em}
\paragraph{Motion vector target.}
Because \mdi{} is synthetic, the object center $\mathbf{c}_t\in\mathbb{R}^2$ is known at each frame. 
For each moving frame pair, we define the normalized 2-D motion target
\vspace{-0em}
\begin{equation}
    \mathbf{m}_t =  \frac{\mathbf{c}_{t+1}-\mathbf{c}_t} {\|\mathbf{c}_{t+1}-\mathbf{c}_t\|_2+\epsilon}  \in \mathbb{R}^{2}.
    \label{eq:mvp_target}
\end{equation}
\vspace{-0em}
This target captures the signed image-plane displacement direction while removing speed as a confounding factor. 
Rather than treating directions as discrete labels for an auxiliary classifier, 
the normalized vector target preserves the geometry of signed displacement: 
opposite motions have opposite targets, and different directions remain geometrically related in the same 2-D space.
%

\vspace{-0.5em}


\paragraph{Motion prediction head.}
A lightweight linear head predicts the motion vector from each pooled delta:
\begin{equation}
    \hat{\mathbf{m}}_t = \mathbf{W}\,\boldsymbol{\delta}_t + \mathbf{b},
    \qquad \mathbf{W} \in \mathbb{R}^{2 \times D},\;\; \mathbf{b} \in \mathbb{R}^{2}.
    \label{eq:pred_head}
\end{equation}
The head introduces only $2D+2$ trainable parameters and is used only during training.

\vspace{-0.5em}


\paragraph{MVP loss.}
We define the Motion Vector Prediction loss as

\begin{equation}
    {L}_{\mathrm{MVP}}
    = \frac{1}{T{-}1} \sum_{t=1}^{T-1}
      \left\| \hat{\mathbf{m}}_t - \mathbf{m}_t \right\|_2^{2}.
    \label{eq:mvp}
\end{equation}
\vspace{-0em}
This objective does not ask the model to memorize a direction word. 
Instead, it directly shapes the projector output so that adjacent-frame changes preserve signed displacement cues before entering the LLM.
We apply MVP supervision at the projector output by default, and compare it with intermediate-layer and readout-position supervision in~\apref{additional_exp}.

\vspace{\secmargin}
\subsection{Training Objective and Inference}
\label{sec:training_obj}
\vspace{-0.5em}

The final training objective combines standard next-token prediction with the MVP loss:
\vspace{-0em}
\begin{equation}
    {L}_{\mathrm{total}} = {L}_{\mathrm{LM}} +\lambda {L}_{\mathrm{MVP}},
    \label{eq:total}
\end{equation}
\vspace{-0em}
where ${L}_{\mathrm{LM}}$ is the next-token cross-entropy on answer tokens and $\lambda$ controls the auxiliary loss weight.
The MVP branch updates only the projector and motion prediction head; the LLM decoder is trained solely by ${L}_{\mathrm{LM}}$.
During inference, the MVP branch is removed, leaving the original input format, token sequence, and decoding procedure unchanged.%
%
%
%

\section{Experimental Results}
\label{sec:results}
\vspace{-0.7em}
We evaluate whether \ours{} turns our diagnosis into practical gains.
We focus on three questions:
\textbf{(i)} Does it reduce the diagnosed binding gap?
\textbf{(ii)} Does \ours{} improve synthetic and real-world motion-direction understanding without real-world direction supervision?
\textbf{(iii)} Does it preserve general video understanding?
Implementation details, benchmark protocols, and additional analyses and ablations are provided in Appendix ~\ref{appen:imple_details}, ~\ref{appen:additional_exp}.

\vspace{\secmargin}
\subsection{Improving Motion Direction without Sacrificing General Video Understanding}
\label{sec:main_results}

\begin{table}[t]
\centering
    \caption{\textbf{\ours{} improves motion direction understanding on both \mds{} and \mdr{}.}
    We report Top-1 accuracy (\%).
    Rows with $\pm$ report the mean and standard deviation over three seeds.
    Best results are in \textbf{bold} and second best are \underline{underlined}.
    }
    \label{tab:modirect_full}
    \setlength{\tabcolsep}{2pt}
    \resizebox{\linewidth}{!}{%
    \begin{tabular}{l cccc ccccc c}
    \toprule
    & \multicolumn{5}{c}{\mds{}} 
    & \multicolumn{4}{c}{\mdr{}} 
    & \\
    \cmidrule(lr){2-6} \cmidrule(lr){7-10}
    Method 
    & P-Syn& C-Syn& P-Real& C-Real & Avg.
    & SSv2~\cite{ssv2} & KTH~\cite{kth} & TOMATO~\cite{tomato} & Avg.
    & \makecell{Overall\\Avg.} \\
    \midrule

    Random Chance               & 25.0 & 25.0 & 25.0 & 25.0 & 25.0 & 50.0 & 50.0 & 19.0 & 39.7 & 31.3 \\
    \midrule
    \rowcolor{gray!15}\multicolumn{11}{l}{\emph{Closed-source models}} \\
    
    GPT-4o~\cite{gpt4o}                      & 46.6 & 37.8 & 48.0 & 40.7 & 43.3 & 51.0 & 52.3 & 34.5 & 45.9 & 44.4 \\
    
    Gemini 2.5 Flash~\cite{gemini2.5flash}            & 58.1 & 43.4 & 61.7 & 50.6 & 53.5 & 65.8 & \underline{69.4} & 25.6 & 53.6 & 53.5 \\


    \midrule
    \rowcolor{gray!15}\multicolumn{11}{l}{\emph{Open-source models}} \\
    Video-LLaVA-7B~\cite{videollava}              & 28.1 & 27.8 & 25.5 & 27.2 & 27.2 & 49.2 & 49.5 & 14.6 & 37.8 & 31.7 \\
    LLaMA-VID-7B~\cite{llama-vid}                & 25.3 & 25.6 & 24.7 & 25.0 & 25.2 & 51.4 & 54.2 & 16.9 & 40.8 & 31.9 \\
    LLaVA-OneVision-7B~\cite{llava_onevision}          & 28.3 & 23.4 & 29.9 & 29.3 & 27.7 & 52.2 & 50.3 & 26.8 & 43.1 & 34.3 \\
    Qwen2.5-VL-7B~\cite{qwen25vl}               & 45.1 & 32.5 & 30.8 & 30.2 & 34.7 & 52.6 & 44.2 & 25.3 & 40.7 & 37.2 \\
    Qwen2.5-VL-72B~\cite{qwen25vl}              & 69.5 & 51.3 & 57.4 & 51.1 & 57.3 & 66.8 & 62.0 & 34.2 & 54.3 & 56.0 \\
    Qwen3-VL-4B~\cite{qwen3vl}                 & 66.6 & 50.2 & 40.8 & 41.1 & 49.7 & 60.9 & 62.1 & 35.0 & 52.7 & 51.0 \\
    InternVL-2.5-4B~\cite{internvl25}             & 31.9 & 31.8 & 30.4 & 31.4 & 31.4 & 66.1 & 50.7 & 26.1 & 47.6 & 38.3 \\
    VideoLLaMA3-7B~\cite{videollama3}              & 56.4 & 46.6 & 50.0 & 48.3 & 50.3 & 56.5 & 52.2 & 19.9 & 42.9 & 47.1 \\
   
    LLaVA-Video-7B~\cite{llava_video}              & 27.6 & 23.4 & 26.9 & 25.8 & 25.9 & 52.2 & 50.3 & 26.8 & 43.1 & 33.3 \\
    LLaVA-Video-72B~\cite{llava_video}             & 36.5 & 28.3 & 28.1 & 29.3 & 30.6 & 50.3 & 56.0 & 27.3 & 44.5 & 36.5 \\
    \midrule
    \rowcolor{gray!15}\multicolumn{11}{l}{\emph{Motion-specialized models}} \\
    Video-LaVIT-7B~\cite{video_lavit}          & 24.1 & 21.0 & 23.8 & 20.2 & 22.3 & 40.3 & 39.0 & 18.6 & 32.6 & 26.7 \\
    ReMoRa-7B~\cite{yashima2026remora}            & 25.4 & 22.7 & 26.2 & 25.7 & 25.0 & 50.8 & 50.3 & 25.6 & 42.2 & 32.4 \\
    Motion-o-7B~\cite{galoaa2026motiono}          & 46.4 & 36.4 & 33.8 & 34.3 & 37.7 & 63.4 & 56.3 & 28.3 & 49.3 & 42.7 \\
    MotionLLM-7B~\cite{chen2025motionllm}         & 24.6 & 24.4 & 25.2 & 23.4 & 24.4 & 50.2 & 50.2 & 21.0 & 40.5 & 31.3 \\
    MotionSight-7B~\cite{du2026motionsight}       & 41.8 & 35.0 & 32.2 & 34.2 & 35.8 & 62.5 & 53.5 & 30.5 & 48.8 & 41.4 \\
    \midrule
    \rowcolor{gray!15}\multicolumn{11}{l}{\emph{Ours}} \\
    LLaVA-Video-7B w/ \mdi{}     & \textbf{99.7}{\scriptsize$\pm$0.2} & 81.1{\scriptsize$\pm$0.4} & 75.3{\scriptsize$\pm$0.7} & 61.0{\scriptsize$\pm$0.5} & 79.3{\scriptsize$\pm$0.4} & 72.0{\scriptsize$\pm$0.6} & 67.9{\scriptsize$\pm$1.6} & \underline{35.8}{\scriptsize$\pm$0.5} & \underline{58.6}{\scriptsize$\pm$0.5} & 70.4{\scriptsize$\pm$0.5} \\
    LLaVA-Video-7B w/ \ours{}         & \underline{99.6}{\scriptsize$\pm$0.1} & 85.5{\scriptsize$\pm$0.9} & \underline{86.1}{\scriptsize$\pm$0.9} & \underline{72.5}{\scriptsize$\pm$0.9} & \underline{85.9}{\scriptsize$\pm$0.7} & \textbf{80.7}{\scriptsize$\pm$0.9} & \textbf{74.6}{\scriptsize$\pm$0.9} & \textbf{38.1}{\scriptsize$\pm$0.7} & \textbf{64.5}{\scriptsize$\pm$0.5} & \underline{76.7}{\scriptsize$\pm$0.3} \\

    


    \midrule



    Qwen2-0.5B Full-FT w/ \mdi{}     & 99.5 & \underline{97.3} & 62.5 & 51.5 & 77.7 & 59.0 & 55.2 & 20.6 & 44.9 & 63.7 \\

    Qwen2-0.5B Full-FT w/ \ours{}     & \textbf{99.7} & \textbf{98.7} & \textbf{91.0} & \textbf{80.1} & \textbf{92.4} & \underline{77.8} & \underline{69.4} & 21.7 & 56.3 & \textbf{76.9} \\
    \bottomrule
    \end{tabular}
    \vspace{-3em}
    }
    \end{table}


\paragraph{The direction binding gap narrows.}
In \figref{concept_vector_analysis} (a), instruction tuning on \mdi{} leaves a widening binding gap ($12.1$ on \cp{}) under visual complexity.
\ours{} reduces this gap to $3.7$ points on \cp{} and narrows it on every OOD domain.

\vspace{-0.5em}
\paragraph{\ours{} substantially improves motion direction.}
\tabref{modirect_full} shows that \mdi{} tuning nearly solves \ps{} ($99.7\%$) but transfers poorly to the hardest synthetic OOD domain, \cp{} ($61.0\%$). 
\ours{} targets this gap, improving \cp{} to $72.5\%$ and raising \mds{} average accuracy from $79.3\%$ to $85.9\%$ over \mdi{} alone.
It also transfers to real videos without real-world tuning, improving \mdr{} Avg. from $58.6\%$ to $64.5\%$.
%
In full fine-tuning, the same trend holds: \ours{} raises \mds{} performance from $77.7\%$ to $92.4\%$ and \mdr{} Avg. from $44.9\%$ to $56.3\%$.

\begin{table}[t]
\centering
\vspace{-1em}
\mpage{0.42}{
\captionof{table}{
\textbf{Where to apply supervision.}
}
\label{tab:ablation_location}
\small
\setlength{\tabcolsep}{3pt}
\resizebox{\linewidth}{!}{%
\begin{tabular}{lccc}
\toprule
\multirow{2}{*}{Stage} & \multicolumn{2}{c}{\md{}} & \multirow{2}{*}{MVBench} \\
\cmidrule(lr){2-3} 
& \textsc{SynBench} & \textsc{RealBench}  \\
\midrule
None & 79.3 & 58.6 & 60.8 \\
\midrule
Vision encoder & 80.4 & 59.3 & \textbf{60.9} \\
Pre-projector & \textbf{86.1} & 62.2 & 60.2 \\
\textbf{Post-projector} & 85.9 & \textbf{64.5} & 60.8 \\
\midrule
LLM visual tokens & 70.0 & 58.3 & 57.4 \\
Final readout & 27.7 & 42.5 & 59.9 \\
\bottomrule
\end{tabular}
}
}
\hfill
\mpage{0.55}{
\captionof{table}{
\textbf{What motion signal to supervise.}
}
\label{tab:ablation_target}
\small
\setlength{\tabcolsep}{3pt}
\resizebox{\linewidth}{!}{%
\begin{tabular}{lccc}
\toprule
\multirow{2}{*}{Supervision target} & \multicolumn{2}{c}{\md{}} & \multirow{2}{*}{MVBench} \\
\cmidrule(lr){2-3} 
& \textsc{SynBench} & \textsc{RealBench}  \\
\midrule
\mdi{} tuning only & 79.3 & 58.6 & \textbf{60.8} \\
\midrule
Frame order & 24.7 & 40.3 & 54.5 \\
Concat. feature deltas & 80.1 & 58.0 & 60.7 \\
\midrule
Delta equivariance & 78.4 & 58.0 & 60.6 \\
\textbf{\ours{}} & \textbf{85.9} & \textbf{64.5} & \textbf{60.8} \\
\bottomrule
\end{tabular}
}
}
\vspace{-1em}
\end{table}

\vspace{-0.5em}

\paragraph{\ours{} preserves general video understanding.}
\begin{table}[t]

\centering
\caption{\textbf{\ours{} preserves general video understanding.}
We compare \ours{} with the base LLaVA-Video-7B on standard and fine-grained video benchmarks.
EgoSchema uses the validation set; Vinoground reports the Group score.}
\label{tab:general_video_understanding}
\vspace{0.5em}
\small
\setlength{\tabcolsep}{4pt}
\resizebox{\textwidth}{!}{%
\begin{tabular}{lcccccc}
\toprule
\multirow{2}{*}{Method} & \multicolumn{6}{c}{Standard video benchmarks} \\
\cmidrule(lr){2-7}
& MVBench~\cite{mvbench} & NExT-QA~\cite{nextqa} & PcptTest~\cite{perceptiontest} & EgoSchema~\cite{mangalam2023egoschema} & TGIF-QA~\cite{tgifqa} & Avg. \\
\midrule
LLaVA-Video-7B~\cite{llava_video}
& 59.5 & \textbf{82.7} & 64.6 & 59.4 & \textbf{80.7} & 69.4 \\
LLaVA-Video-7B w/ \ours{}
& \textbf{60.8$\pm$0.1} & 81.3$\pm$1.2 & \textbf{65.8$\pm$0.1} & \textbf{60.8$\pm$1.1} & 80.3$\pm$0.2 & \textbf{69.8$\pm$0.3} \\

\midrule
Qwen2-0.5B Full-FT w/ videochat2-it~\cite{mvbench} + \mdi{}
& 47.4 & 67.8 & 48.4 & 47.2 & \textbf{82.7} & 58.7 \\
Qwen2-0.5B Full-FT w/  \ours{}
& \textbf{48.9} & \textbf{69.3} & \textbf{49.2} & \textbf{47.6} & \textbf{82.7} & \textbf{59.5} \\
\bottomrule
\end{tabular}
}
\resizebox{\textwidth}{!}{%
\begin{tabular}{lccccc}
\toprule
\multirow{2}{*}{Method} & \multicolumn{5}{c}{Fine-grained video benchmarks} \\
\cmidrule(lr){2-6}
& TempComp.~\cite{tempcompass} & Vinoground~\cite{zhang2024vinoground} & FAVOR~\cite{tufavor} & MotionBench~\cite{hong2025motionbench} & Avg. \\
\midrule
LLaVA-Video-7B~\cite{llava_video}
& 70.5 & 15.6 & 47.2 & 56.0 & 47.3 \\
LLaVA-Video-7B w/ \ours{}
& \textbf{73.4$\pm$0.2} & \textbf{15.9$\pm$0.6} & \textbf{48.3$\pm$0.4} & \textbf{56.3$\pm$1.1} & \textbf{48.5$\pm$0.3} \\

\midrule
Qwen2-0.5B Full-FT w/ videochat2-it~\cite{mvbench} + \mdi{}
& 56.3 & 5.8 & \textbf{32.0} & 37.1 & 32.8 \\
Qwen2-0.5B Full-FT w/  \ours{}
& \textbf{57.2} & \textbf{6.8} & 30.1 & \textbf{39.1} & \textbf{33.3} \\
\bottomrule
\end{tabular}
}
\end{table}


Table~\ref{tab:general_video_understanding} shows that although \ours{} targets motion direction, it preserves aggregate general video performance in Standard~\cite{mvbench, nextqa, perceptiontest, mangalam2023egoschema, tgifqa}
video benchmarks and improves Fine-grained~\cite{tempcompass,zhang2024vinoground,tufavor,hong2025motionbench} video benchmarks.
For LLaVA-Video-7B, the Standard average stays on par ($69.4 \to 69.8$) while the Fine-grained average increases from $47.3$ to $48.5$.
In the full fine-tuning setup, \ours{} shows the same trend over \mdi{} on both averages ($58.7 \to 59.5$, $32.8 \to 33.3$).
Thus, direction supervision does not trade off broader video capability.

\vspace{\secmargin}
\subsection{Ablation Study}
\label{sec:ablation}
\paragraph{Where to apply supervision.}
In \tabref{ablation_location}, we apply MVP supervision at various locations in a Video-LLM.
Post-projector supervision performs best on average, while readout supervision performs poorly despite the binding gap being observed there. 
This supports strengthening motion signals at the vision-language interface, before the LLM transforms them for answer generation.

\vspace{-0.5em}
\paragraph{What motion signal to supervise.}
In \tabref{ablation_target}, we compare \ours{} with various supervision targets including a direction-aware delta equivariance objective. 
\ours{} performs best, indicating that signed 2-D motion-vector supervision is most effective.

%
\section{Conclusions}
\label{sec:conclusion}
\vspace{-0.5em}
In this work, we identify \emph{directional motion blindness}, a systematic failure of Video-LLMs to resolve signed image-plane motion direction.
Through controlled diagnosis on \md{}, we localize the failure to a \emph{direction binding gap}: motion direction remains linearly accessible inside the model, but is not reliably connected to the correct verbal response. 
Instruction tuning closes this gap on the source domain, but visual complexity reopens it through a magnitude deficit in the shared direction representation. 
We address this with \ours{}, a training-only objective that predicts motion vectors from adjacent-frame projector-feature deltas, strengthening signed displacement cues without changing inference. 
Trained only with synthetic motion supervision, \ours{} improves synthetic and real-world direction understanding while preserving general video understanding. 
These results suggest that diagnosis-driven supervision can help close perception--language gaps in Video-LLMs.

\paragraph{Acknowledgment.}
This work was supported by the National Research Foundation of Korea (NRF) grants funded by the Korea government (MSIT) (RS-2025-02216217, 50\%) and (RS-2026-25616405, 25\%). Additionally, it was supported in part by the Institute of Information \& Communications Technology Planning \& Evaluation (IITP) grant funded by the Korea Government (MSIT) under grant RS-2024-00353131 (10\%) and the Institute of Information \& Communications Technology Planning \& Evaluation (IITP)-ITRC (Information Technology Research Center) grant funded by the Korea government (MSIT) (IITP-2026-RS-2023-00259004, 10\%). This research was also supported by the MSIT (Ministry of Science and ICT), Korea, under the National Program for Excellence in SW (2023-0-00042, 5\%) supervised by the IITP (Institute of Information \& Communications Technology Planning \& Evaluation) in 2026.%

{\small
\bibliographystyle{plainnat}
\bibliography{main}

@STRING{NeurIPS = "Neural Information Processing Systems"}

@STRING{CVPR	= "IEEE Conference on Computer Vision and Pattern Recognition"}

@STRING{ECCV	= "European Conference on Computer Vision"}

@STRING{ICCV	= "IEEE International Conference on Computer Vision"}

@STRING{ICPR 	= "International Conference on Pattern Recognition"}

@STRING{TPAMI	= "TPAMI"}

@STRING{NeurIPS	= "NeurIPS"}

@STRING{CVPR	= "CVPR"}

@STRING{ECCV	= "ECCV"}

@STRING{ICCV	= "ICCV"}

@STRING{ICPR 	= "ICPR"}

@STRING{ICML 	= "ICML"}

@STRING{ICLR 	= "ICLR"}

@STRING{EMNLP 	= "EMNLP"}

@STRING{ACL 	= "ACL"}

@STRING{WACV 	= "WACV"}

@book{gibson2014ecological,
  title={The ecological approach to visual perception: classic edition},
  author={Gibson, James J},
  year={2014},
  publisher={Psychology press}
}

@article{born2005structure,
  title={Structure and function of visual area MT},
  author={Born, Richard T and Bradley, David C},
  journal={Annu. Rev. Neurosci.},
  volume={28},
  number={1},
  pages={157--189},
  year={2005},
  publisher={Annual Reviews}
}

@article{nakayama1985,
  author  = {Nakayama, Ken},
  title   = {Biological image motion processing: a review},
  journal = {Vision Research},
  volume  = {25},
  number  = {5},
  pages   = {625--660},
  year    = {1985},
  doi     = {10.1016/0042-6989(85)90171-3}
}

@inproceedings{orgadllms2025,
  title={LLMs Know More Than They Show: On the Intrinsic Representation of LLM Hallucinations},
  author={Orgad, Hadas and Toker, Michael and Gekhman, Zorik and Reichart, Roi and Szpektor, Idan and Kotek, Hadas and Belinkov, Yonatan},
  booktitle=ICLR,
  year={2025}
}

@article{park2025bridging_bindinggap,
  title={Bridging the Knowledge-Prediction Gap in LLMs on Multiple-Choice Questions},
  author={Park, Yoonah and Pyun, Haesung and Jo, Yohan},
  journal={arXiv preprint arXiv:2509.23782},
  year={2025}
}

@inproceedings{sun2025probing_bindinggap,
  title={Probing for arithmetic errors in language models},
  author={Sun, Yucheng and Stolfo, Alessandro and Sachan, Mrinmaya},
  booktitle=EMNLP,
  year={2025}
}

@inproceedings{marks2024the,
title={The Geometry of Truth: Emergent Linear Structure in Large Language Model Representations of True/False Datasets},
author={Samuel Marks and Max Tegmark},
booktitle={First Conference on Language Modeling},
year={2024}
}

@inproceedings{li2023inference,
  title={Inference-time intervention: Eliciting truthful answers from a language model},
  author={Li, Kenneth and Patel, Oam and Vi{\'e}gas, Fernanda and Pfister, Hanspeter and Wattenberg, Martin},
  booktitle=NeurIPS,
  year={2023}
}

@inproceedings{tigges2024language,
title={Language Models Linearly Represent Sentiment},
author={Curt Tigges and Oskar John Hollinsworth and Atticus Geiger and Neel Nanda},
booktitle=ICML,
year={2024},
}

@inproceedings{rimsky-etal-2024-steering,
  title={Steering llama 2 via contrastive activation addition},
  author={Rimsky, Nina and Gabrieli, Nick and Schulz, Julian and Tong, Meg and Hubinger, Evan and Turner, Alexander},
  booktitle={Proceedings of the 62nd Annual Meeting of the Association for Computational Linguistics (Volume 1: Long Papers)},
  year={2024}
}

@inproceedings{arditi2024refusal,
  title={Refusal in language models is mediated by a single direction},
  author={Arditi, Andy and Obeso, Oscar and Syed, Aaquib and Paleka, Daniel and Panickssery, Nina and Gurnee, Wes and Nanda, Neel},
  booktitle=NeurIPS,
  year={2024}
}

@article{llava_onevision,
title={{LL}a{VA}-OneVision: Easy Visual Task Transfer},
author={Bo Li and Yuanhan Zhang and Dong Guo and Renrui Zhang and Feng Li and Hao Zhang and Kaichen Zhang and Peiyuan Zhang and Yanwei Li and Ziwei Liu and Chunyuan Li},
journal={TMLR},
year={2025},
}

@inproceedings{llama-vid,
  title={Llama-vid: An image is worth 2 tokens in large language models},
  author={Li, Yanwei and Wang, Chengyao and Jia, Jiaya},
  booktitle=ECCV,
  year={2024},
}

@inproceedings{videollava,
  title={Video-llava: Learning united visual representation by alignment before projection},
  author={Lin, Bin and Ye, Yang and Zhu, Bin and Cui, Jiaxi and Ning, Munan and Jin, Peng and Yuan, Li},
  booktitle=EMNLP,
  year={2024}
}

@inproceedings{dance,
  title={Disentangled Concepts Speak Louder Than Words: Explainable Video Action Recognition},
  author={Lee, Jongseo and Lee, Wooil and Park, Gyeong-Moon and Kim, Seong Tae and Choi, Jinwoo},
  booktitle=NeurIPS,
  year={2025}
}

@inproceedings{internvl25,
  title={Internvl: Scaling up vision foundation models and aligning for generic visual-linguistic tasks},
  author={Chen, Zhe and Wu, Jiannan and Wang, Wenhai and Su, Weijie and Chen, Guo and Xing, Sen and Zhong, Muyan and Zhang, Qinglong and Zhu, Xizhou and Lu, Lewei and others},
  booktitle=CVPR,
  year={2024}
}

@article{qwen3vl,
  title={Qwen3-vl technical report},
  author={Bai, Shuai and Cai, Yuxuan and Chen, Ruizhe and Chen, Keqin and Chen, Xionghui and Cheng, Zesen and Deng, Lianghao and Ding, Wei and Gao, Chang and Ge, Chunjiang and others},
  journal={arXiv preprint arXiv:2511.21631},
  year={2025}
}

@article{qwen25vl,
  title={Qwen2.5-VL Technical Report},
      author={Shuai Bai and Keqin Chen and Xuejing Liu and Jialin Wang and Wenbin Ge and Sibo Song and Kai Dang and Peng Wang and Shijie Wang and Jun Tang and Humen Zhong and Yuanzhi Zhu and Mingkun Yang and Zhaohai Li and Jianqiang Wan and Pengfei Wang and Wei Ding and Zheren Fu and Yiheng Xu and Jiabo Ye and Xi Zhang and Tianbao Xie and Zesen Cheng and Hang Zhang and Zhibo Yang and Haiyang Xu and Junyang Lin},
  journal={arXiv preprint arXiv:2502.13923},
  year={2025}
}

@misc{llava_next_video,
  title={LLaVA-NeXT: A Strong Zero-shot Video Understanding Model},
  url={https://llava-vl.github.io/blog/2024-04-30-llava-next-video/},
  author={Zhang, Yuanhan and Li, Bo and Liu, haotian and Lee, Yong jae and Gui, Liangke and Fu, Di and Feng, Jiashi and Liu, Ziwei and Li, Chunyuan},
  year={2024}
}

@article{llava_video,
title={{LL}a{VA}-Video: Video Instruction Tuning With Synthetic Data},
author={Yuanhan Zhang and Jinming Wu and Wei Li and Bo Li and Zejun MA and Ziwei Liu and Chunyuan Li},
journal={TMLR},
year={2025},
}

@inproceedings{internvideo2,
  title={Internvideo2: Scaling foundation models for multimodal video understanding},
  author={Wang, Yi and Li, Kunchang and Li, Xinhao and Yu, Jiashuo and He, Yinan and Chen, Guo and Pei, Baoqi and Zheng, Rongkun and Wang, Zun and Shi, Yansong and others},
  booktitle=ECCV,
  year={2024},
}

@inproceedings{nvila,
  title     = {{NVILA}: Efficient Frontier Visual Language Models},
  author    = {Liu, Zhijian and Zhu, Ligeng and Shi, Baifeng and Zhang, Zhuoyang and Lou, Yuming and Yang, Shang and Xi, Haocheng and Cao, Shiyi and Gu, Yuxian and Li, Dacheng and Li, Xiuyu and Tang, Haotian and Fang, Yunhao and Chen, Yukang and Hsieh, Cheng-Yu and Huang, De-An and Cheng, An-Chieh and Hu, Jinyi and Liu, Sifei and Krishna, Ranjay and Molchanov, Pavlo and Kautz, Jan and Yin, Hongxu and Han, Song and Lu, Yao},
  booktitle = CVPR,
  year      = {2025}
}

@inproceedings{mplug_owl3,
  title     = {{mPLUG-Owl3}: Towards Long Image-Sequence Understanding in Multi-Modal Large Language Models},
  author    = {Ye, Jiabo and Xu, Haiyang and Liu, Haowei and Hu, Anwen and Yan, Ming and Qian, Qi and Zhang, Ji and Huang, Fei and Zhou, Jingren},
  booktitle = ICLR,
  year      = {2025}
}

@inproceedings{mash-vlm,
  title={Mash-vlm: Mitigating action-scene hallucination in video-llms through disentangled spatial-temporal representations},
  author={Bae, Kyungho and Kim, Jinhyung and Lee, Sihaeng and Lee, Soonyoung and Lee, Gunhee and Choi, Jinwoo},
  booktitle=CVPR,
  year={2025}
}

@article{gpt4o,
  title={Gpt-4o system card},
  author={Hurst, Aaron and Lerer, Adam and Goucher, Adam P and Perelman, Adam and Ramesh, Aditya and Clark, Aidan and Ostrow, AJ and Welihinda, Akila and Hayes, Alan and Radford, Alec and others},
  journal={arXiv preprint arXiv:2410.21276},
  year={2024}
}

@article{videollama2,
  title={Videollama 2: Advancing spatial-temporal modeling and audio understanding in video-llms},
  author={Cheng, Zesen and Leng, Sicong and Zhang, Hang and Xin, Yifei and Li, Xin and Chen, Guanzheng and Zhu, Yongxin and Zhang, Wenqi and Luo, Ziyang and Zhao, Deli and others},
  journal={arXiv preprint arXiv:2406.07476},
  year={2024}
}

@article{videollama3,
  title={Videollama 3: Frontier multimodal foundation models for image and video understanding},
  author={Zhang, Boqiang and Li, Kehan and Cheng, Zesen and Hu, Zhiqiang and Yuan, Yuqian and Chen, Guanzheng and Leng, Sicong and Jiang, Yuming and Zhang, Hang and Li, Xin and others},
  journal={arXiv preprint arXiv:2501.13106},
  year={2025}
}

@article{gemini2.5flash,
  title={Gemini 2.5: Pushing the frontier with advanced reasoning, multimodality, long context, and next generation agentic capabilities},
  author={Comanici, Gheorghe and Bieber, Eric and Schaekermann, Mike and Pasupat, Ice and Sachdeva, Noveen and Dhillon, Inderjit and Blistein, Marcel and Ram, Ori and Zhang, Dan and Rosen, Evan and others},
  journal={arXiv preprint arXiv:2507.06261},
  year={2025}
}

@inproceedings{merv,
title={Unifying Specialized Visual Encoders for Video Language Models},
author={Jihoon Chung and Tyler Zhu and Max Gonzalez Saez-Diez and Juan Carlos Niebles and Honglu Zhou and Olga Russakovsky},
booktitle=ICML,
year={2025}
}

@inproceedings{ssv2,
  title={The" something something" video database for learning and evaluating visual common sense},
  author={Goyal, Raghav and Ebrahimi Kahou, Samira and Michalski, Vincent and Materzynska, Joanna and Westphal, Susanne and Kim, Heuna and Haenel, Valentin and Fruend, Ingo and Yianilos, Peter and Mueller-Freitag, Moritz and others},
  booktitle=ICCV,
  year={2017}
}

@inproceedings{tomato,
    title={TOMATO: Assessing Visual Temporal Reasoning Capabilities in Multimodal Foundation Models},
    author={Shangguan, Ziyao and Li, Chuhan and Ding, Yuxuan and Zheng, Yanan and Zhao, Yilun and Fitzgerald, Tesca and Cohan, Arman},
    booktitle=ICLR,
    year={2025}
}

@inproceedings{litemporal,
  title={Temporal Reasoning Transfer from Text to Video},
  author={Li, Lei and Liu, Yuanxin and Yao, Linli and Zhang, Peiyuan and An, Chenxin and Wang, Lean and Sun, Xu and Kong, Lingpeng and Liu, Qi},
  booktitle=ICLR,
  year={2025}
}

@inproceedings{vniah,
  title={Needle In A Video Haystack: A Scalable Synthetic Evaluator for Video MLLMs},
  author={Zhao, Zijia and Lu, Haoyu and Huo, Yuqi and Du, Yifan and Yue, Tongtian and Guo, Longteng and Wang, Bingning and Liu, Jing and others},
  booktitle=ICLR,
  year={2025}
}

@inproceedings{zhou2025mlvu,
  title={Mlvu: Benchmarking multi-task long video understanding},
  author={Zhou, Junjie and Shu, Yan and Zhao, Bo and Wu, Boya and Liang, Zhengyang and Xiao, Shitao and Qin, Minghao and Yang, Xi and Xiong, Yongping and Zhang, Bo and others},
  booktitle=CVPR,
  year={2025}
}

@inproceedings{hong2025motionbench,
  title={Motionbench: Benchmarking and improving fine-grained video motion understanding for vision language models},
  author={Hong, Wenyi and Cheng, Yean and Yang, Zhuoyi and Wang, Weihan and Wang, Lefan and Gu, Xiaotao and Huang, Shiyu and Dong, Yuxiao and Tang, Jie},
  booktitle=CVPR,
  year={2025}
}

@inproceedings{videomme,
  title={Video-mme: The first-ever comprehensive evaluation benchmark of multi-modal llms in video analysis},
  author={Fu, Chaoyou and Dai, Yuhan and Luo, Yongdong and Li, Lei and Ren, Shuhuai and Zhang, Renrui and Wang, Zihan and Zhou, Chenyu and Shen, Yunhang and Zhang, Mengdan and others},
  booktitle=CVPR,
  year={2025}
}

@inproceedings{actionatlas,
  title={Actionatlas: A videoqa benchmark for domain-specialized action recognition},
  author={Salehi, Mohammadreza and Park, Jae S and Yadav, Tanush and Kusupati, Aditya and Krishna, Ranjay and Choi, Yejin and Hajishirzi, Hannaneh and Farhadi, Ali},
  booktitle=NeurIPS,
  year={2024}
}

@inproceedings{tempcompass,
  title={Tempcompass: Do video llms really understand videos?},
  author={Liu, Yuanxin and Li, Shicheng and Liu, Yi and Wang, Yuxiang and Ren, Shuhuai and Li, Lei and Chen, Sishuo and Sun, Xu and Hou, Lu},
  booktitle=ACL,
  year={2024}
}

@inproceedings{mangalam2023egoschema,
  title={Egoschema: A diagnostic benchmark for very long-form video language understanding},
  author={Mangalam, Karttikeya and Akshulakov, Raiymbek and Malik, Jitendra},
  booktitle=NeurIPS,
  year={2023}
}

@inproceedings{tufavor,
  title={FAVOR-Bench: A Comprehensive Benchmark for Fine-Grained Video Motion Understanding},
  author={Tu, Chongjun and Zhang, Lin and Chen, Pengtao and Ye, Peng and Zeng, Xianfang and Cheng, Wei and YU, Gang and Chen, Tao},
  booktitle=NeurIPS,
  year=2025
}

@article{zhang2024vinoground,
  title={Vinoground: Scrutinizing lmms over dense temporal reasoning with short videos},
  author={Zhang, Jianrui and Cai, Mu and Lee, Yong Jae},
  journal={arXiv preprint arXiv:2410.02763},
  year={2024}
}

@inproceedings{perceptiontest,
  title={Perception test: A diagnostic benchmark for multimodal video models},
  author={Patraucean, Viorica and Smaira, Lucas and Gupta, Ankush and Recasens, Adria and Markeeva, Larisa and Banarse, Dylan and Koppula, Skanda and Malinowski, Mateusz and Yang, Yi and Doersch, Carl and others},
  booktitle=NeurIPS,
  year={2023}
}

@inproceedings{li2024vitatecs,
  title={Vitatecs: A diagnostic dataset for temporal concept understanding of video-language models},
  author={Li, Shicheng and Li, Lei and Liu, Yi and Ren, Shuhuai and Liu, Yuanxin and Gao, Rundong and Sun, Xu and Hou, Lu},
  booktitle=ECCV,
  year={2024},
}

@inproceedings{EMA,
  title={Efficient motion-aware video mllm},
  author={Zhao, Zijia and Huo, Yuqi and Yue, Tongtian and Guo, Longteng and Lu, Haoyu and Wang, Bingning and Chen, Weipeng and Liu, Jing},
  booktitle=CVPR,
  year={2025}
}

@inproceedings{mapflow,
title={Map the Flow: Revealing Hidden Pathways of Information in Video{LLM}s},
author={Minji Kim and Taekyung Kim and Bohyung Han},
booktitle=ICLR,
year={2026}
}

@misc{nostalgebraist2020logitlens,
  author       = {nostalgebraist},
  title        = {Interpreting {GPT}: The Logit Lens},
  year         = {2020},
  month        = {August},
  howpublished = {LessWrong},
  url          = {https://www.lesswrong.com/posts/AcKRB8wDpdaN6v6ru/interpreting-gpt-the-logit-lens},
  note         = {Accessed: 2026-02-22}
}

@inproceedings{mvbench,
  title={Mvbench: A comprehensive multi-modal video understanding benchmark},
  author={Li, Kunchang and Wang, Yali and He, Yinan and Li, Yizhuo and Wang, Yi and Liu, Yi and Wang, Zun and Xu, Jilan and Chen, Guo and Luo, Ping and others},
  booktitle=CVPR,
  year={2024}
}

@inproceedings{videochatgpt,
  title={Video-chatgpt: Towards detailed video understanding via large vision and language models},
  author={Maaz, Muhammad and Rasheed, Hanoona and Khan, Salman and Khan, Fahad},
  booktitle={ACL},
  year={2024}
}

@inproceedings{kth,
  title={Recognizing human actions: a local SVM approach},
  author={Schuldt, Christian and Laptev, Ivan and Caputo, Barbara},
  booktitle=ICPR,
  year={2004},
}

@inproceedings{nextqa,
  title={Next-qa: Next phase of question-answering to explaining temporal actions},
  author={Xiao, Junbin and Shang, Xindi and Yao, Angela and Chua, Tat-Seng},
  booktitle=CVPR,
  year={2021}
}

@inproceedings{tgifqa,
  title={Tgif-qa: Toward spatio-temporal reasoning in visual question answering},
  author={Jang, Yunseok and Song, Yale and Yu, Youngjae and Kim, Youngjin and Kim, Gunhee},
  booktitle=CVPR,
  year={2017}
}

@inproceedings{vlm4d,
  title={Vlm4d: Towards spatiotemporal awareness in vision language models},
  author={Zhou, Shijie and Vilesov, Alexander and He, Xuehai and Wan, Ziyu and Zhang, Shuwang and Nagachandra, Aditya and Chang, Di and Chen, Dongdong and Wang, Xin Eric and Kadambi, Achuta},
  booktitle=ICCV,
  year={2025}
}

@article{place365,
  title={Places: A 10 million image database for scene recognition},
  author={Zhou, Bolei and Lapedriza, Agata and Khosla, Aditya and Oliva, Aude and Torralba, Antonio},
  journal=TPAMI,
  volume={40},
  number={6},
  pages={1452--1464},
  year={2017},
  publisher={IEEE}
}

@inproceedings{coco,
  title={Microsoft coco: Common objects in context},
  author={Lin, Tsung-Yi and Maire, Michael and Belongie, Serge and Hays, James and Perona, Pietro and Ramanan, Deva and Doll{\'a}r, Piotr and Zitnick, C Lawrence},
  booktitle=ECCV,
  year={2014}
}

@article{tvbench,
  author = {Daniel Cores and Michael Dorkenwald and Manuel Mucientes and Cees G. M. Snoek and Yuki M. Asano},
  title = {TVBench: Redesigning Video-Language Evaluation},
  journal = {arXiv:2410.07752},
  year = {2024}
}

@inproceedings{du2026motionsight,
title={MotionSight: Boosting Fine-Grained Motion Understanding in Multimodal {LLM}s},
author={Yipeng Du and Tiehan Fan and Kepan Nan and Rui Xie and Penghao Zhou and Xiang Li and Jian Yang and Zhenheng Yang and Ying Tai},
booktitle=ICLR,
year={2026}
}

@inproceedings{yashima2026remora,
  title={{ReMoRa}: Multimodal Large Language Model based on Refined Motion Representation for Long-Video Understanding},
  author={Yashima, Daichi and Kurita, Shuhei and Oda, Yusuke and Sugiura, Komei},
  booktitle=CVPR,
  year={2026}
}

@article{galoaa2026motiono,
  title={Motion-o: Trajectory-Grounded Video Reasoning},
  author={Galoaa, Bishoy and Moezzi, Shayda and Bai, Xiangyu and Ostadabbas, Sarah},
  journal={arXiv preprint arXiv:2603.18856},
  year={2026}
}

@inproceedings{backbone_temporal_reasoning,
  title={Improve Temporal Reasoning in Multimodal Large Language Models via Video Contrastive Decoding},
  author={Qi, Daiqing and Guo, Dongliang and Yuan, Hanzhang and Zhao, Handong and Hu, Mengxuan and Yang, Lehan and Li, Sheng},
  booktitle=NeurIPS,
  year={2025}
}

@inproceedings{backbone_arrow_of_time,
title={Seeing the Arrow of Time in Large Multimodal Models},
author={Zihui Xue and Mi Luo and Kristen Grauman},
booktitle=NeurIPS,
year={2025},
}

@inproceedings{lora,
title={Lo{RA}: Low-Rank Adaptation of Large Language Models},
author={Edward J Hu and yelong shen and Phillip Wallis and Zeyuan Allen-Zhu and Yuanzhi Li and Shean Wang and Lu Wang and Weizhu Chen},
booktitle=ICLR,
year={2022},
}

@inproceedings{st_llm,
  title={ST-LLM: Large Language Models Are Effective Temporal Learners},
  author={Liu, Ruyang and Li, Chen and Tang, Haoran and Ge, Yixiao and Shan, Ying and Li, Ge},
  booktitle=ECCV,
  year={2024}
}

@inproceedings{tdn,
  title={{TDN}: Temporal Difference Networks for Efficient Action Recognition},
  author={Wang, Limin and Tong, Zhan and Ji, Bin and Wu, Gangshan},
  booktitle=CVPR,
  year={2021}
}

@inproceedings{tea,
  title={{TEA}: Temporal Excitation and Aggregation for Action Recognition},
  author={Li, Yan and Ji, Bin and Shi, Xintian and Zhang, Jianguo and Kang, Bin and Wang, Limin},
  booktitle={CVPR},
  year={2020}
}

@inproceedings{motionsqueeze,
  title={{MotionSqueeze}: Neural Motion Feature Learning for Video Understanding},
  author={Kwon, Heeseung and Kim, Manjin and Kwak, Suha and Cho, Minsu},
  booktitle={ECCV},
  year={2020},
}

@inproceedings{mars,
  title={{MARS}: Motion-Augmented {RGB} Stream for Action Recognition},
  author={Crasto, Nieves and Weinzaepfel, Philippe and Alahari, Karteek and Schmid, Cordelia},
  booktitle={CVPR},
  year={2019}
}

@inproceedings{hidden_two_stream,
  title={Hidden Two-Stream Convolutional Networks for Action Recognition},
  author={Zhu, Yi and Lan, Zhenzhong and Newsam, Shawn and Hauptmann, Alexander G.},
  booktitle={ACCV},
  year={2018},
}

@inproceedings{d3d,
  title={{D3D}: Distilled {3D} Networks for Video Action Recognition},
  author={Stroud, Jonathan C. and Ross, David A. and Sun, Chen and Deng, Jia and Sukthankar, Rahul},
  booktitle=WACV,
  year={2020}
}

@article{motionmae,
  title={Self-supervised Video Representation Learning with Motion-Aware Masked Autoencoders},
  author={Yang, Haosen and Huang, Deng and Wen, Bin and Wu, Jiannan and Yao, Hongxun and Jiang, Yi and Zhu, Xiatian and Yuan, Zehuan},
  journal={arXiv preprint arXiv:2210.04154},
  year={2022}
}

@inproceedings{mosi,
  title={Self-supervised Motion Learning from Static Images},
  author={Huang, Ziyuan and Zhang, Shiwei and Jiang, Jianwen and Tang, Mingqian and Jin, Rong and Ang, Marcelo H.},
  booktitle={CVPR},
  year={2021}
}

@inproceedings{video_lavit,
  title={Video-{L}a{VIT}: Unified Video-Language Pre-training with Decoupled Visual-Motional Tokenization},
  author={Jin, Yang and Sun, Zhicheng and Xu, Kun and Xu, Kun and Chen, Liwei and Jiang, Hao and Huang, Quzhe and Song, Chengru and Liu, Yuliang and Zhang, Di and Song, Yang and Gai, Kun and Mu, Yadong},
  booktitle={ICML},
  year={2024},
}

@inproceedings{flow4agent,
  title={{Flow4Agent}: Long-form Video Understanding via Motion Prior from Optical Flow},
  author={Liu, Ruyang and Sun, Shangkun and Tang, Haoran and Li, Ge and Gao, Wei},
  booktitle={ICCV},
  year={2025}
}

@article{moose,
  title={{MOOSE}: Pay Attention to Temporal Dynamics for Video Understanding via Optical Flows},
  author={Nguyen, Hong and Tran, Dung and Hoang, Hieu and Nguyen, Phong and Narayanan, Shrikanth},
  journal={arXiv preprint arXiv:2506.01119},
  year={2025}
}

@article{phyvllm,
  title={{PhyVLLM}: Physics-Guided Video Language Model with Motion-Appearance Disentanglement},
  author={Zhan, Yu-Wei and Wang, Xin and Chen, Hong and Feng, Tongtong and Feng, Wei and Wang, Ren and Li, Guangyao and Li, Qing and Zhu, Wenwu},
  journal={arXiv preprint arXiv:2512.04532},
  year={2025}
}

@article{chen2025motionllm,
  title={Motionllm: Understanding human behaviors from human motions and videos},
  author={Chen, Ling-Hao and Lu, Shunlin and Zeng, Ailing and Zhang, Hao and Wang, Benyou and Zhang, Ruimao and Zhang, Lei},
  journal=TPAMI,
  year={2025}
}

@inproceedings{mindcube,
  title={Spatial mental modeling from limited views},
  author={Yin, Baiqiao and Wang, Qineng and Zhang, Pingyue and Zhang, Jianshu and Wang, Kangrui and Wang, Zihan and Zhang, Jieyu and Chandrasegaran, Keshigeyan and Liu, Han and Krishna, Ranjay and others},
  booktitle=ICLR,
  year={2026}
}

@inproceedings{robinson2022leveraging,
  title={Leveraging Large Language Models for Multiple Choice Question Answering},
  author={Robinson, Joshua and Rytting, Christopher Michael and Wingate, David},
  booktitle=ICLR,
  year={2023}
}

@inproceedings{fu2024ocrbenchv2improvedbenchmark,
  title={OCRBench v2: An Improved Benchmark for Evaluating Large Multimodal Models on Visual Text Localization and Reasoning},
  author={Ling Fu and Zhebin Kuang and Jiajun Song and Mingxin Huang and Biao Yang and Yuzhe Li and Linghao Zhu and Qidi Luo and Xinyu Wang and Hao Lu and Zhang Li and Guozhi Tang and Bin Shan and Chunhui Lin and Qi Liu and Binghong Wu and Hao Feng and Hao Liu and Can Huang and Jingqun Tang and Wei Chen and Lianwen Jin and Yuliang Liu and Xiang Bai},
  booktitle=NeurIPS,
  year={2025}
}

@inproceedings{wei2022chain,
  title={Chain-of-thought prompting elicits reasoning in large language models},
  author={Wei, Jason and Wang, Xuezhi and Schuurmans, Dale and Bosma, Maarten and Xia, Fei and Chi, Ed and Le, Quoc V and Zhou, Denny and others},
  booktitle=NeurIPS,
  year={2022}
}

@inproceedings{kojima2022large,
  title={Large language models are zero-shot reasoners},
  author={Kojima, Takeshi and Gu, Shixiang Shane and Reid, Machel and Matsuo, Yutaka and Iwasawa, Yusuke},
  booktitle=NeurIPS,
  year={2022}
}

@inproceedings{chen2024spatialvlm,
  title={Spatialvlm: Endowing vision-language models with spatial reasoning capabilities},
  author={Chen, Boyuan and Xu, Zhuo and Kirmani, Sean and Ichter, Brian and Driess, Danny and Florence, Pete and Sadigh, Dorsa and Guibas, Leonidas and Xia, Fei},
  booktitle=CVPR,
  year={2024}
}

@inproceedings{cheng2024spatialrgpt,
  title={Spatialrgpt: Grounded spatial reasoning in vision-language models},
  author={Cheng, An-Chieh and Yin, Hongxu and Fu, Yang and Guo, Qiushan and Yang, Ruihan and Kautz, Jan and Wang, Xiaolong and Liu, Sifei},
  booktitle=NeurIPS,
  year={2024}
}

@inproceedings{gurnee2023language,
  title={Language models represent space and time},
  author={Gurnee, Wes and Tegmark, Max},
  booktitle=ICLR,
  year={2024}
}

@inproceedings{zhang2025lmms,
  title={Lmms-eval: Reality check on the evaluation of large multimodal models},
  author={Zhang, Kaichen and Li, Bo and Zhang, Peiyuan and Pu, Fanyi and Cahyono, Joshua Adrian and Hu, Kairui and Liu, Shuai and Zhang, Yuanhan and Yang, Jingkang and Li, Chunyuan and others},
  booktitle={NAACL},
  year={2025}
}

@article{wolf2019huggingface,
  title={Huggingface's transformers: State-of-the-art natural language processing},
  author={Wolf, Thomas and Debut, Lysandre and Sanh, Victor and Chaumond, Julien and Delangue, Clement and Moi, Anthony and Cistac, Pierric and Rault, Tim and Louf, R{\'e}mi and Funtowicz, Morgan and others},
  journal={arXiv preprint arXiv:1910.03771},
  year={2019}
}

@inproceedings{siglip,
  title={Sigmoid loss for language image pre-training},
  author={Zhai, Xiaohua and Mustafa, Basil and Kolesnikov, Alexander and Beyer, Lucas},
  booktitle=CVPR,
  year={2023}
}

@article{qwen2,
  title={Qwen2 Technical Report},
  author={Yang, An and Yang, Baosong and Hui, Binyuan and Zheng, Bo and Yu, Bowen and Zhou, Chang and Li, Chengpeng and Li, Chengyuan and Liu, Dayiheng and Huang, Fei and others},
  journal={arXiv preprint arXiv:2407.10671},
  year={2024}
}
}
\newpage

\appendix

\appendix

\section*{Appendix}
\paragraph{Table of Contents}\leavevmode\\
\vspace{0.3em}
\hrule
\vspace{0.5em}
\tocsection{A}{Implementation Details}{appen:imple_details}
\tocsubsection{A.1}{Evaluation}{appen:evaluation}
\tocsubsection{A.2}{Model Card}{appen:model_card}
\tocsubsection{A.3}{Training Details: LoRA Fine-tuning}{appen:lora_finetune_setup}

\tocsubsection{A.4}{Training Details: Full Fine-tuning with \ours{}}{appen:training_from_scratch}


\tocsection{B}{Dataset and Benchmarks}{appen:md}
\tocsubsection{B.1}{\md{}} {appen:md}

\tocsubsection{B.2}{General Video Benchmarks}{appen:data_public}

\tocsection{C}{Analysis Details}{appen:anal_details}

\tocsubsection{C.1}{Linear Probing Details}{appen:probing_details}

\tocsubsection{C.2}{Logit Lens Details}{appen:logit_lens}

\tocsection{D}{Additional Analysis}{appen:videollms_instruction_dataset_anal}

\tocsubsection{D.1}{Video LLMs Instruction-tuning Dataset Analysis}{appen:videollms_instruction_dataset_anal}

\tocsubsection{D.2}{Input-side Scaffolds: Design and Full Results}{appen:input_scaffolds}


\tocsubsection{D.3}{Motion Direction Binding Gap Beyond MCQ Format}{appen:gap_beyond_mcq}

\tocsubsection{D.4}{Motion Direction Binding Gap Across Video-LLMs}{appen:binding_gap_across_model}




\tocsection{E}{Additional Experimental Results}{appen:additional_exp}
\tocsubsection{E.1}{Ablation Study of \ours{}}{appen:ablation}


\tocsubsection{E.2}{\ours{} Across Video-LLM Backbones}{appen:ablation_backbone}

\tocsubsection{E.3}{Additional Analysis of \ours{}}{appen:add_analysis_delta}

\tocsection{F}{Case Study}{appen:case_study}

\tocsection{G}{Limitations}{appen:limit}
\tocsection{H}{Broader Impacts}{appen:broader_impacts}

\vspace{0.5em}
\hrule
\vspace{0.3em}

\newpage
\section{Implementation Details}
To support full reproducibility, we will publicly release the complete training and evaluation code, along with model checkpoints and the full \md{} in the future.

This section reports the implementation details for all experiments in the paper.
Unless otherwise stated, all reported experiments follow three global conventions.
We use $8 \times$ NVIDIA A6000 48GB GPUs for training and $8 \times$ NVIDIA RTX 4090 GPUs for evaluation.
We use \texttt{lmms-eval}~\cite{zhang2025lmms} wherever the benchmark is supported; for the remaining benchmarks we follow each benchmark's official evaluation script, or implement the protocol directly when no official harness is available.
We uniformly sample $T{=}8$ frames per video at both training and inference time.
We will release our full training and evaluation code upon publication.
\label{appen:imple_details}
\subsection{Evaluation}
In this section, we provide the full evaluation details.
Unless otherwise specified, all inference runs are conducted using LMMS-Eval~\cite{zhang2025lmms} and Hugging Face~\cite{wolf2019huggingface} implementations.
We provide the sources and model cards of all evaluated models in Appendix~\ref{appen:model_card}.
We will publicly release \mds{}, \mdr{}, and all trained model weights to support reproducibility.
\label{appen:evaluation}

\subsubsection{\md{}}
We evaluate both \mds{} and \mdr{} using the LMMS-Eval framework.
For each video, we uniformly sample 8 frames and ask a four-way multiple-choice question about the signed image-plane motion direction.
The model is allowed to generate up to 16 new tokens.
We parse the generated response by first extracting an answer option, and then compare it with the ground-truth option to compute accuracy.
All evaluations are run on eight NVIDIA RTX 4090 GPUs with a per-device batch size of 1.

\subsubsection{Standard Video Benchmarks}

\paragraph{MVBench.}
MVBench~\cite{mvbench} evaluates general video understanding across diverse temporal reasoning tasks.

\paragraph{NExT-QA.}
NExT-QA~\cite{nextqa} evaluates causal and temporal reasoning in video question answering. We sample 16 frames per video and report the resulting accuracy.

\paragraph{Perception Test.}
Perception Test~\cite{perceptiontest} evaluates multimodal video understanding across perception-oriented reasoning tasks.

\paragraph{EgoSchema.}
EgoSchema~\cite{mangalam2023egoschema} evaluates long-form egocentric video understanding through multiple-choice question answering. We use the publicly released subset, sample 32 frames per video, and report accuracy.

\paragraph{TGIF-QA.}
TGIF-QA~\cite{tgifqa} evaluates short-video question answering with emphasis on actions, transitions, and temporal dynamics.

\subsubsection{Fine-grained Video Benchmarks}

\paragraph{TempCompass.}
TempCompass~\cite{tempcompass} evaluates fine-grained temporal reasoning in videos, including event order, duration, and temporal relations. We exclude the captioning subtask.

\paragraph{VinoGround.}
VinoGround~\cite{zhang2024vinoground} evaluates compositional video-language understanding by testing whether models can distinguish subtle visual and temporal differences. We sample 16 frames per video and report only the group score, the strictest of the three official metrics (text, video, and group score).

\paragraph{FAVOR-Bench.}
FAVOR-Bench~\cite{tufavor} evaluates fine-grained video understanding under temporally sensitive question-answering settings.

\paragraph{MotionBench.}
MotionBench~\cite{hong2025motionbench} evaluates motion-centric video understanding with emphasis on fine-grained motion perception and reasoning.

\subsection{Model Card}
\label{appen:model_card}

In this section, we provide the Hugging Face checkpoints for all evaluated models.
All open-source models use greedy decoding with a maximum of 1024 new tokens. Following prior multiple-choice direction evaluation settings, we use a maximum of 16 new tokens because the answer is determined from the generated option token.

\begin{table}[h]
\centering
\caption{Hugging Face checkpoints and GitHub repositories for all evaluated models.}
\label{tab:model_card}
\resizebox{\linewidth}{!}{
\begin{tabular}{l l}
\toprule
\textbf{Model} & \textbf{HF Checkpoint / GitHub} \\
\midrule
GPT                   & \texttt{\url{https://platform.openai.com/docs/models}} \\
Gemini                & \texttt{\url{https://ai.google.dev/}} \\
Qwen2.5-VL-7B         & \texttt{Qwen/Qwen2.5-VL-7B-Instruct} \\
Qwen2.5-VL-72B        & \texttt{Qwen/Qwen2.5-VL-72B-Instruct} \\
Qwen3-VL-4B           & \texttt{Qwen/Qwen3-VL-4B-Instruct} \\
VideoLLaMA3-2B        & \texttt{DAMO-NLP-SG/VideoLLaMA3-2B} \\
VideoLLaMA3-7B        & \texttt{DAMO-NLP-SG/VideoLLaMA3-7B} \\
Video-LLaVA-7B        & \texttt{LanguageBind/Video-LLaVA-7B} \\
LLaVA-OneVision-7B    & \texttt{lmms-lab/llava-onevision-qwen2-7b-ov} \\
LLaVA-Video-7B        & \texttt{lmms-lab/LLaVA-Video-7B-Qwen2} \\
LLaVA-Video-72B       & \texttt{lmms-lab/LLaVA-Video-72B-Qwen2} \\
LLaMA-VID-7B          & \texttt{YanweiLi/llama-vid-7b-full-224-video-fps-1} \\
InternVL-2.5-4B       & \texttt{OpenGVLab/InternVL2\_5-4B} \\
mPLUG-Owl3-7B         & \texttt{\url{https://github.com/x-plug/mplug-owl}} \\ 
Video-LaVIT-7B        & \texttt{rain1011/Video-LaVIT-v1} \\
ReMoRa-7B             & \texttt{naisekizero/ReMoRa} \\
Motion-o-7B           & \texttt{bishoygaloaa/motion-o} \\
MotionLLM-7B          & \texttt{\url{https://github.com/IDEA-Research/MotionLLM}} \\
MotionSight-7B        & \texttt{\url{https://github.com/NJU-PCALab/MotionSight}} \\


\midrule
\ours{} (LLaVA-Video-7B) & \texttt{\url{https://github.com/KHU-VLL/DeltaDirect}} \\
\bottomrule
\end{tabular}
}
\end{table}

\subsection{Training Details: LoRA Fine-tuning}
\label{appen:lora_finetune_setup}
We perform LoRA fine-tuning with the official LLaVA-NeXT repository\footnote{\url{https://github.com/LLaVA-VL/LLaVA-NeXT}} for LLaVA-Video-7B~\cite{llava_video} and LLaVA-OneVision-7B~\cite{llava_onevision}. LoRA fine-tuning takes approximately 7 hours.
\subsubsection{\ours{}}
We provide detailed training configurations for the \ours{} variant and the associated LoRA setup.
We use LLaVA-Video-Qwen2 as the backbone model.
This backbone consists of a Qwen2-7B~\citep{qwen2} large language model, a SigLIP~\citep{siglip} vision encoder with patch size 14 and input resolution $384 \times 384$, and a two-layer GELU-based multimodal projector.
The vision encoder is kept frozen throughout training.

\paragraph{Auxiliary Motion Direction Objective.}
\ours{} introduces an auxiliary supervision signal for motion direction.
Specifically, we compute temporal differences between consecutive frame features, apply mean pooling, and feed the resulting representation into a linear prediction head.
The auxiliary direction loss is jointly optimized with the main objective using a loss weight of $\lambda_{\text{direct}} = 1.0$.

\paragraph{Optimization Details.}
Training is performed for 1 epoch using AdamW with a learning rate of $1 \times 10^{-5}$, cosine learning-rate decay, a warmup ratio of 0.03, and zero weight decay.
We enable mixed precision training with bf16 and tf32, as well as gradient checkpointing.
The effective batch size is 144, obtained from 6 GPUs, a per-device batch size of 12, and gradient accumulation over 2 steps.
Training uses DeepSpeed ZeRO-2.

\paragraph{Fine-tuning and LoRA Setup.}
We fine-tune the multimodal projector and the large language model while keeping the vision encoder frozen.
The projector and large language model use separate learning rates of $2 \times 10^{-5}$ and $1 \times 10^{-5}$, respectively.
LoRA is applied only to the large language model, with rank $r=64$, scaling factor $\alpha=128$, and dropout 0.05.
The multimodal projector is fully fine-tuned.

\paragraph{Video Processing.}
Videos are processed using 8 frames with forced sampling.
We apply bilinear spatial pooling with stride 2 to reduce the spatial resolution of visual features.





\subsection{Training Details: Full Fine-tuning with \ours{}}
\label{appen:training_from_scratch}

\paragraph{Setup.}
We jointly optimize the vision encoder, projector, and language model of a Qwen2-0.5B~\cite{qwen2} backbone paired with a SigLIP~\cite{siglip} vision encoder and a two-layer GELU projector, initialized from the Stage-0 LLaVA-OneVision-0.5B~\cite{llava_onevision} checkpoint ($\sim$893M trainable parameters). Full fine-tuning takes approximately 52 hours.
For the training data, we subsample VideoChat2-Instruction (VideoChat2-IT)~\cite{mvbench} using the dataset configuration~\footnote{\url{https://github.com/byminji/map-the-flow/blob/main/docs/TRAIN.md}}, resulting in approximately 354K training samples.
\paragraph{Objective.}
We combine the standard language-modeling loss with the auxiliary motion-direction loss of \ours{}.
Since not all training samples contain motion-direction vector annotations, we apply the auxiliary loss only to annotated samples and mask it out otherwise.
The final objective is
\[
\mathcal{L}
=
\mathcal{L}_{\mathrm{LM}}
+
\lambda_{\mathrm{direct}} \, m \, \mathcal{L}_{\mathrm{direct}},
\]
where $m=1$ for samples with motion-direction vector annotations and $m=0$ otherwise.
\paragraph{Hyperparameters.}
We train for 2 epochs with AdamW at learning rate $2{\times}10^{-5}$ for the language model and projector, and $2{\times}10^{-6}$ for the vision encoder.
We use a cosine schedule with warmup ratio 0.03, no weight decay, and an effective batch size of 96 (batch size 2 $\times$ gradient accumulation 6 across 8 GPUs).
Each video is force-sampled to $T{=}8$ frames with spatial average pooling at stride 2.
For more details, please refer to Table~\ref{tab:hp-fullft}.

\begin{table}[ht]
\centering
\small
\caption{Hyperparameters for full fine-tuning.}
\label{tab:hp-fullft}
\begin{tabular}{ll}
\toprule
\textbf{Hyperparameter} & \textbf{Value} \\
\midrule
Epochs                       & 2 \\
Total steps                  & 9{,}438 \\
LR (LLM, projector)          & $2\times 10^{-5}$ \\
LR (vision tower)            & $2\times 10^{-6}$ \\
LR schedule / warmup ratio   & cosine / 0.03 \\
Weight decay                 & 0 \\
Optimizer                    & AdamW (fused) \\
Per-device batch size        & 2 \\
Gradient accumulation        & 6 \\
GPUs                         & 8 \\
Effective batch size         & 96 \\
Precision                    & bf16 + TF32 \\
Gradient checkpointing       & on \\
DeepSpeed                    & ZeRO Stage 2 \\
Max sequence length          & 32{,}768 \\
Frames $T$                   & 16 \\
\midrule
Training samples             & 453{,}054 \\
\hspace{1em} Synthetic Shape-Plain  & 98{,}399 \\
\hspace{1em} VideoChat2-IT~\cite{mvbench}          & 354{,}655 \\
\bottomrule
\end{tabular}
\end{table}

%
%

\newpage

\section{Dataset and Benchmarks}
\label{appen:datasets}

\subsection{\md{}}
\label{appen:md}
In the main text, we introduced \md{} as a dataset family for motion direction instruction tuning and evaluation. In this section, we provide detailed descriptions of its three components: \mdi{} for instruction tuning, \mds{} as a controlled synthetic benchmark for analysis and evaluation, and \mdr{} for real-world evaluation.




\subsubsection{\mdi{}}
\label{appen:mdi}

\begin{figure}[h]
    \centering

    \begin{subfigure}{0.75\linewidth}
        \centering
        \includegraphics[width=\linewidth]{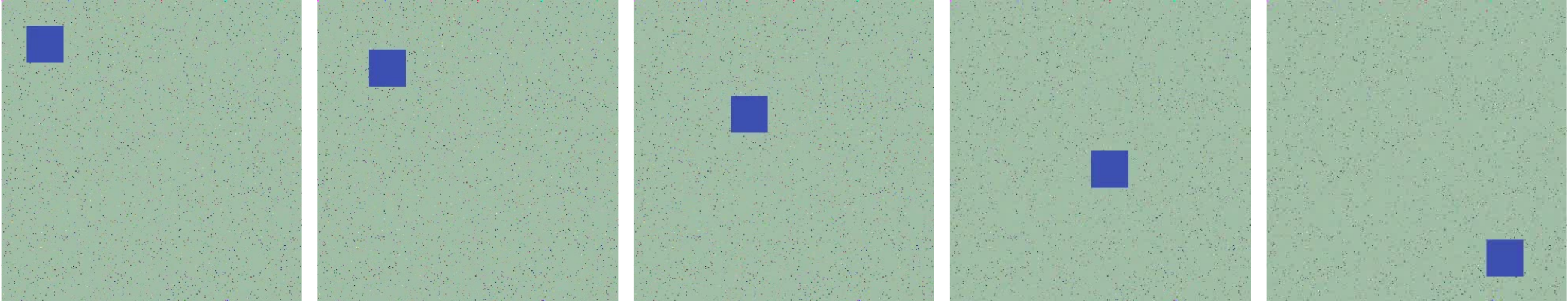}
        \caption{\mdi{} Perturbed Linear motion example}
        \label{fig:mds_example_lp}
    \end{subfigure}

    \vspace{0.4em}

    \begin{subfigure}{0.75\linewidth}
        \centering
        \includegraphics[width=\linewidth]{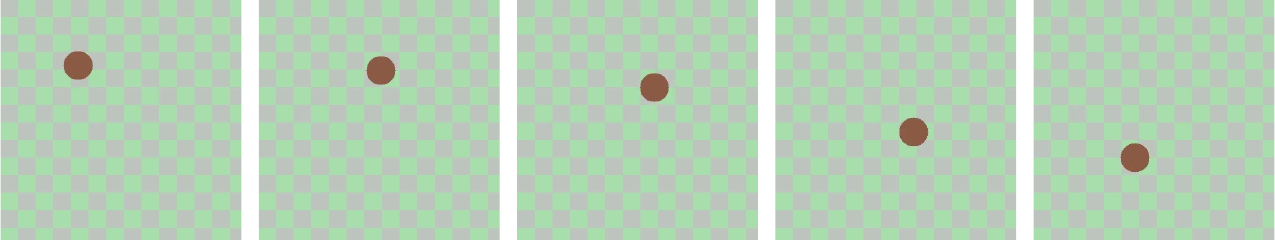}
        \caption{\mdi{} Roundtrip motion example} 
        \label{appen:oc_example_roundtrip}
    \end{subfigure}

    \caption{Examples of \mdi{}.}
    \label{fig:mdi_example}
\end{figure}

\mdi{} is a synthetic instruction-tuning dataset constructed from the \ps{} (\psyn{}) domain, which serves as the source domain for training. The dataset contains 100,000 video--question--answer (QA) pairs with diverse motion-related question formats. Each video consists of 8 frames at a resolution of $384 \times 384$, depicting a single object with diverse appearance, motion patterns, and procedural background textures.
Examples are shown in~\figref{mdi_example}.

\paragraph{QA composition and answer formats.}
The dataset contains both multiple-choice and open-ended QA formats, encouraging the model to learn both discrete decision boundaries and natural language grounding of motion concepts. The QA formats include direction prediction, motion description, trajectory understanding, object appearance recognition, location grounding, motion existence prediction, and rotation reasoning. A detailed breakdown of QA categories and their frequencies is provided in \tabref{mdi_stat} (a). Direction-related QA pairs consist of both open-ended responses and multiple-choice questions. Specifically, we include both 5-way MCQ settings (4 cardinal directions + stationary) and 9-way MCQ settings (8 directions + stationary) to support both coarse and fine-grained motion understanding. Trajectory MCQs additionally require the model to reason about compound motion patterns, such as sequential direction changes in zigzag or roundtrip trajectories. The dataset also includes appearance-related questions about object and background properties, location grounding questions based on 3$\times$3 spatial regions, and motion existence questions that determine whether an object is moving or stationary. Rotation-related QA pairs are generated only for circular-motion videos, where the model predicts clockwise (CW) or counterclockwise (CCW) motion directions using either multiple-choice or open-ended formats. Some QA categories naturally appear less frequently because they are only valid for specific motion types. For example, rotation QA pairs are generated only for circular-motion videos, while trajectory MCQs are constructed only from multi-stage motion patterns such as zigzag and roundtrip motions.

\paragraph{Dataset statistics.}
\tabref{mdi_stat} further summarizes the distributions of motion patterns, motion directions, and procedural background textures. The dataset includes diverse motion patterns, including perturbed linear motion, zigzag trajectories, circular motion, roundtrip motion, and static scenes. Motion directions are approximately uniformly distributed across all 8 directions, reducing directional bias during instruction tuning. To increase visual diversity and reduce low-level shortcut learning, the dataset incorporates diverse procedural background textures, including gradients, checkerboards, stripes, noise-based textures, paper-like patterns, and vignette-style backgrounds. The final background distribution is not perfectly uniform because all generated samples are filtered using minimum foreground--background contrast constraints to ensure object visibility.

\begin{table}[htbp]
\centering
\caption{\mdi{} statistics.}
\label{tab:mdi_stat}

\setlength{\tabcolsep}{5pt}
\renewcommand{\arraystretch}{1.1}
\small

\begin{subtable}{0.95\linewidth}
\centering
\caption{QA types}
\label{tab:mdi_qa_type}
\begin{tabular}{llc}
\toprule
QA type & Description & Cnt \\
\midrule
direction MCQ (9-way) & 8 directions + stationary, 9 choices & 17,973 \\
direction MCQ (5-way) & 4 directions + stationary, 5 choices & 17,903 \\
direction open & open-ended direction answer &  13,953 \\
description & motion description & 9,885 \\
appearance & object/background appearance description & 10,393 \\
location MCQ & $3{\times}3$ region, 9 choices & 6,461 \\
location open & open-ended location answer & 6,386 \\
move or not MCQ & yes/no multiple choice & 5,128 \\
move or not open & yes/no open-ended answer & 5,168 \\
trajectory MCQ & compound direction, 4 choices & 4,039 \\
rotation MCQ & CW/CCW multiple choice & 1,294 \\
rotation open & CW/CCW open-ended answer & 1,417 \\
\midrule
Total & -- & 100,000 \\
\bottomrule
\end{tabular}
\end{subtable}

\vspace{0.75em}

\begin{subtable}{0.31\linewidth}
\centering
\caption{Motion}
\label{tab:mdi_motion}
\begin{tabular}{lc}
\toprule
Type & Cnt \\
\midrule
perturbed linear & 29,942 \\
zigzag & 20,126 \\
circular & 19,830 \\
roundtrip & 14,972 \\
static & 15,130 \\
\bottomrule
\end{tabular}
\end{subtable}
\hfill
\begin{subtable}{0.31\linewidth}
\centering
\caption{Direction}
\label{tab:mdi_direction}
\begin{tabular}{lc}
\toprule
Dir & Cnt \\
\midrule
right & 8,268 \\
left & 8,087 \\
up & 8,122 \\
down & 8,176 \\
top-right & 8,068 \\
top-left & 8,067 \\
bottom-right & 8,083 \\
bottom-left & 8,169 \\
\bottomrule
\end{tabular}
\end{subtable}
\hfill
\begin{subtable}{0.31\linewidth}
\centering
\caption{Background}
\label{tab:mdi_background}
\begin{tabular}{lc}
\toprule
Texture & Cnt \\
\midrule
gradient & 23,528 \\
stripes & 9,840 \\
solid & 9,701 \\
checker & 9,690 \\
noise & 9,407 \\
gaussian noise & 9,407 \\
blobs & 9,096 \\
speckle & 8,953 \\
paper & 5,917 \\
vignette & 4,406 \\
\bottomrule
\end{tabular}
\end{subtable}

\end{table}

\subsubsection{\mds{}}
\label{appen:mds}

\mds{} is a controlled synthetic benchmark designed to isolate motion direction understanding. 
It follows a 2$\times$2 factorial design over foreground type and background type:
\ps{} (\psyn), \cs{} (\csyn), \pp{} (\preal), and \cp{} (\creal).

\paragraph{Dataset size and balance.}
Each domain contains 6{,}000 video samples, with 1{,}500 samples per direction (left, right, up, down). This balanced design ensures that performance differences cannot be attributed to class imbalance.

\begin{figure}[h]
    \centering

    \begin{subfigure}{0.75\linewidth}
        \centering
        \includegraphics[width=\linewidth]{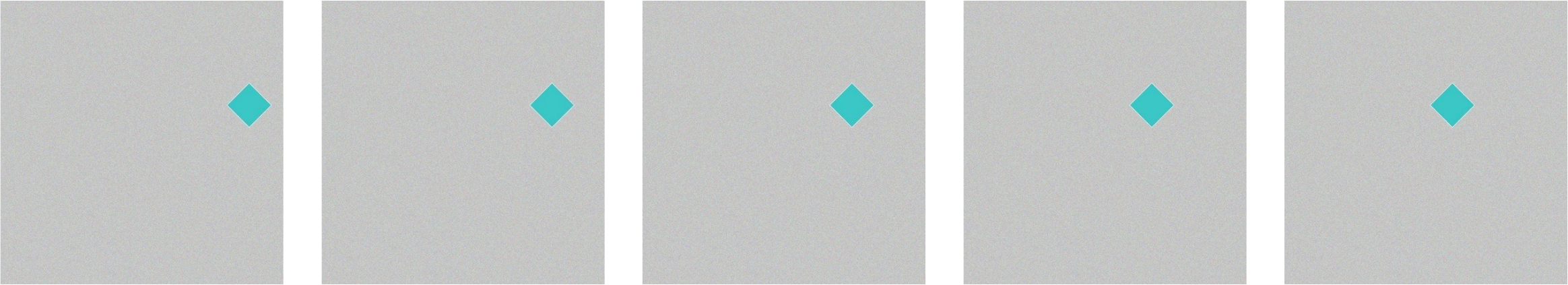}
        \caption{\ps{}}
        \label{fig:sc_example}
    \end{subfigure}

    \vspace{0.4em}

    \begin{subfigure}{0.75\linewidth}
        \centering
        \includegraphics[width=\linewidth]{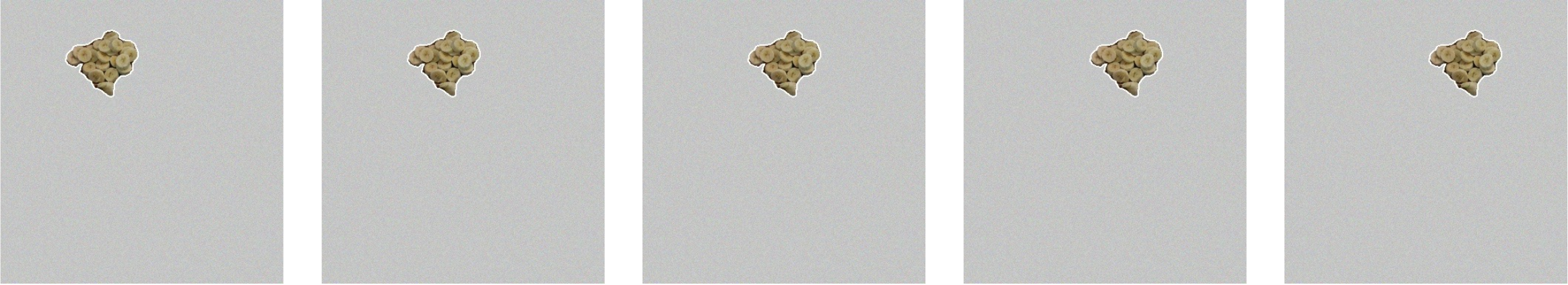}
        \caption{\cs{}}
        \label{appen:oc_example}
    \end{subfigure}
    
    \vspace{0.4em}

    \begin{subfigure}{0.75\linewidth}
        \centering
        \includegraphics[width=\linewidth]{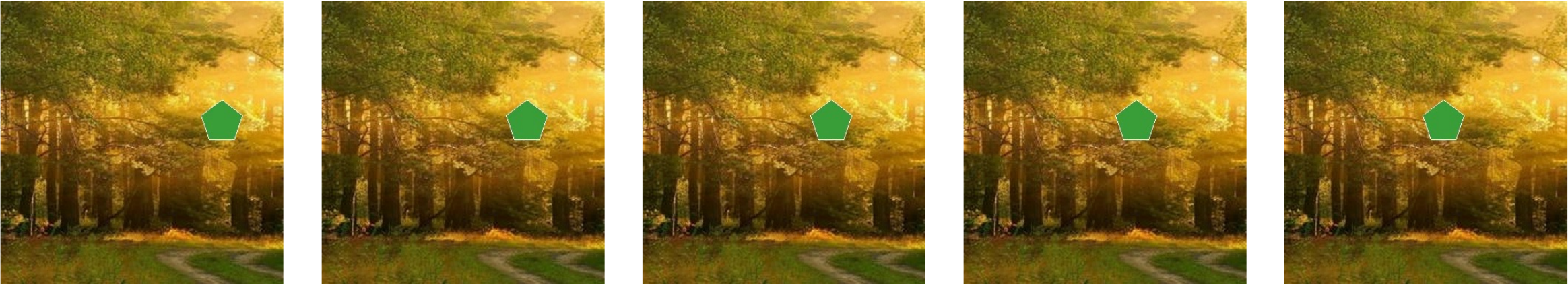}
        \caption{\pp{}}
        \label{fig:sp_example}
    \end{subfigure}
    
    \vspace{0.4em}

    \begin{subfigure}{0.75\linewidth}
        \centering
        \includegraphics[width=\linewidth]{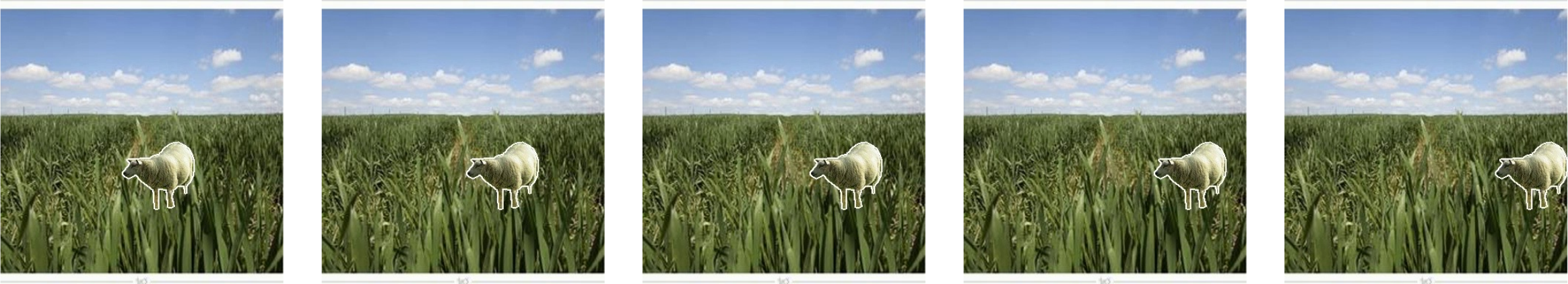}
        \caption{\cp{}}
        \label{fig:op_example} 
    \end{subfigure}

    \caption{Examples of \mds{}.}
    \label{fig:mds_example}
\end{figure}

\paragraph{Video generation.}
Each video contains a single object moving with constant velocity along a linear trajectory. The starting position is sampled uniformly within a valid region, and the velocity is chosen such that (i) the object remains within the frame for the entire sequence and (ii) the total displacement exceeds a minimum threshold. This avoids degenerate cases such as negligible motion or early exit from the frame.
The example is shown in~\figref{mds_example}.

\paragraph{Task formulation.}
We adopt a four-way MCQ format, where the answer options correspond to four signed directions. The order of answer options is randomized for each sample, so that the correct answer letter is not a fixed alias of the motion direction. This design directly tests whether the model can bind the perceived direction to the prompt-specific answer option.

\paragraph{Question format.}
Each video is paired with a direction question: \emph{``From the viewer's perspective, in which direction is the object moving in this video?''} We use a 4-way multiple-choice setting with candidates \{Left, Right, Up, Down\}. Candidate order is randomly shuffled to eliminate answer-position bias.

\subsubsection{\mdr{}}
\label{appen:mdr}
\mdr{} is a real-world motion direction benchmark constructed by curating samples 
from existing video datasets, including Something-Something-V2 (SSv2), KTH, and TOMATO. 
For SSv2 and KTH, we reformulate each video into a motion direction multiple-choice question that asks:
\emph{``From the viewer's perspective, in which direction is the something or person moving in this video?''}

\begin{figure}[h]
    \centering

    \begin{subfigure}{0.75\linewidth}
        \centering
        \includegraphics[width=\linewidth]{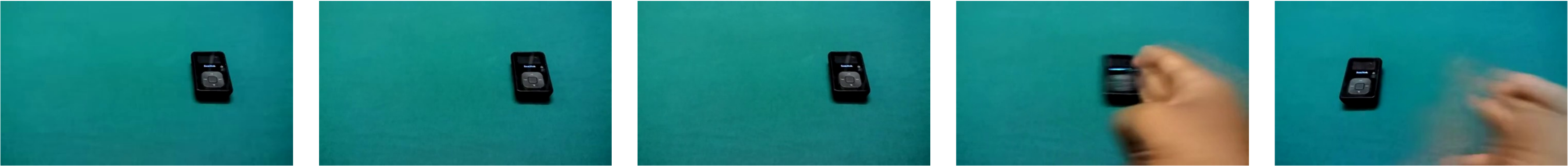}
        \caption{Something-Something v2 ~\cite{ssv2}}
        \label{fig:modirect_real_ssv2_example}
    \end{subfigure}

    \vspace{0.4em}

    \begin{subfigure}{0.75\linewidth}
        \centering
        \includegraphics[width=\linewidth]{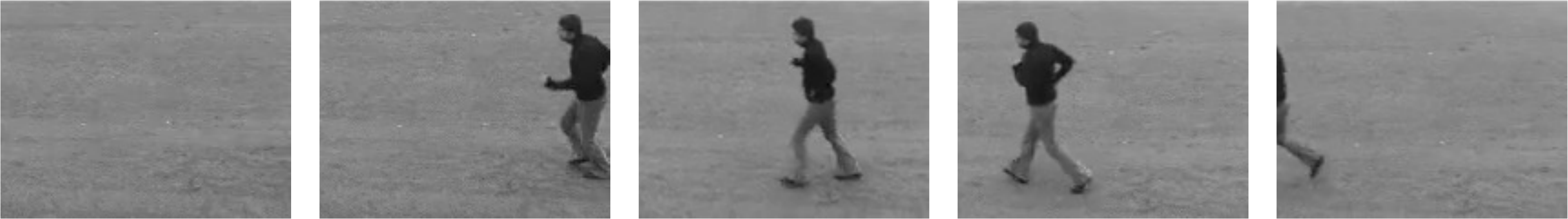}
        \caption{KTH ~\cite{kth}}
        \label{fig:modirect_real_kth_example}
    \end{subfigure}

    \vspace{0.4em}

    \begin{subfigure}{0.75\linewidth}
        \centering
        \includegraphics[width=\linewidth]{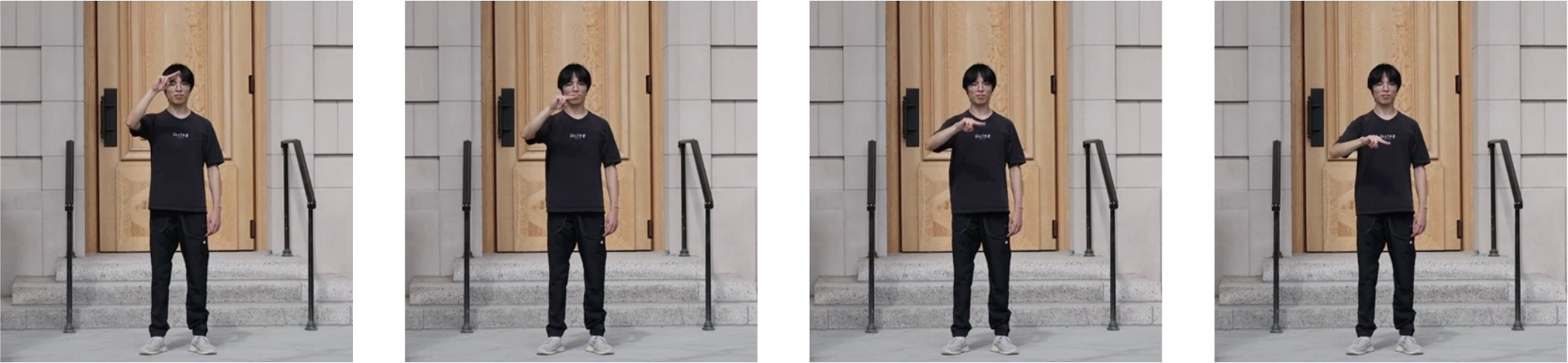}
        \caption{TOMATO ~\cite{tomato}}
        \label{fig:modirect_real_tomato_hum_example}
    \end{subfigure}

    \vspace{0.4em}
    


    \caption{Examples of \mdr{}.}
    \label{fig:mdr_example}
\end{figure}

\paragraph{Something-Something V2.}
We select all samples from four action categories that explicitly encode left--right motion:
\emph{``Pulling something from left to right,''} \emph{``Pulling something from right to left,''} \emph{``Pushing something from left to right,''} and \emph{``Pushing something from right to left.''}.  Each sample is reformulated into a binary multiple-choice question, resulting in a chance performance of 50\%. This filtering results in 722 samples. We include all available samples from the selected categories, without additional curation or filtering. The example is shown in \figref{mdr_example} (a).

\paragraph{KTH.}
We use samples from the \emph{walking}, \emph{jogging}, and \emph{running} classes. Motion direction (left vs.\ right) is determined from human skeleton trajectories by measuring the dominant horizontal displacement of key joints. This results in 899 samples. We include all videos from the selected action classes, without manual selection. Each sample is cast as a binary MCQ (left vs.\ right), yielding a chance accuracy of 50\%. The example is shown in \figref{mdr_example} (b).

\paragraph{TOMATO.}
We include 403 samples from the TOMATO~\cite{tomato} benchmark, which evaluates visual temporal reasoning across three demonstration categories: \emph{human-centric}, \emph{real-world object-centric}, and \emph{simulated} scenarios. Human-centric videos involve human body movements or interactions, such as hand gestures or full-body actions. Object-centric videos focus on the temporal dynamics of real-world objects in natural scenes. Simulated videos depict simplified synthetic environments with abstract moving entities designed to isolate temporal reasoning from appearance complexity. We use the original annotations and convert them into a unified multiple-choice format. 
Each sample is cast as an MCQ with its original answer options; since the number of options varies across questions, the expected chance accuracy is 19.0\%. Examples are shown in \figref{mdr_example} (c).

\paragraph{Evaluation format.}
All datasets are converted into a multiple-choice format to ensure consistent evaluation. This allows direct comparison with synthetic benchmarks while testing real-world generalization.

\subsection{General Video Benchmarks}
\label{appen:data_public}

\subsubsection{Standard Video Benchmarks}

\paragraph{MVBench.}
MVBench~\cite{mvbench}  is a multi-modal video understanding benchmark for evaluating temporal comprehension in Multi-modal Large Language Models (MLLMs). It consists of 20 video tasks constructed via a static-to-dynamic transformation and evaluates models using automatically generated multiple-choice questions based on video annotations.

\paragraph{NExT-QA.}
NExT-QA~\cite{nextqa} is a video question answering benchmark designed to evaluate deeper video understanding beyond description, focusing on causal reasoning, temporal action reasoning, and scene comprehension. It provides both multiple-choice and open-ended QA tasks and reveals that existing models remain weak in causal and temporal reasoning despite strong performance on shallow scene description.

\paragraph{Perception Test.}
Perception Test~\cite{perceptiontest} is a multimodal video benchmark designed to evaluate perception and reasoning abilities of pre-trained multimodal models across video, audio, and text modalities. It focuses on skills such as memory, abstraction, physics, and semantics, and provides densely annotated real-world videos with multiple-choice and grounded video question answering for comprehensive evaluation.

\paragraph{EgoSchema.}
EgoSchema~\cite{mangalam2023egoschema} is a long-form video question answering benchmark designed to evaluate long-term video understanding in vision--language models.
It contains over 5,000 human-curated multiple-choice questions derived from Ego4D videos, requiring reasoning over approximately three-minute-long clips and demonstrating the difficulty of long temporal reasoning for current models.
In this work, we use the subset of EgoSchema included in our evaluation suite and report multiple-choice accuracy following the standard evaluation protocol.

\paragraph{TGIF-QA.}
TGIF-QA~\cite{tgifqa} is a video question answering benchmark that extends visual question answering from images to videos, focusing on spatio-temporal reasoning. It introduces three video-specific QA tasks and provides a large-scale dataset designed to evaluate models’ ability to reason over temporal dynamics in videos.

\subsubsection{Fine-grained video benchmarks}

\paragraph{TempCompass.}
TempCompass~\cite{tempcompass} is a benchmark designed to evaluate temporal perception abilities of Video LLMs across diverse temporal aspects and task formats. It introduces videos that differ only in specific temporal properties and provides multiple evaluation tasks to measure nuanced temporal understanding beyond single-frame bias.

\paragraph{VinoGround.}
VinoGround~\cite{zhang2024vinoground} is a benchmark for evaluating temporal reasoning in short videos using temporal counterfactual video--caption pairs. It consists of 1,000 natural video-caption examples designed to test models’ ability to distinguish temporal differences between actions and object transformations.

\paragraph{FAVOR-Bench.}
FAVOR-Bench~\cite{tufavor} is a benchmark for evaluating fine-grained motion understanding in Multimodal Large Language Models (MLLMs). It provides manually annotated videos with both multiple-choice and open-ended tasks to assess detailed temporal motion comprehension.

\paragraph{MotionBench.}
MotionBench~\cite{hong2025motionbench} is a benchmark designed to evaluate fine-grained motion comprehension in video understanding models. It assesses motion-level perception through diverse motion-oriented question types collected from varied real-world video sources.

\newpage
\section{Analysis Details}
\label{appen:anal_details}

\subsection{Linear Probing Details}
\label{appen:probing_details}

This appendix specifies the linear probing protocol used in \secref{diagnosis}, \secref{inst_tuning}, and the cross-backbone direction binding gap analysis in \apref{binding_gap_across_model}.
The goal of this protocol is to measure whether motion-direction information is linearly accessible at different stages of the Video-LLM pipeline, rather than to introduce an additional high-capacity classifier.
We therefore use the same lightweight probe design and optimization setting across all stages and backbones.

\paragraph{Probe protocol.}
For each stage, we freeze the underlying Video-LLM and train a single linear classifier on top of the extracted features.
The probe is optimized with cross-entropy loss using standardized inputs, where the mean and standard deviation are computed only from the training split to avoid test-set leakage.
Because the probe contains no hidden layers, its performance reflects the linear decodability of the underlying representation rather than the expressive power of the probe itself.
Unless otherwise specified, we use this probing protocol for all linear-probing results reported in the main paper and appendix.
We use the same hyperparameters for all stages and all backbones, summarized in \tabref{probe_hp}.

\begin{table}[H]
\centering
\caption{Linear probing hyperparameters used throughout the paper.}
\label{tab:probe_hp}
\begin{tabular}{ll}
\toprule
Hyperparameter & Value \\
\midrule
Probe architecture & Single linear layer + softmax \\
Loss & Cross-entropy \\
Optimizer & AdamW \\
Learning rate & $10^{-3}$ \\
Weight decay & $10^{-2}$ \\
Batch size & $64$ \\
Epochs & $50$ \\
Train--test split & $70 / 30$ stratified \\
\bottomrule
\end{tabular}
\end{table}

\paragraph{Pipeline stages.}
We probe four stages along the Video-LLM pipeline.
First, we probe the visual encoder output $\mathbf{V} \in \mathbb{R}^{T \times M \times D_v}$.
Second, we probe the projector output $\mathbf{F} \in \mathbb{R}^{T \times N \times D}$, which maps visual features into the language-model embedding space.
Third, we probe the LLM visual-token hidden state $\mathbf{z}^{\ell}_{t} \in \mathbb{R}^{D}$ at frame $t$ and layer $\ell$.
Following \secref{anal_setup}, $\mathbf{z}^{\ell}_{t}$ denotes the spatially pooled hidden state over the visual-token positions corresponding to frame $t$.
Finally, we probe the readout token hidden state $\mathbf{h}^{\ell} \in \mathbb{R}^{D}$ at layer $\ell$.
The readout token is the last token in the prefill and is directly used by the LM head for next-token prediction.
In our MCQ template, this position immediately precedes the answer character, making it the relevant representation for analyzing answer-option binding.

\subsection{Logit Lens Details}
\label{appen:logit_lens}

While linear probing verifies whether motion direction is linearly decodable from an internal representation, it does not test whether that representation is already aligned with the LM head for answer selection. To measure this alignment, we use the logit lens at the readout token position.

\paragraph{Formulation.}
The logit lens~\citep{nostalgebraist2020logitlens} feeds an intermediate hidden state directly to the frozen LM head, without any additional training, to inspect what the model would predict if decoding stopped at that layer. For layer $\ell \in \{1,\ldots,L\}$, we compute the logit-lens letter prediction at the readout token position as
\begin{equation}
\hat{y}^{\ell}_{\text{lens}}
=
\arg\max_{c \in \{A,B,C,D\}}
\left( \mathbf{W}_u \cdot \mathrm{Norm}(\mathbf{h}^{\ell}) \right)_{t_c},
\label{eq:logit_lens}
\end{equation}
where $\mathbf{h}^{\ell} \in \mathbb{R}^{D}$ is the readout token hidden state at layer $\ell$,
$\mathrm{Norm}(\cdot)$ is the final pre-unembedding normalization layer of the LLM,
$\mathbf{W}_u \in \mathbb{R}^{V \times D}$ is the frozen LM head, and
$t_c$ is the vocabulary index of answer-letter token $c$.

\paragraph{Protocol.}
Unless otherwise specified, we use the same prompt format, video preprocessing, answer-token candidates, and held-out evaluation split as in the corresponding MCQ evaluation.
We apply the frozen final normalization layer and LM head to each intermediate readout-token hidden state and evaluate whether the highest-scoring answer-letter token matches the ground-truth option.
No additional parameters are trained for this analysis.
\paragraph{Interpretation.}
Unlike a learned linear probe, the logit lens evaluates whether the model's own LM head can already read out the correct answer from a given layer.
Thus, a gap between linear-probe accuracy and logit-lens accuracy indicates that motion-direction information is present in the representation but is not yet aligned with the answer-selection space.


\newpage

\section{Additional Analysis}
\label{appen:additional_anal}



\subsection{Video LLMs Instruction-tuning Dataset Analysis}
\label{appen:videollms_instruction_dataset_anal}
This section provides details of the instruction-data analysis summarized in \secref{data_analysis}. We describe the direction supervision estimation pipeline, its validation via human annotation, and the resulting dataset statistics.
\begin{table}[h]
    \centering
    \caption{\textbf{Motion direction supervision is scarce in video instruction-tuning data.} 
    Proportion of QA pairs identified by two stages of the Direction Supervision Estimation pipeline. 
    \textbf{Keyword Matched} reports the number of QA pairs that contain explicit directional keywords and the corresponding total number of QA pairs in each dataset. 
    \textbf{Semantically Direction-dependent QA} reports the subset of keyword-matched QA pairs that require motion direction understanding after semantic filtering, where the total corresponds to the size of the keyword-matched subset. \textbf{Final Ratio} denotes the proportion of semantically direction-dependent QA pairs relative to the total number of QA pairs in the dataset.}
    \label{tab:data_analysis}
    \resizebox{\linewidth}{!}{
    \begin{tabular}{lccccc}
        \toprule
        \multirow{2}{*}{Dataset}
        & \multicolumn{2}{c}{Keyword Matched}
        & \multicolumn{2}{c}{Semantically Direction-dependent QA}
        & \multirow{2}{*}{Final Ratio} \\
        \cmidrule(lr){2-3}
        \cmidrule(lr){4-5}
        & Matched & Total
        & Directional QA & Matched
        & \\
        \midrule
        VideoChat2-IT~\citep{mvbench} & 143,759 & 957,999 & 18,134 & 143,759 & 1.89\% \\
        VideoChatGPT100K~\citep{videochatgpt} & 23,156 & 99,814 & 989 & 23,156 & 0.99\% \\
        LLaVA-OneVision (video)~\citep{llava_onevision} & 47,240 & 487,632 & 3,766 & 47,240 & 0.77\% \\
        LLaVA-Video-178K~\citep{llava_video} & 802,700 & 5,699,294 & 51,776 & 802,700 & 0.91\% \\
        \bottomrule
    \end{tabular}

}

\end{table}

\subsubsection{Motion Direction Supervision Estimation Pipeline}
\label{appen:motion_direction_supervision_pipline}
\paragraph{Keyword pre-filtering for motion direction.}
The instruction data contains a diverse set of QA pairs. To efficiently identify candidate samples, we first apply a lightweight pre-filtering step based on keyword matching. We define a keyword set $\mathcal{K} = \{\text{left, right, up, down}\}$, which captures the primary horizontal and vertical directions. A QA pair is retained if any keyword in $\mathcal{K}$ appears as a case-insensitive whole-word match in either the question or the ground-truth answer. This step intentionally retains a broad set of candidate QA pairs, including cases where directional terms appear without requiring motion direction reasoning. For example, words such as ``right'' may refer to correctness rather than spatial direction, and phrases like ``close up'' may include directional terms without implying object motion. These cases cannot be reliably disambiguated by keyword matching alone and are further filtered in the subsequent semantic classification stage.
\paragraph{Semantic filtering for motion direction.}
We apply a semantic filtering step to determine whether a QA pair truly requires motion direction understanding. We formulate this as a binary classification problem, where each QA pair is labeled as \textit{directional} or \textit{non-directional}. The classifier takes both the question and the ground-truth answer as input, and predicts whether motion direction reasoning is required to answer the question correctly. We implement this classifier using a GPT-4o model via API. This step filters out semantically irrelevant cases that pass the keyword pre-filtering, yielding a refined subset of direction-dependent QA pairs.

\begin{figure}[ht]
    \centering
    \begin{subfigure}[t]{0.48\linewidth}
        \centering
        \includegraphics[width=\linewidth]{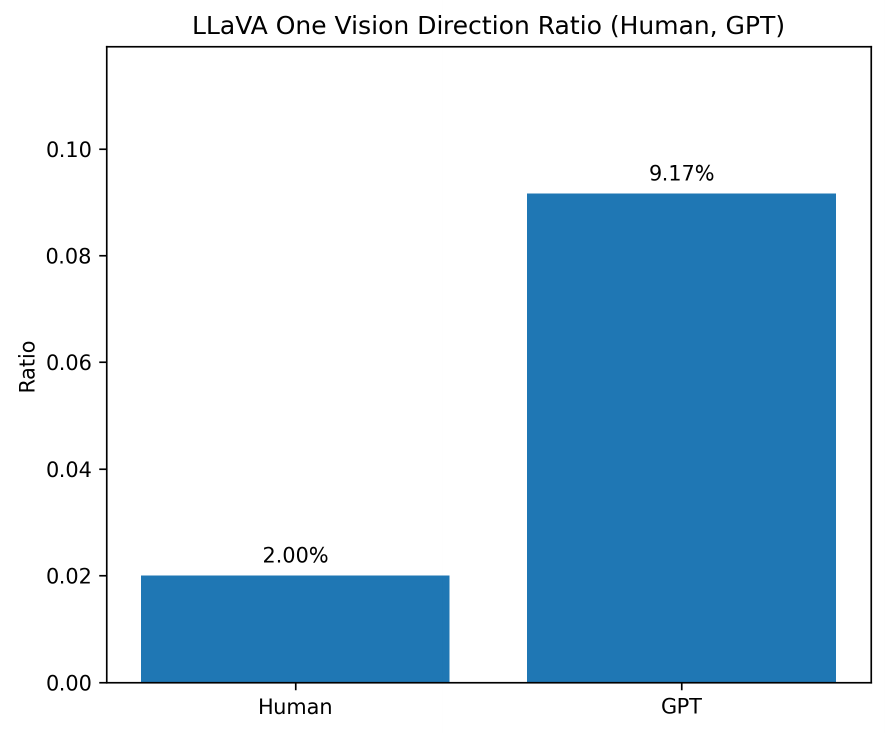}
        \caption{}
        \label{fig:llava_onevision_ratio}
    \end{subfigure}
    \begin{subfigure}[t]{0.48\linewidth}
        \centering
        \includegraphics[width=\linewidth]{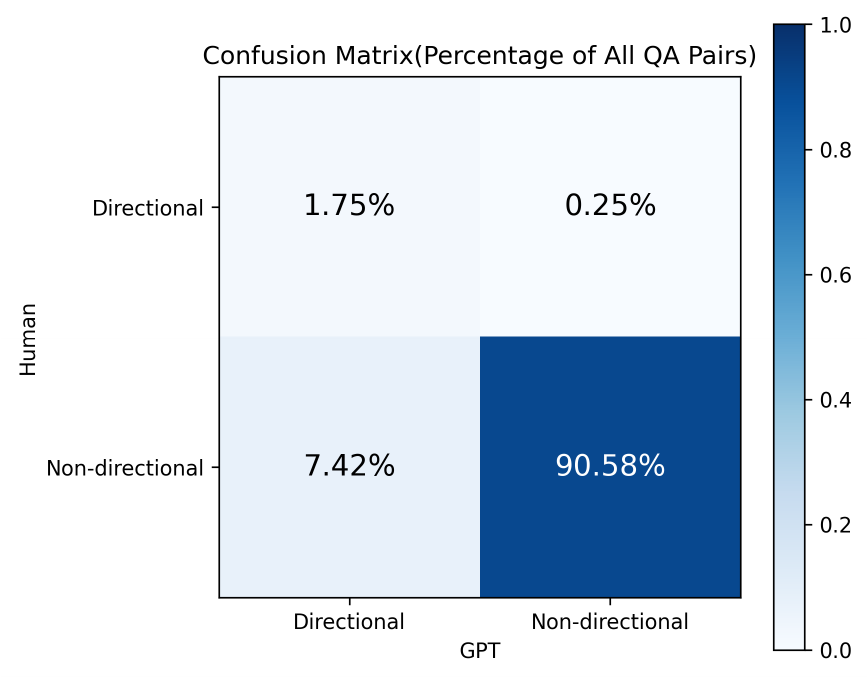}
        \caption{}
        \label{fig:llava_onevision_confmat}
    \end{subfigure}
    \hfill
    \caption{\textbf{LLM-based semantic classification tends to over-predict direction.}
    (a) Comparison of overall direction ratios. While human annotations show that only a small fraction of QA pairs require direction understanding, the LLM predicts direction at a substantially higher rate, further supporting the over-prediction tendency.
    (b) Confusion matrix comparing LLM predictions with human annotations. The model more often labels non-direction QA pairs as direction than it misses true direction cases, indicating a tendency to over-predict direction.}
    \label{fig:llava_onevision_combined}
\end{figure}

\subsubsection{Reliability of Semantic Filtering}
\label{appen:gpt_reliable}
\paragraph{Human verification.}
Since GPT-based semantic filtering may cause classification errors, we validate its reliability against human annotations. We randomly sample 5{,}000 QA pairs from the LLaVA-OneVision instruction dataset~\cite{llava_onevision} and label each pair as \textit{directional} or \textit{non-directional}. Annotators follow the same labeling criteria used for the GPT-based classifier and are not exposed to its predictions. The resulting annotations are used to analyze agreement and error patterns between GPT and human judgments. 
\paragraph{Error analysis.}
Using these annotations, we analyze the agreement between GPT predictions and human labels, as shown in~\figref{llava_onevision_combined}. \figref{llava_onevision_combined} (a) shows that GPT overestimates the proportion of direction-dependent QA pairs. \figref{llava_onevision_combined} (b) further shows that this discrepancy is primarily driven by false positives. Specifically, the classifier incorrectly labels non-directional QA pairs as directional, while missing very few true directional cases. This results in high recall (87.5\%) but low precision (19.1\%), indicating that the classifier rarely misses truly directional QA pairs while substantially over-predicting directional ones. Since the analysis is based on sampled data, we account for sampling variability by computing a 95\% confidence interval ($9.17\% \pm 2.8\%$). Even under this uncertainty, the estimated proportion remains substantially higher than the human-annotated rate, indicating consistent overestimation. Because the classifier tends to include non-directional cases rather than exclude true directional ones, we interpret this estimate as an upper bound on the true amount of motion direction supervision.

\subsubsection{Results}
\label{appen:videollms_instruction_dataset_anal_result}
We quantify the presence of motion direction supervision in video instruction-tuning datasets using the proposed two-stage pipeline. As shown in \tabref{data_analysis}, keyword-based pre-filtering retains a substantial portion of QA pairs (23--29\%), but this proportion drops sharply after semantic filtering. Across datasets, only 0.77--1.89\% of QA pairs require motion direction understanding, remaining consistently around 1\% despite differences in dataset size and composition. Given that GPT-based semantic filtering tends to overestimate due to false positives (\apref{gpt_reliable}), these values should be interpreted as upper bounds. Even under this conservative interpretation, motion direction supervision remains limited across all datasets.



\subsection{Input-side Scaffolds: Design and Full Results}
\label{appen:input_scaffolds}
In this section, we provide a detailed description of the prompting design and full results underlying this finding.  We study input-side scaffolds from two perspectives. On the visual side, we augment video frames with explicit directional cues. On the text side, we modify the prompt to guide step-by-step reasoning about object motion. We consider multiple visual and text scaffolds and evaluate all visual–text combinations across synthetic and real-world benchmarks. We describe the visual and text prompting strategies and then report the full results for all scaffold configurations.


\subsubsection{Visual Prompting}
\label{appen:input_scaffolds_visual_prompting}
\begin{figure}[H]
    \centering

    \begin{subfigure}{0.32\linewidth}
        \centering
        \includegraphics[width=\linewidth]{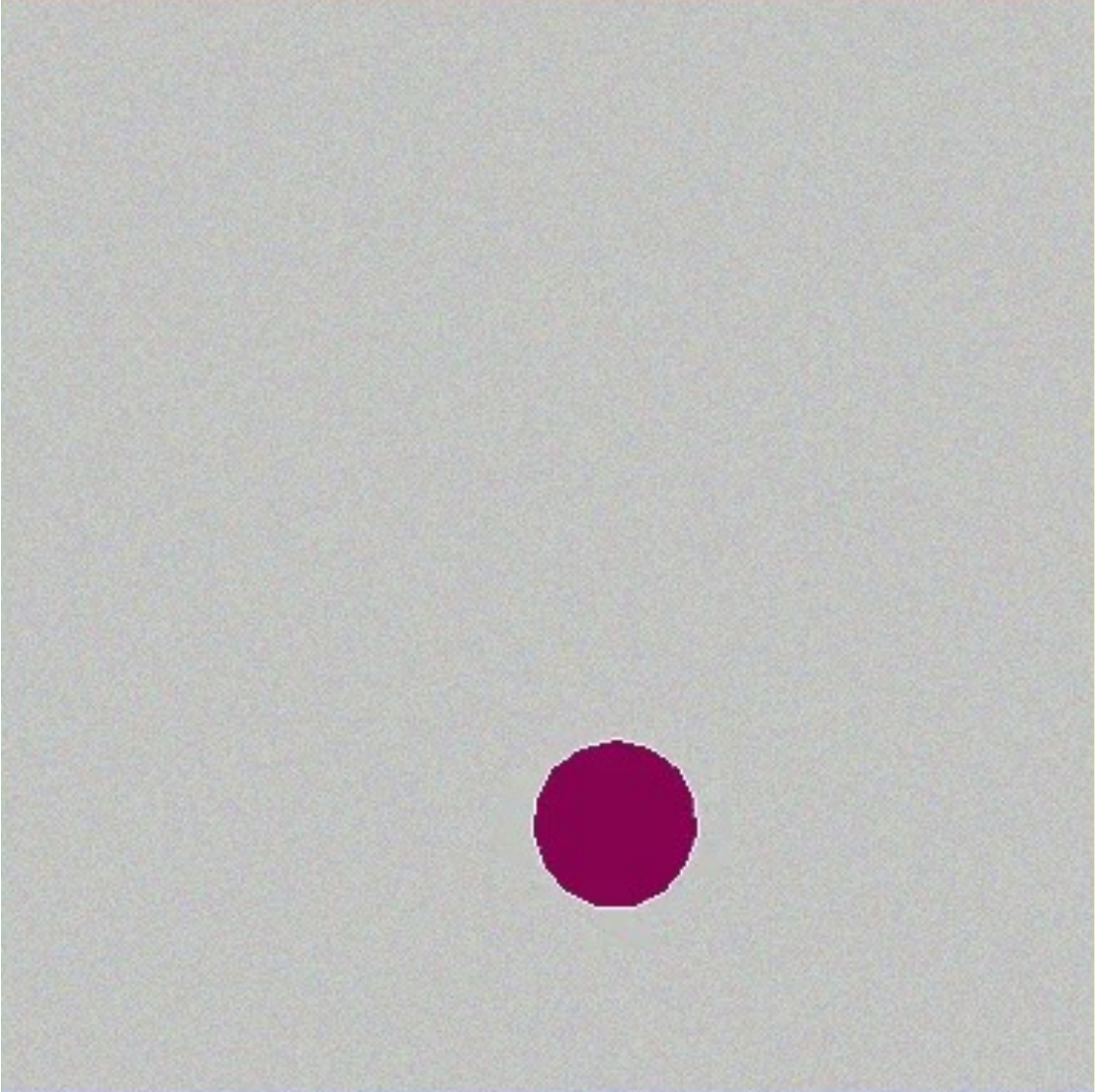}
        \caption{Plain}
        \label{fig:visual_prompting_plain}
    \end{subfigure}
    \hfill
    \begin{subfigure}{0.32\linewidth}
        \centering
        \includegraphics[width=\linewidth]{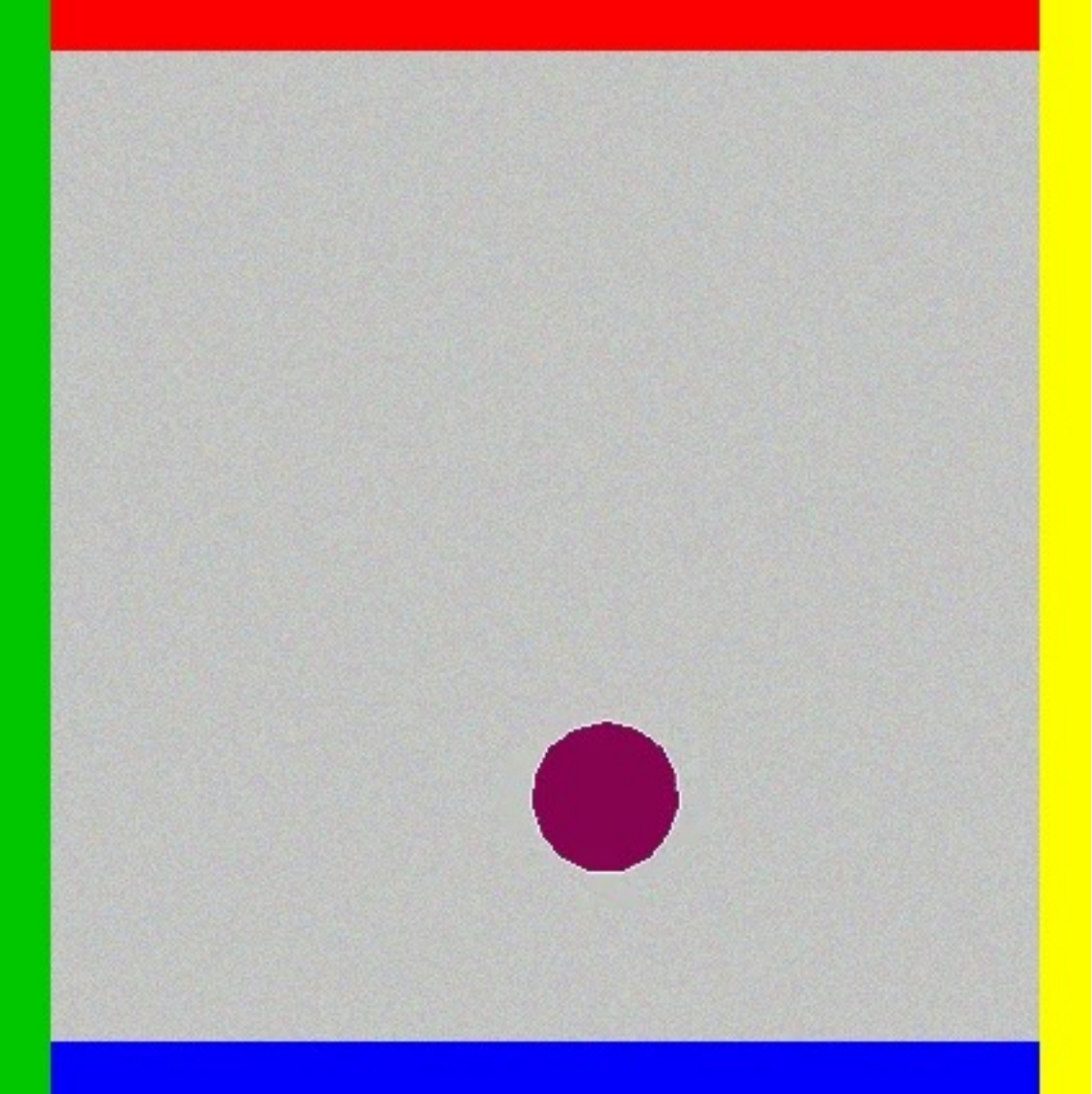}
        \caption{Colored Edge}
        \label{fig:visual_prompting_color}
    \end{subfigure}
    \hfill
    \begin{subfigure}{0.32\linewidth}
        \centering
        \includegraphics[width=1.03\linewidth]{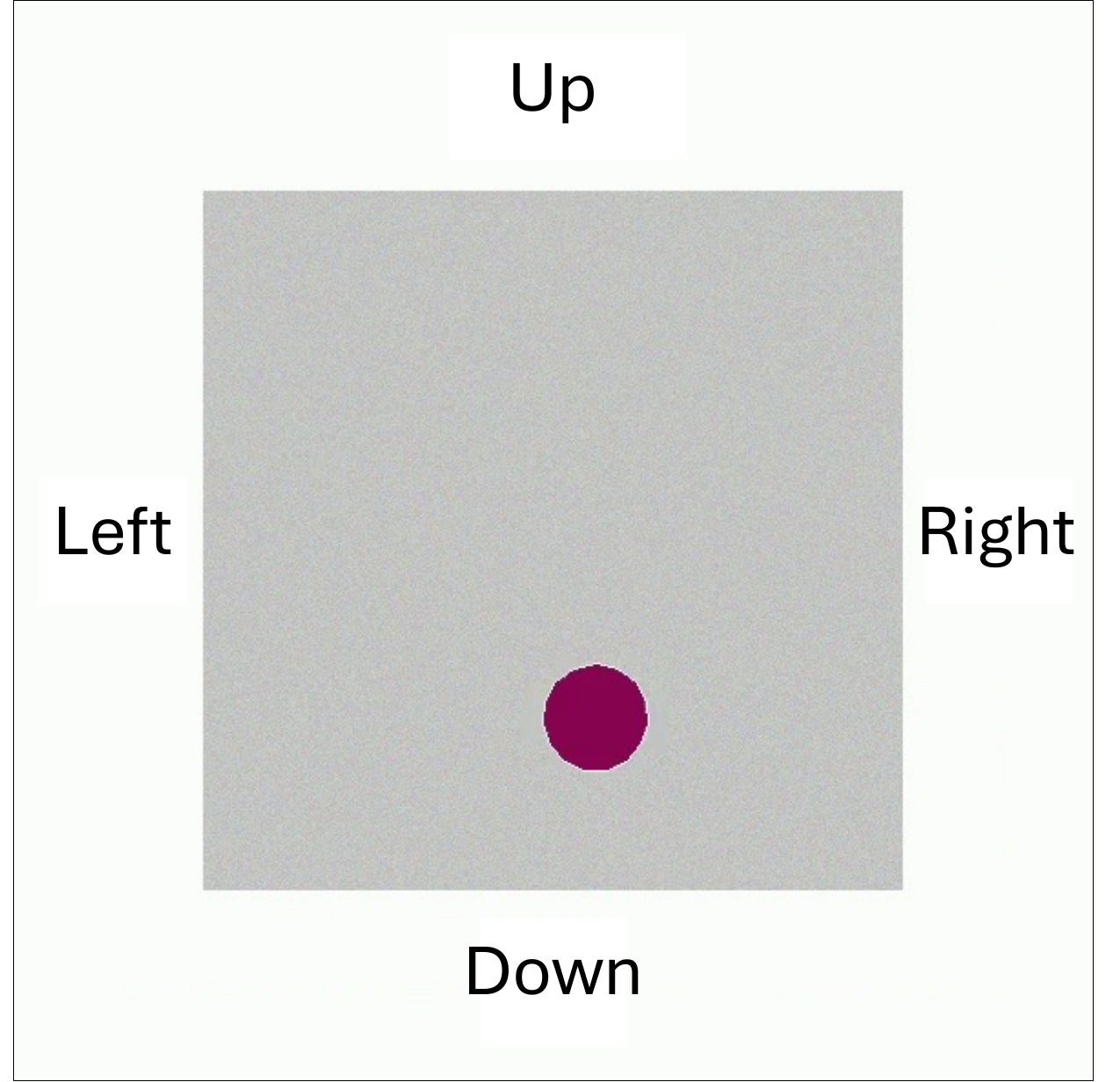}
        \caption{Text Edge}
        \label{fig:visual_prompting_text}
    \end{subfigure}

    \caption{\textbf{Visual prompting examples.}
    We compare the original input with two visual cue variants designed to make directional information more explicit.
    (a) \textit{Plain} shows the unmodified video frame without any additional visual cue.
    (b) \textit{Colored Edge} marks the four image borders with distinct colors.
    (c) \textit{Text Edge} annotates the borders with directional words.}
    \label{fig:visual_prompting}
\end{figure}

Visual prompting introduces explicit directional cues into the video frames, providing additional spatial signals that can guide direction inference~\cite{internvl25, fu2024ocrbenchv2improvedbenchmark}.
We consider three visual conditions:
\emph{(i) Plain} uses the original video without any additional visual cues.
\emph{(ii) Color Edges} assign distinct colors to the four image borders, allowing the model to relate the object's position to the corresponding boundary cues. 
\emph{(iii) Text Edges} explicitly annotate each border with directional words (e.g., ``Up'', ``Down'', ``Left'', ``Right''), providing a more explicit association between position and direction. 
Examples of all conditions are shown in \figref{visual_prompting}.

\subsubsection{Text Prompt}
\label{appen:input_scaffolds_text_prompting}
Text prompting modifies the input question to elicit structured reasoning about motion direction. 
Rather than determining direction from a single query, these prompts guide the model to explicitly reason over object positions and their temporal changes.

\begin{figure}[H]
    \centering
    \includegraphics[width=0.7\linewidth]{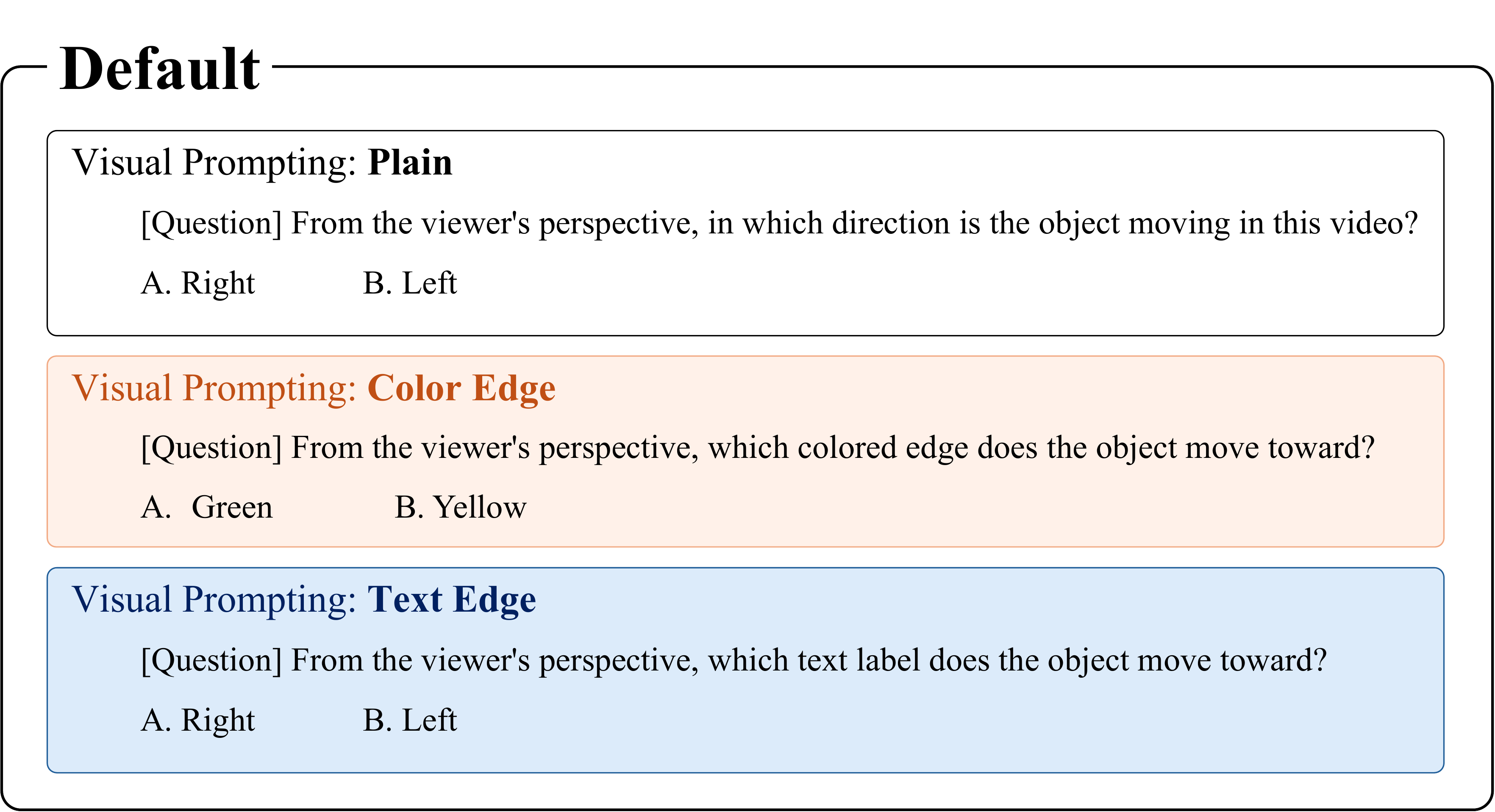}
    \caption{\textbf{Default Prompting Example}}
    \label{fig:default_prompt}
\end{figure}

\begin{figure}[H]
    \centering
    \includegraphics[width=0.7\linewidth]{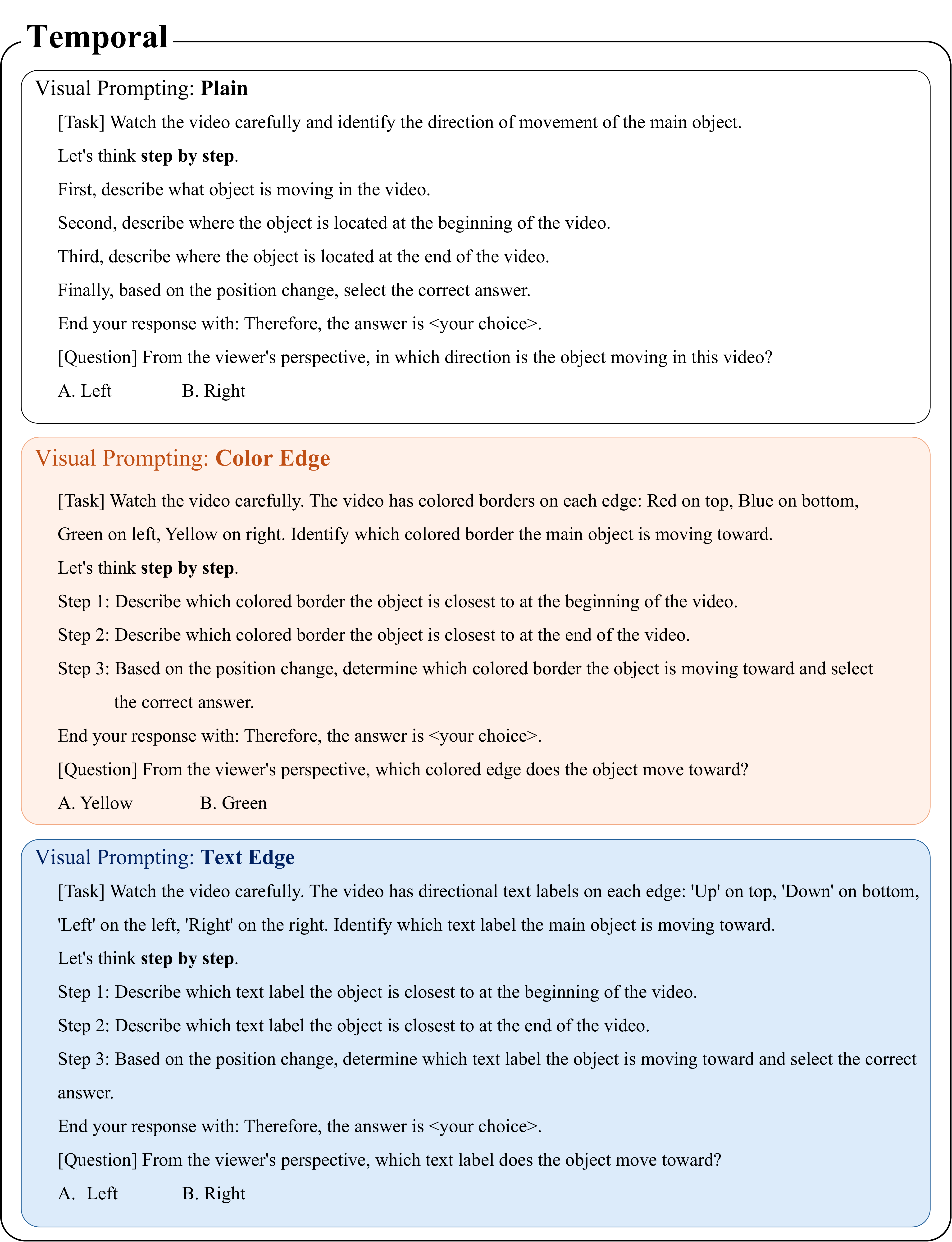}
    \caption{\textbf{Temporal Prompting Example.}}
    \label{fig:step_by_step_prompt}
\end{figure}

\begin{figure}[H]
    \centering
    \includegraphics[width=0.7\linewidth]{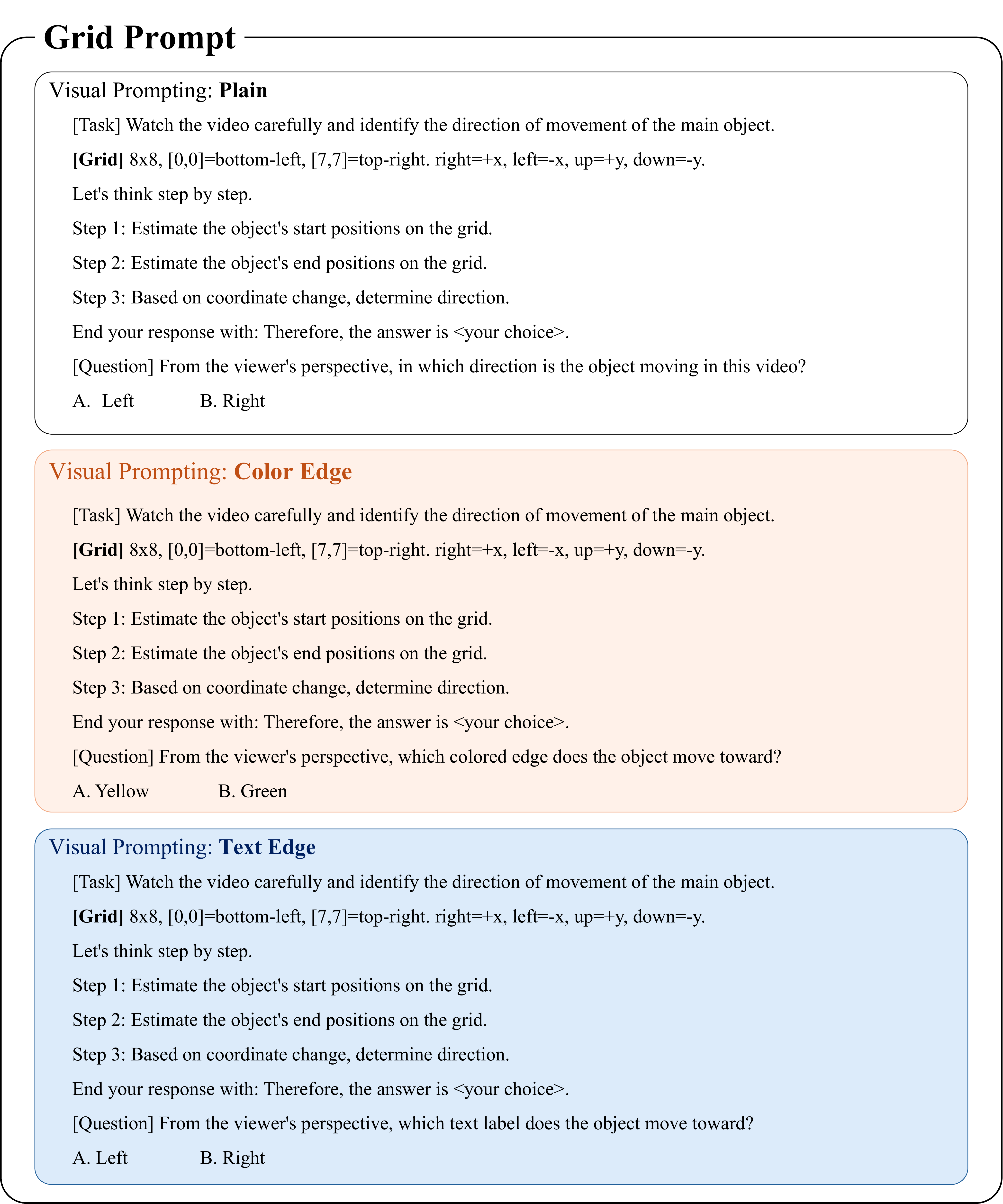}
    \caption{\textbf{Grid Prompting Example.}}
    \label{fig:grid_prompt}
\end{figure}

We consider three variants:
\emph{(i) Default prompting} asks the direction question directly without additional guidance. 
\emph{(ii) Temporal prompting} decomposes the task into intermediate steps, asking the model to describe the object and its positions at the beginning and end of the video before determining direction. 
\emph{(iii) Grid-based prompting} follows the same step-by-step structure as temporal prompting, but additionally introduces an explicit coordinate system over the image, requiring the model to estimate start and end positions within the grid and determine direction from coordinate change. Examples of all conditions are shown in \figref{default_prompt}, \figref{step_by_step_prompt}, \figref{grid_prompt}.

\subsubsection{Full Results}




\begin{table}[h]
\centering
\caption{\textbf{Prompting does not reliably recover motion direction reasoning.} Accuracy under combinations of visual and text prompting.}
\label{tab:prompting_results}
\vspace{0.5em}
\small
\setlength{\tabcolsep}{6pt}
\begin{tabular}{llcc}
\toprule
Visual prompt & Text prompt & \psyn{} & SSv2~\cite{ssv2} \\
\midrule
\multicolumn{2}{l}{Random chance} & 25.0 & 50.0 \\
\midrule
\multirow{3}{*}{Plain}
 & Default       & 27.6 & 52.2 \\
 & Temporal  & 29.0 & \textbf{57.1} \\
 & Grid      & 28.2 & 53.6 \\
\midrule
\multirow{3}{*}{Colored edge}
 & Default       & 25.5 & 48.6 \\
 & Temporal  & 29.4 & 52.9 \\
 & Grid      & 23.0 & 49.2 \\
\midrule
\multirow{3}{*}{Text edge}
 & Default       & 32.0 & 49.9 \\
 & Temporal  & \textbf{34.7} & 54.2 \\
 & Grid      & 29.1 & 50.1 \\
\bottomrule
\end{tabular}
\end{table}

\tabref{prompting_results} reports accuracy under all scaffold combinations on \ps{} and SSv2~\cite{ssv2}. On \ps{}, the best-performing configuration, \textit{Temporal + Text edge}, reaches $34.7\%$, only a modest improvement over the raw baseline ($27.6\%$). 

On SSv2, the best result is achieved by \textit{Temporal prompting} at $57.1\%$, but this gain is not consistently preserved across other combinations. Visual prompting alone (e.g., Colored edge: $48.6\%$, Text edge: $49.9\%$) generally underperforms the raw baseline ($52.2\%$), and combining visual and text prompts does not yield reliable improvements.

Overall, the results show that neither visual cues nor structured text prompting consistently recovers motion direction reasoning, and in many cases, additional prompting introduces noise that degrades performance.

\subsection{Motion Direction Binding Gap Beyond MCQ Format}
\label{appen:gap_beyond_mcq}

In this section, we verify that the direction binding gap is not an artifact of the letter-MCQ format.
We evaluate vanilla LLaVA-Video~\citep{llava_video} on \ps{} under two additional answer formats that preserve the direction question but change the required output: \emph{direction-word MCQ}, where the four options are the direction words \{left, right, up, down\} rather than the letters A/B/C/D, and \emph{open-ended generation}, where the model generates a free-text answer that we score with GPT-4o~\citep{gpt4o}.
For each format, we measure the final-readout direction probe accuracy and the corresponding task accuracy.

\begin{table}[h]
\centering
\caption{\justifying \textbf{The direction binding gap persists beyond the MCQ format.}
Final-readout direction probe accuracy and task accuracy on vanilla LLaVA-Video, evaluated on \ps{} under three answer formats: letter MCQ (default), direction-word MCQ, and open-ended generation.
The final-readout direction probe stays above $90\%$ in all three formats, while task accuracy remains low, leaving a binding gap of more than $60$\,pp in every format.
Open-ended generation is scored with GPT-4o.}
\label{appentab:gap_beyond_mcq}
\setlength{\tabcolsep}{8pt}
\renewcommand{\arraystretch}{1.15}
\begin{tabular}{lccc}
\toprule
Answer format & Probe (\%) & Acc.\ (\%) & Gap (pp) \\
\midrule
Letter MCQ             & $95.3$ & $27.6$ & $67.7$ \\
Direction-word MCQ     & $92.3$ & $27.8$ & $64.5$ \\
Open-ended generation  & $93.3$ & $19.7$ & $73.6$ \\
\bottomrule
\end{tabular}
\end{table}

\paragraph{The binding gap persists across answer formats.}
\apreftab{gap_beyond_mcq} reports the results.
The final-readout direction probe stays above $90\%$ in all three formats, indicating that motion direction is linearly accessible regardless of the required output format.
Yet the task accuracy remains low across the board, ranging from $19.7\%$ for open-ended generation to $27.8\%$ for direction-word MCQ.
The direction binding gap therefore exceeds $60$\,pp in every format we tested.

\paragraph{The binding gap is not an MCQ artifact.}
The persistence of the gap under both direction-word MCQ and open-ended generation rules out the letter-MCQ format as the source of directional motion blindness.
The bottleneck lies in converting the linearly accessible direction signal into the correct verbal answer, regardless of whether that answer is a letter, a word, or a free-text response.

\subsection{Motion Direction Binding Gap Across Video-LLMs}
\label{appen:binding_gap_across_model}

\paragraph{Setup.}
We test whether the direction binding gap observed in LLaVA-Video~(\secref{readout_binding}) generalizes to other Video-LLMs.
For each model, we measure two quantities on the \ps{} domain of \mds{}:
(i)~direction probing accuracy at the final readout token, following \apref{probing_details}, and
(ii)~MCQ accuracy from greedy generation, parsed as the first emitted letter among \texttt{\{A, B, C, D\}}.
We define the binding gap as the difference between the two, following \secref{readout_binding}.
We evaluate every model on $6{,}000$ balanced samples ($1{,}500$ per direction, four directions).

\paragraph{Models.}
We evaluate nine Video-LLMs spanning different vision encoders, projectors, LLM families (Qwen2 / Qwen2.5 / Qwen3 / Vicuna), and parameter scales ($2$B-$7$B).
We list the full model set with citations in \tabref{cross_model_binding_gap}.

\paragraph{The binding gap is consistent across Video-LLMs.}
\begin{table}[htpb]
\centering
\caption{\justifying \textbf{The direction binding gap is universal across Video-LLMs.}
Direction probing accuracy and MCQ accuracy at the final readout state, evaluated on \ps{} in \mds{} ($6{,}000$ balanced samples).
Direction is linearly decodable in every backbone, yet MCQ accuracy stays near the $25$\,\% chance baseline in seven of the nine models.
Even Qwen3-VL, the strongest MCQ baseline, retains a $31.2$\,pp gap.}
\label{tab:cross_model_binding_gap}
\vspace{0.5em}
\resizebox{0.85\linewidth}{!}{%
\begin{tabular}{lccc}
\toprule
Method & Direction Probe (\%) & MCQ Acc. (\%) & Gap (pp) \\
\midrule
LLaVA-Video-7B (Qwen2)~\cite{llava_video}            & 95.3 & 27.6 & 67.7 \\
mPLUG-Owl3-7B~\cite{mplug_owl3}                    & 90.5 & 25.6 & 64.9 \\
LLaVA-OneVision-7B (Qwen2)~\cite{llava_onevision}    & 90.8 & 28.3 & 62.5 \\
VideoLLaMA3-2B (Qwen2)~\cite{videollama3}            & 93.6 & 31.2 & 62.4 \\
InternVL2.5-4B (Qwen2.5)~\cite{internvl25}           & 91.1 & 31.9 & 59.2 \\
Qwen2.5-VL-7B~\cite{qwen25vl}                        & 95.8 & 45.1 & 50.7 \\
LLaVA-NeXT-Video-7B (Vicuna)~\cite{llava_next_video} & 75.3 & 25.8 & 49.5 \\
Video-LLaVA-7B (Vicuna)~\cite{videollava}            & 77.1 & 28.1 & 49.0 \\
Qwen3-VL-4B~\cite{qwen3vl}                           & \textbf{97.8} & \textbf{66.6} & \textbf{31.2} \\
\bottomrule
\end{tabular}%
}
\end{table}
\tabref{cross_model_binding_gap} reports direction probing and MCQ accuracy on the nine evaluated Video-LLMs.
Direction probing is high in every model ($75.3$\,\%-$97.8$\,\%), confirming that motion direction is linearly decodable from the readout token across architectures.
MCQ accuracy, however, stays near the $25$\,\% chance baseline in seven of the nine models; only Qwen2.5-VL ($45.1$\,\%) and Qwen3-VL ($66.6$\,\%) achieve meaningful direction MCQ out of the box.
The resulting binding gap exceeds $30$ percentage points in every Video-LLM we evaluate.
Even Qwen3-VL — the most MCQ-aligned vanilla model — exhibits a $31.2$\,pp gap (probe $97.8$\,\% vs.\ MCQ $66.6$\,\%): the binding gap persists even when downstream MCQ capability is high.
The gap is therefore not specific to a particular vision encoder, projector, or LLM family, but reflects a shared failure mode across the Video-LLMs we evaluate.
\newpage

\section{Additional Experimental Results}
\label{appen:additional_exp}
%
%

\subsection{Ablation Study of \ours{}}
\label{appen:ablation}

This section provides ablation studies for \ours{}.
Our goal is to verify that the improvement does not come from additional training alone, but from the proposed motion direction supervision and its feature construction.
We therefore compare alternative variants that modify the auxiliary objective, the motion feature used for supervision, and the training configuration while keeping the remaining components fixed.
%

\subsubsection{Setup}
\label{appen:ablation_setup}

We vary one design axis at a time around an anchor configuration and report the impact on direction performance.
The anchor configuration follows Section~\ref{sec:deltadirect}: LLaVA-Video (Qwen2~\cite{qwen2} LLM, SigLIP~\cite{siglip} vision encoder), $T=8$ frames, LoRA rank $r=64$ with $\alpha=128$ on the LLM, a fully fine-tuned projector, $\lambda_{\mathrm{MVP}}=1.0$, mean spatial pooling, a linear motion head, the subtraction temporal operator, and MSE loss against unit-vector targets.

\paragraph{Evaluation suite.}
We report seven direction tasks and one general-video task.
The seven direction tasks are: 4-way directional MCQ on the four \mds{} domains (chance $25\%$, $1{,}500$ samples per direction), and the three \mdr{} sub-tasks KTH, SSv2, and TOMATO.
For compactness, we abbreviate the four \mds{} domains as \psyn{}, \csyn{}, \preal{}, and \creal{}, where P/C denote Primitive/Cutout foregrounds and Syn/Real denote synthetic/real backgrounds.
\emph{Dir avg.}\ denotes the unweighted mean over these seven tasks.
MVBench (mean over 20 sub-tasks) monitors general video understanding.
%

\subsubsection{Supervision Location}
\label{appen:ablation_location}
\begin{table*}[h]
\centering
\caption{\textbf{Ablation: location of MVP supervision (full breakdown).}
During \mdi{} instruction tuning, we apply MVP supervision at different points along the Video-LLM pipeline and report Top-1 accuracy (\%) on the four \mds{} domains (P-Syn, C-Syn, P-Real, C-Real; chance $25\%$) and three \mdr{} sub-tasks (SSv2, KTH, TOMATO).
Avg.\ Direction is the unweighted mean over \mds{} and \mdr{}; MVBench~\cite{mvbench} monitors general video understanding.
$\ell_v$ and $\ell$ denote the layer index of the vision encoder~\cite{siglip} ($L_v=27$) and LLM~\cite{qwen2} ($L=28$), respectively.}
\label{appentab:ablation_supervision_location}
\setlength{\tabcolsep}{4pt}
\renewcommand{\arraystretch}{1.15}
\footnotesize
\resizebox{\textwidth}{!}{%
\begin{tabular}{ll cccc ccc c c}
\toprule
\multirow{2}{*}{Stage}
& \multirow{2}{*}{Feat.}
& \multicolumn{4}{c}{\mds{}}
& \multicolumn{3}{c}{\mdr{}}
& \multirow{2}{*}{\makecell{Avg.\\ Direction}}
& \multirow{2}{*}{MVBench~\cite{mvbench}} \\
\cmidrule(lr){3-6} \cmidrule(lr){7-9}
&
& P-Syn & C-Syn & P-Real & C-Real
& SSv2~\cite{ssv2} & KTH~\cite{kth} & TOMATO~\cite{tomato}
& & \\
\midrule
None & --
& 99.7 & 81.1 & 75.3 & 61.0 & 72.0 & 67.9 & 35.8
& 70.4 & 60.8 \\
\midrule
\multirow{3}{*}{Vision encoder}
& $\mathbf{V}^{\ell_v=7}$
& 99.8 & 81.6 & 76.6 & 62.3 & 72.6 & 66.7 & 35.5
& 70.7 & 60.6 \\
& $\mathbf{V}^{\ell_v=14}$
& 99.8 & 82.3 & 76.7 & 62.7 & 73.4 & 69.2 & 35.2
& 71.3 & 60.9 \\
& $\mathbf{V}^{\ell_v=21}$
& 99.7 & 81.6 & 76.3 & 62.1 & 73.1 & 67.3 & 35.5
& 70.8 & 60.7 \\
\addlinespace[2pt]
Pre-projector & $\mathbf{V}^{L_v}$
& 99.8 & 86.2 & 86.9 & 71.4 & 78.1 & 71.6 & 37.0
& 75.9 & 60.2 \\
\addlinespace[2pt]
\textbf{Post-projector (Ours)} & $\mathbf{F}$
& \textbf{99.6} & \textbf{85.5} & \textbf{86.1} & \textbf{72.5}
& \textbf{80.7} & \textbf{74.6} & \textbf{38.1}
& \textbf{76.7} & \textbf{60.8} \\
\addlinespace[2pt]
\multirow{3}{*}{LLM visual tokens}
& $\mathbf{z}^{\ell=7}$
& 27.2 & 27.1 & 27.6 & 26.9 & 50.6 & 42.0 & 20.6
& 31.7 & 56.6 \\
& $\mathbf{z}^{\ell=14}$
& 83.8 & 72.9 & 63.1 & 60.0 & 73.8 & 68.6 & 32.5
& 65.0 & 57.4 \\
& $\mathbf{z}^{\ell=21}$
& 79.0 & 62.1 & 63.6 & 51.5 & 66.9 & 71.4 & 26.6
& 60.2 & 57.5 \\
\addlinespace[2pt]
Final readout & $\mathbf{h}^{L}$
& 28.5 & 26.7 & 30.1 & 25.4 & 56.8 & 50.2 & 20.5
& 34.0 & 59.9 \\
\midrule
\multirow{2}{*}{Two-tap}
& $\mathbf{F}, \mathbf{z}^{L}$
& 63.6 & 38.0 & 53.3 & 41.3 & 60.8 & 55.1 & 27.3
& 48.5 & 60.3 \\
& $\mathbf{F}, \mathbf{z}^{\ell=4}$
& 45.2 & 32.2 & 38.3 & 32.8 & 54.8 & 54.7 & 24.1
& 40.3 & 57.6 \\
\bottomrule
\end{tabular}%
}
\end{table*}

%
\paragraph{Post-projector supervision yields the largest gain.}
We compare supervision at the projector output (anchor) against four alternatives: (i) vision-encoder intermediate layers (SigLIP blocks 7, 14, 21), (ii) the pre-projector vision-encoder output, (iii) LLM hidden states (layers 7, 14, 21) and the final readout token, and (iv) two-tap combinations of pre- and post-LLM supervision.
\apreftab{ablation_supervision_location} shows that supervision is effective only before the LLM, and that the projector output is the best point: \ours{} improves Dir avg.\ over the no-auxiliary baseline by $+6.3$\,pp ($70.4 \to 76.7$) while leaving MVBench unchanged ($60.8$).
Pre-projector supervision is the runner-up ($+5.5$\,pp), but it slightly degrades MVBench ($60.2$ vs.\ $60.8$).

Vision-encoder taps leave the downstream representation almost unchanged ($+0.3$ to $+0.9$\,pp).
SigLIP already encodes motion direction densely, and supervising it adds gradient that does not propagate to the LLM-side readout.
LLM-side taps fail in a different way: early-layer supervision (L7) collapses MCQ formatting and pushes \mds{} accuracies to chance, while mid- and late-layer supervision (L14, L21) and readout supervision underperform even the no-auxiliary baseline.
Two-tap combinations inherit this degradation.
Direct supervision of post-LLM hidden states interferes with the autoregressive readout that produces the answer letter.

\subsubsection{Direction Loss}
\label{appen:ablation_loss}
\begin{table}[h]
\centering
\caption{\textbf{Ablation: direction loss.} 
We compare MSE on $(\cos\theta, \sin\theta)$ (anchor) with a cosine loss. 
The cosine loss is $4.3$\,pp lower on Avg.\ Direction and comparable on MVBench.}
\label{tab:ablation_loss}
\setlength{\tabcolsep}{4pt}
\renewcommand{\arraystretch}{1.15}
\resizebox{\columnwidth}{!}{%
\begin{tabular}{l cccc ccc c c}
\toprule
\multirow{2}{*}{Loss}
& \multicolumn{4}{c}{\mds{}}
& \multicolumn{3}{c}{\mdr{}}
& \multirow{2}{*}{\makecell{Avg.\\ Direction}}
& \multirow{2}{*}{MVBench~\cite{mvbench}} \\
\cmidrule(lr){2-5} \cmidrule(lr){6-8}
& P-Syn & C-Syn & P-Real & C-Real
& SSv2~\cite{ssv2} & KTH~\cite{kth} & TOMATO~\cite{tomato}
& & \\
\midrule
\textbf{MSE (anchor)}
& 99.6 & \textbf{85.5} & \textbf{86.1} & \textbf{72.5}
& \textbf{80.7} & \textbf{74.6} & \textbf{38.1}
& \textbf{76.7} & \textbf{60.8} \\
Cosine
& \textbf{99.7} & 84.0 & 78.2 & 64.5
& 75.9 & 69.3 & 35.5
& 72.4 & 60.4 \\
\bottomrule
\end{tabular}%
}
\end{table}

\paragraph{MSE on the unit-vector target works best.}
We compare the anchor loss, MSE on $(\cos\theta, \sin\theta)$, against a cosine loss $1-\cos(\hat{\mathbf{m}}_t, \mathbf{m}_t)$.
As shown in \tabref{ablation_loss}, MSE outperforms the cosine loss by $4.3$\,pp on Avg.\ Direction ($76.7$ vs.\ $72.4$) and is slightly better on MVBench ($60.8$ vs.\ $60.4$).
Cosine loss is invariant to the magnitude of the prediction, whereas MSE on a unit-norm target additionally pulls the predicted vector toward unit length, which provides a tighter supervision signal.
We therefore use MSE as the default direction loss.

\subsubsection{Frame-Pair Operator}
\label{appen:ablation_temporal_op}
\begin{table}[h]
\centering
\caption{\textbf{Ablation: frame-pair operator.} 
We compare subtraction with a learned concatenation operator. 
Subtraction outperforms concatenation on every direction task, while MVBench is comparable.}
\label{tab:ablation_temporal_op}
\setlength{\tabcolsep}{4pt}
\renewcommand{\arraystretch}{1.15}
\resizebox{\columnwidth}{!}{%
\begin{tabular}{l cccc ccc c c}
\toprule
\multirow{2}{*}{Operator}
& \multicolumn{4}{c}{\mds{}}
& \multicolumn{3}{c}{\mdr{}}
& \multirow{2}{*}{\makecell{Avg.\\ Direction}}
& \multirow{2}{*}{MVBench~\cite{mvbench}} \\
\cmidrule(lr){2-5} \cmidrule(lr){6-8}
& P-Syn & C-Syn & P-Real & C-Real
& SSv2~\cite{ssv2} & KTH~\cite{kth} & TOMATO~\cite{tomato}
& & \\
\midrule
\textbf{Subtraction (anchor)}
& \textbf{99.6} & \textbf{85.5} & \textbf{86.1} & \textbf{72.5}
& \textbf{80.7} & \textbf{74.6} & \textbf{38.1}
& \textbf{76.7} & \textbf{60.8} \\
Concatenation
& 99.5 & 79.4 & 76.4 & 60.6
& 74.9 & 68.9 & 36.2
& 70.8 & 60.6 \\
\bottomrule
\end{tabular}%
}
\end{table}

\paragraph{The subtraction prior is useful.}
We compare the anchor operator, subtraction ($\mathbf{F}_{t+1}-\mathbf{F}_t$), with a learned concatenation operator ($\mathrm{Linear}([\mathbf{F}_t; \mathbf{F}_{t+1}])$).
As shown in \tabref{ablation_temporal_op}, concatenation underperforms subtraction by $5.9$\,pp on Avg.\ Direction ($70.8$ vs.\ $76.7$), while MVBench is comparable ($60.6$ vs.\ $60.8$).
The linear layer over the concatenated features can in principle represent subtraction, so this gap suggests that the model does not recover the frame difference on its own under the limited supervision of \mdi{}.
%
We therefore adopt subtraction as the default operator, which also adds no parameters.

\subsubsection{Loss Weight}
\label{appen:ablation_lambda}
\begin{table}[h]
\centering
\caption{\textbf{Ablation: loss weight $\lambda_{\mathrm{MVP}}$.} 
The total objective is $L = L_{\mathrm{LM}} + \lambda_{\mathrm{MVP}} \cdot L_{\mathrm{MVP}}$. 
Avg.\ Direction increases monotonically with $\lambda_{\mathrm{MVP}}$ in the explored range, while MVBench is comparable.}
\label{tab:ablation_lambda}
\setlength{\tabcolsep}{4pt}
\renewcommand{\arraystretch}{1.15}
\resizebox{\columnwidth}{!}{%
\begin{tabular}{c cccc ccc c c}
\toprule
\multirow{2}{*}{$\lambda_{\mathrm{MVP}}$}
& \multicolumn{4}{c}{\mds{}}
& \multicolumn{3}{c}{\mdr{}}
& \multirow{2}{*}{\makecell{Avg.\\ Direction}}
& \multirow{2}{*}{MVBench~\cite{mvbench}} \\
\cmidrule(lr){2-5} \cmidrule(lr){6-8}
& P-Syn & C-Syn & P-Real & C-Real
& SSv2~\cite{ssv2} & KTH~\cite{kth} & TOMATO~\cite{tomato}
& & \\
\midrule
$0.1$
& \textbf{99.8} & 82.5 & 76.3 & 62.2
& 72.8 & 68.6 & 35.5
& 71.1 & 60.7 \\
$0.5$
& \textbf{99.8} & 84.5 & 82.4 & 67.4
& 76.6 & 67.3 & 36.5
& 73.5 & 60.4 \\
$\mathbf{1.0}$ \textbf{(anchor)}
& 99.6 & \textbf{85.5} & \textbf{86.1} & \textbf{72.5}
& \textbf{80.7} & \textbf{74.6} & \textbf{38.1}
& \textbf{76.7} & \textbf{60.8} \\
\bottomrule
\end{tabular}%
}
\end{table}

\paragraph{A larger $\lambda_{\mathrm{MVP}}$ improves direction performance without hurting MVBench.}
The total training objective is $L = L_{\mathrm{LM}} + \lambda_{\mathrm{MVP}} \cdot L_{\mathrm{MVP}}$, and we sweep $\lambda_{\mathrm{MVP}} \in \{0.1, 0.5, 1.0\}$.
As shown in \tabref{ablation_lambda}, Avg.\ Direction increases monotonically with $\lambda_{\mathrm{MVP}}$ ($71.1 \to 73.5 \to 76.7$).
MVBench varies by at most $0.4$\,pp across the sweep ($60.4$--$60.8$), suggesting that $\lambda_{\mathrm{MVP}}$ strengthens the direction-specific signal without degrading general video understanding.
We use $\lambda_{\mathrm{MVP}}=1.0$ as the default.
Since Avg.\ Direction does not saturate within this range, larger values may yield further gains, which we leave for future work.

\subsubsection{LoRA Rank}
\label{appen:ablation_rank}
\begin{table}[h]
\centering
\caption{\textbf{Ablation: LoRA rank.} 
LoRA rank $r$ with $\alpha = 2r$ on the LLM; the projector is fully fine-tuned. 
Direction performance is monotone in rank; MVBench is essentially flat across the range.}
\label{tab:ablation_rank}
\setlength{\tabcolsep}{4pt}
\renewcommand{\arraystretch}{1.15}
\resizebox{\columnwidth}{!}{%
\begin{tabular}{c cccc ccc c c}
\toprule
\multirow{2}{*}{Rank $r$}
& \multicolumn{4}{c}{\mds{}}
& \multicolumn{3}{c}{\mdr{}}
& \multirow{2}{*}{\makecell{Avg.\\ Direction}}
& \multirow{2}{*}{MVBench~\cite{mvbench}} \\
\cmidrule(lr){2-5} \cmidrule(lr){6-8}
& P-Syn & C-Syn & P-Real & C-Real
& SSv2~\cite{ssv2} & KTH~\cite{kth} & TOMATO~\cite{tomato}
& & \\
\midrule
$16$
& 98.0 & 70.8 & 73.5 & 57.5
& 67.3 & 58.1 & 34.0
& 65.6 & 60.2 \\
$32$
& 99.3 & 80.1 & 80.7 & 63.4
& 72.2 & 64.4 & 34.5
& 70.7 & 60.5 \\
$\mathbf{64}$ \textbf{(anchor)}
& \textbf{99.6} & \textbf{85.5} & \textbf{86.1} & \textbf{72.5}
& \textbf{80.7} & \textbf{74.6} & \textbf{38.1}
& \textbf{76.7} & \textbf{60.8} \\
\bottomrule
\end{tabular}%
}
\end{table}

\paragraph{Direction performance is monotone in rank; $r=64$ is on the budget knee.}
We apply LoRA with rank $r$ and $\alpha = 2r$ to the LLM, with $r \in \{16, 32, 64\}$, while the projector is fully fine-tuned in all settings.
$r=16$ already learns the source domain task (\ps{} $98.0\%$) but degrades on harder splits and on real-world directional tasks.
MVBench is essentially flat across the range ($60.2$-$60.8$), so the cost of higher rank is borne entirely by the direction signal.
We adopt $r=64$ as the anchor.

\subsubsection{Number of Frames}
\label{appen:ablation_frames}
\begin{table}[h]
\centering
\caption{\textbf{Ablation: frame count.}
Frames per video $T$ at fixed effective batch size ($144$).
$T=8$ is the optimal frame budget for both baseline and \ours{}; $T=16$ and $T=32$ both under-perform.
The relative gain of \ours{} over baseline is preserved at $T=32$.
\emph{Avg.\ Direction} is the unweighted mean over the seven direction tasks.}
\label{tab:ablation_frames}
\setlength{\tabcolsep}{4pt}
\renewcommand{\arraystretch}{1.15}
\resizebox{\columnwidth}{!}{%
\begin{tabular}{ll cccc ccc c c}
\toprule
\multirow{2}{*}{Method}
& \multirow{2}{*}{$T$}
& \multicolumn{4}{c}{\mds{}}
& \multicolumn{3}{c}{\mdr{}}
& \multirow{2}{*}{\makecell{Avg.\\ Direction}}
& \multirow{2}{*}{MVBench~\cite{mvbench}} \\
\cmidrule(lr){3-6} \cmidrule(lr){7-9}
& 
& P-Syn & C-Syn & P-Real & C-Real
& SSv2~\cite{ssv2} & KTH~\cite{kth} & TOMATO~\cite{tomato}
& & \\
\midrule
\multirow{2}{*}{Baseline}
& $8$
& \textbf{99.7} & 81.1 & 75.3 & 61.0 & 72.0 & 67.9 & 35.8
& 70.4 & \textbf{60.8} \\
& $32$
& 96.6 & 71.1 & 63.3 & 51.2
& 67.3 & 59.8 & 27.8
& 62.4 & 60.1 \\
\midrule
\multirow{3}{*}{\textbf{\ours{}}}
& $\mathbf{8}$ \textbf{(anchor)}
& 99.6 & \textbf{85.5} & \textbf{86.1} & \textbf{72.5}
& \textbf{80.7} & \textbf{74.6} & \textbf{38.1}
& \textbf{76.7} & \textbf{60.8} \\
& $16$
& 96.9 & 79.8 & 82.1 & 68.2
& 75.9 & 66.3 & 31.0
& 71.5 & 59.6 \\
& $32$
& 96.4 & 77.6 & 82.3 & 68.1
& 76.5 & 69.6 & 28.5
& 71.3 & 59.4 \\
\bottomrule
\end{tabular}%
}
\end{table}

\paragraph{Eight input frames are sufficient for learning motion direction.}
We compare $T \in \{8, 16, 32\}$ frames per video at fixed effective batch size ($144$).
$T=8$ (anchor) is best on Dir avg.\ ($76.7$); $T=16$ drops to $71.5$ and $T=32$ to $71.3$.
The relative gain of \ours{} over baseline is preserved at $T=32$ ($+8.9$\,pp; baseline drops by $8.0$\,pp from $T=8$ to $T=32$).
The synthetic clips contain a single coherent motion that $8$ frames already capture, so extra frames add gradient noise (smaller per-step batch, more padding) without adding new content.
%



\subsubsection{Supervision Target}
In ~\tabref{ablation_target} of the main paper, we compare \ours{} against three alternative auxiliary objectives.
%
\paragraph{Frame order.}
We attach a small classification head to LLM hidden states at layer 16 that predicts each frame's temporal position via cross-entropy with target $[0, 1, \dots, T{-}1]$.
This post-LLM objective pressures the language model to maintain temporally discriminative representations, without requiring any direction labels.

\paragraph{Concatenating feature deltas.}
We compute per-pair frame deltas $\Delta_t = F_{t+1} - F_t$ from the projector output, average them over the spatial dimension, and concatenate the resulting $T{-}1$ motion tokens to the LLM input sequence.
No auxiliary loss or learnable parameter is introduced; the LLM must learn to exploit the appended motion tokens through the standard language modeling objective alone.

\paragraph{Delta equivariance.}
We collect all spatially averaged frame deltas across the batch from the projector output and compute their pairwise cosine similarity matrix.
An MSE loss aligns this representation similarity matrix with the corresponding direction ground-truth similarity matrix.
This regularizer constrains relative structure rather than regressing direction directly, and introduces no additional learnable parameters.
\paragraph{Results.}\ours{} outperforms all alternatives, and the direction-agnostic objectives provide limited gains, confirming that explicit direction supervision on the signed 2-D motion vector is necessary.

\subsection{\ours{} Across Video-LLM Backbones}
\label{appen:ablation_backbone}
To verify that \ours{} generalizes beyond our main backbone, we apply it to additional Video-LLMs with minimal architecture-specific changes. We keep the loss formulation, head design, and training recipe identical across backbones, adjusting only the layer from which the head reads visual features.
\begin{table*}[h]
\centering
\caption{\textbf{Ablation: backbone robustness.} 
\ours{} replicated across three Video-LLM backbones. 
\ours{} consistently improves Avg.\ Direction across most backbones.}
\label{tab:ablation_backbone}
\setlength{\tabcolsep}{3pt}
\renewcommand{\arraystretch}{1.15}
\resizebox{\textwidth}{!}{%
\begin{tabular}{ll cccc ccc c}
\toprule
\multirow{2}{*}{Backbone}
& \multirow{2}{*}{Variant}
& \multicolumn{4}{c}{\mds{}}
& \multicolumn{3}{c}{\mdr{}}
& \multirow{2}{*}{\makecell{Avg.\\ Direction}} \\
\cmidrule(lr){3-6} \cmidrule(lr){7-9}
&
& P-Syn & C-Syn & P-Real & C-Real
& SSv2~\cite{ssv2} & KTH~\cite{kth} & TOMATO~\cite{tomato}
& \\
\midrule
\multirow{3}{*}{LLaVA-Video-7B
~\cite{llava_video}}
& Vanilla
& 27.6 & 23.4 & 26.9 & 25.8
& 52.2 & 50.3 & 26.8
& 33.3 \\
& \mdi{}
& 99.7$\pm$0.2 & 81.1$\pm$0.4 & 75.3$\pm$0.7 & 61.0$\pm$0.5
& 72.0$\pm$0.6 & 67.9$\pm$1.6 & 35.8$\pm$0.5
& 70.4$\pm$0.5 \\
& \textbf{\ours{}}
& 99.6$\pm$0.1 & 85.5$\pm$0.9 & 86.1$\pm$0.9 & 72.5$\pm$0.9
& 80.7$\pm$0.9 & 74.6$\pm$0.9 & 38.1$\pm$0.7
& 76.7$\pm$0.3 \\
\midrule
\multirow{3}{*}{LLaVA-OneVision-7B~\cite{llava_onevision}}
& Vanilla
& 28.3 & 23.4 & 29.9 & 29.3
& 52.2 & 50.3 & 26.8
& 34.3 \\
& \mdi{}
& 99.8 & 88.2 & 82.2 & 76.9
& 65.0 & 51.6 & 31.3
& 70.7 \\
& \textbf{\ours{}}
& 99.6 & 89.6 & 83.2 & 78.3
& 66.6 & 51.5 & 32.0
& 71.5 \\
\midrule
\multirow{3}{*}{mPLUG-Owl3-7B~\cite{mplug_owl3}}
& Vanilla
& 25.6 & 24.2 & 25.8 & 25.4
& 43.4 & 50.3 & 21.3
& 30.9 \\
& \mdi{}
& 100.0 & 99.9 & 93.6 & 89.5
& 94.4 & 91.3 & 37.2
& 86.6 \\
& \textbf{\ours{}}
& 100.0 & 99.2 & 92.6 & 92.8
& 93.2 & 87.1 & 31.5
& 85.2 \\
\bottomrule
\end{tabular}%
}
\end{table*}

\paragraph{\ours{} improves direction accuracy across all tested backbones.}
We replicate \ours{} on three Video-LLM backbones: LLaVA-Video-7B, LLaVA-OneVision-7B, and mPLUG-Owl3-7B.
\ours{} improves Avg.\ Direction over the corresponding instruction-tuning baseline on every backbone, with consistent gains on LLaVA-Video ($+6.3$\,pp) and LLaVA-OneVision ($+0.8$\,pp).
The smaller gain on LLaVA-OneVision reflects a stronger baseline (Avg.\ Direction $70.7\%$), already close to that of LLaVA-Video.
We adopt LLaVA-Video-7B as the primary backbone for the strongest baseline and the cleanest signal for ablation analysis throughout the paper.

\subsection{Additional Analysis of \ours{}}
\label{appen:add_analysis_delta}

The previous subsection shows that \ours{} substantially improves motion direction accuracy.
We now ask \emph{why}, tracing the effect from the projector output through the LLM using the diagnostic framework from \secref{diagnosis} and \secref{inst_tuning}.

\begin{table}[h]
\centering
\caption{\textbf{\ours{} improves cross-domain direction probing at the projector output.}
We train a direction probe on \ps{} projector outputs and evaluate out of domain.
OOD Avg. is the macro average over \csyn{}, \preal{}, and \creal{}.}
\label{appentab:cross_domain_probing}
\vspace{0.3em}
\small
\setlength{\tabcolsep}{5pt}
\begin{tabular}{lcc}
\toprule
Model & OOD Avg. & \creal{} \\
\midrule
LLaVA-Video-7B & 42.8 & 28.5 \\
LLaVA-Video-7B w/ \mdi{} & 42.2 & 27.5 \\
LLaVA-Video-7B w/ \ours{} & \textbf{68.6} & \textbf{55.4} \\
\bottomrule
\end{tabular}
\end{table}


\vspace{\paramargin}



\paragraph{Domain-invariant direction signals emerge at the projector output.}
To locate where the improvement originates, we probe the projector output $\mathbf{F}$.
Specifically, we train a four-way direction probe on \ps{} and evaluate it on the OOD domains of \mds{} (\apreftab{cross_domain_probing}).
Instruction tuning alone remains close to the vanilla model, with an OOD average of $42.2\%$ and only $27.5\%$ on \cp{}.
With \ours{}, the OOD average increases to $68.6\%$, and \cp{} accuracy rises to $55.4\%$.


\vspace{\paramargin}
\paragraph{A small projector shift amplifies the OOD direction signal.}
\begin{table}[h]
\centering
\caption{\textbf{\ours{} selectively amplifies the direction concept vector magnitude on OOD domains.} 
Direction concept-vector magnitude $\|\mathbf{v}^{\ell=21}_d\|$ at layer 21 under \mdi{} and \ours{}. 
The relative change $\Delta(\%)$ grows from near-zero on the source domain \ps{} to $+20.4\%$ on the most OOD \cp{}.}
\label{appentab:delta_direct_magnitude_amplification}
\small
\setlength{\tabcolsep}{8pt}
\renewcommand{\arraystretch}{1.15}
\begin{tabular}{l ccc}
\toprule
Domain & \mdi{} & \ours{} & $\Delta(\%)$ \\
\midrule
\ps{} (Source domain) & 28.92 & 28.29 & $-2.2$ \\
\cs{}       & 20.49 & 20.85 & $+1.8$ \\
\pp{}       & 18.47 & 21.37 & $+15.7$ \\
\cp{} (Hardest OOD) & 14.15 & 17.03 & $+20.4$ \\
\bottomrule
\end{tabular}
\end{table}
The LLaVA-Video-7B projector is a two-layer MLP that maps SigLIP vision tokens into the LLM embedding space.
Each row of a linear layer's weight matrix is one output channel, so we compare the \mdi{} and \ours{} projectors via row-wise cosine similarity and relative magnitude difference across both layers.
\ours{} preserves the direction of every projector row and rescales only its magnitude.
Row-wise cosine similarity with the \mdi{} projector stays above $0.99$, while the relative magnitude difference is approximately $4\%$.
This magnitude-only shift selectively amplifies the motion-direction signal on OOD domains.
\apreftab{delta_direct_magnitude_amplification} shows that the direction concept vector magnitude at layer 21 rises by $+20.4\%$ on the most OOD \cp{}, while the source domain \ps{} stays nearly unchanged ($-2.2\%$).
This is consistent with the magnitude deficit diagnosed in \secref{magnitude_deficit}.

\newpage

\section{Case Study}
\label{appen:case_study}

\subsection{Direction-aware Motion Understanding}
\label{appen:case_direction}

We present additional qualitative examples demonstrating how direction supervision improves motion-aware video understanding under open-ended generation settings.
Unlike multiple-choice evaluation, open-ended video description requires the model to jointly capture object identity, motion dynamics, and temporal relationships without explicit answer candidates.

\begin{figure}[t]
    \centering
    \includegraphics[width=\linewidth]{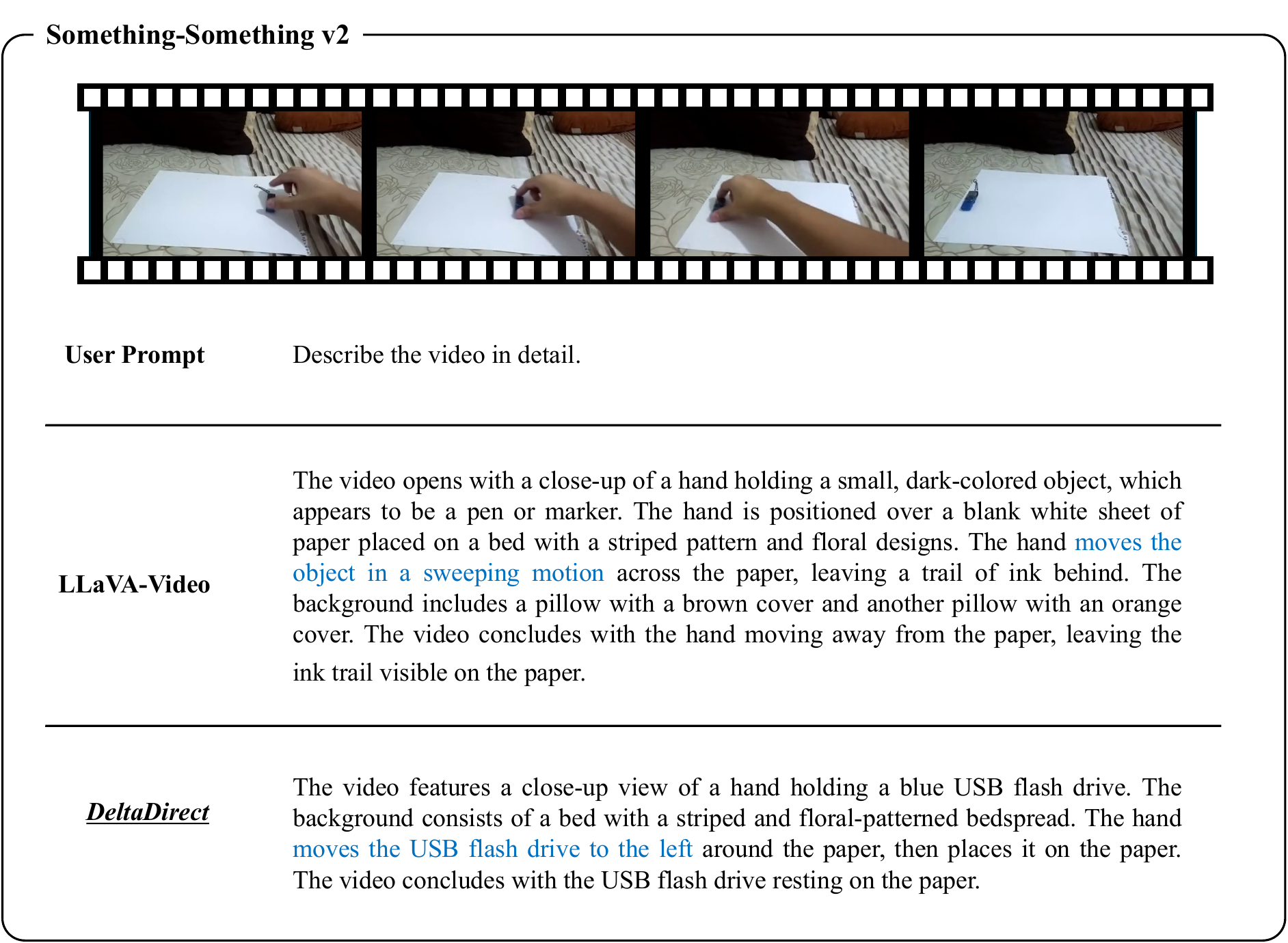}
    \caption{\textbf{Qualitative comparison on Something-Something v2 with an open-ended description prompt.} Compared to the baseline LLaVA-Video, \ours{} generates a more grounded description of both the object and its leftward motion. Blue text highlights motion- and direction-related expressions.}
    \label{fig:ssv2_case_study}
\end{figure}

Figure~\ref{fig:ssv2_case_study} shows an example from Something-Something v2.
Given the prompt ``Describe the video in detail,'' the baseline LLaVA-Video produces a partially wrong description, incorrectly interpreting the manipulated object as a ``pen'' or ``marker'' and describing a nonexistent ``ink trail'' on the paper. Although the model recognizes that motion occurs, it describes the movement only vaguely as a ``sweeping motion'' without explicitly grounding the motion direction or trajectory. In contrast, \ours{} generates a more spatially and temporally grounded description by explicitly identifying the leftward movement of the object before the final placement event on the paper.
This example suggests that direction supervision improves not only directional prediction itself, but also motion grounding in free-form video generation.

\begin{figure}[t]
    \centering
    \includegraphics[width=\linewidth]{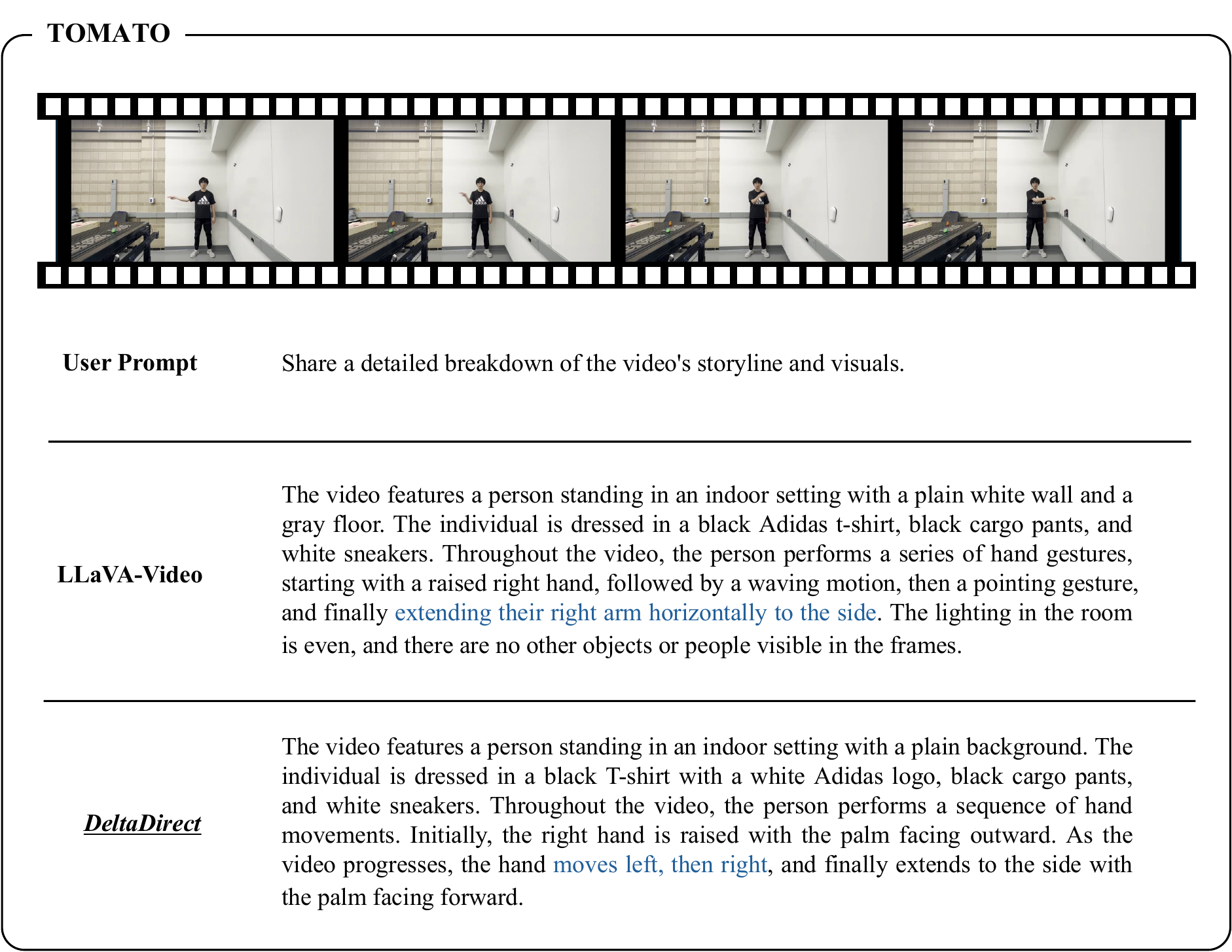}
    \caption{\textbf{Qualitative comparison on TOMATO with an open-ended description prompt.} Compared to the baseline LLaVA-Video, \ours{} provides a more precise description of the person’s hand motion, explicitly capturing its directional changes from left to right. Blue text highlights motion- and direction-related expressions.}
    \label{fig:tomato_case_study}
\end{figure}

Figure~\ref{fig:tomato_case_study} presents an additional example from TOMATO under the prompt ``Share a detailed breakdown of the video's storyline and visuals.''
While the baseline model summarizes the activity only coarsely as a sequence of generic hand gestures followed by a final arm extension, \ours{} produces a more fine-grained motion description by explicitly identifying intermediate directional transitions, including leftward and rightward hand movements before the final extension. Rather than compressing the action into a single high-level summary, the generated response decomposes the motion into sequential motion events that more closely align with the observed video dynamics.

Overall, these examples suggest that direction supervision encourages more precise motion-aware descriptions, improving the grounding of motion trajectories and directional transitions in open-ended video understanding.

\subsection{Failure Cases in Open-ended Generation}
\label{appen:failure_case}

\begin{figure}[t]
    \centering
    \includegraphics[width=\linewidth]{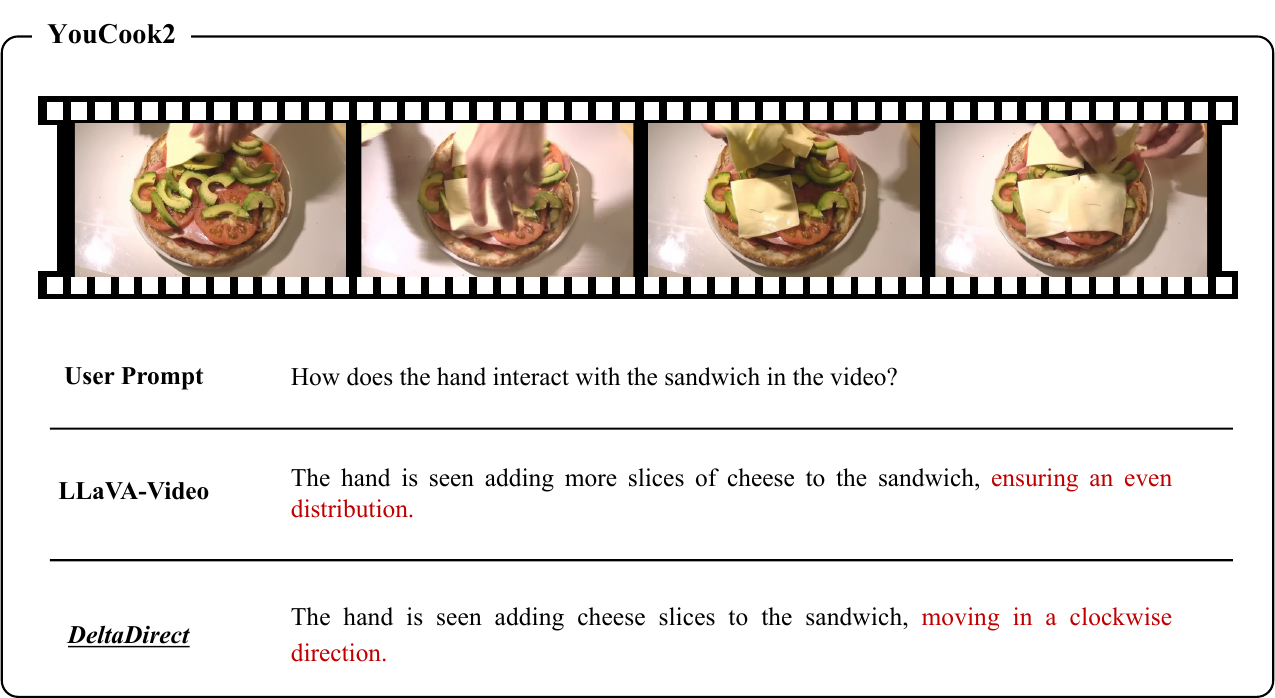}
    \caption{\textbf{Qualitative comparison on YouCook2 under an open-ended video understanding prompt.} Compared to the baseline LLaVA-Video, \ours{} tends to generate explicit directional motion descriptions, which can sometimes be unsupported by the actual video content. Red text highlights potentially incorrect expressions.}
    \label{fig:youcook2_case_study}
\end{figure}

Although direction supervision often improves motion-aware generation, it can sometimes encourage the model to produce overly specific motion descriptions even when the visual evidence is ambiguous. \figref{youcook2_case_study} shows an example from YouCook2 under the prompt ``How does the hand interact with the sandwich in the video?'' The baseline LLaVA-Video generates a simple but visually grounded response describing the placement of cheese slices on the sandwich. In contrast, \ours{} additionally introduces the phrase ``moving in a clockwise direction,'' even though the observed hand movement does not clearly exhibit a clockwise trajectory. This example suggests that explicit motion supervision may increase the tendency to generate detailed motion descriptions, occasionally leading to unsupported or inaccurate motion interpretations in open-ended video understanding settings.

\newpage

\section{Limitations}
\label{appen:limit}
This work focuses on signed 2-D image-plane motion direction, a basic but limited form of motion understanding.
Our benchmarks and supervision do not cover depth motion, rotation, acceleration, non-rigid deformation, multi-object interaction, or long-horizon event dynamics.
Moreover, \ours{} uses synthetic videos where the 2-D motion vector is analytically available; extending the same supervision to arbitrary real videos may require pseudo-labels from tracking, optical flow, or human annotation.
Our controlled synthetic benchmark assumes a single dominant moving object and a clear viewer-centric direction label, whereas real videos may contain camera motion, occlusion, or multiple objects moving in different directions.
Finally, our main analysis uses LoRA post-finetuning as the primary training regime.
We include from-scratch full fine-tuning results to verify that the benefit of \ours{} is not limited to adapter-based training, but a broader full-training study across larger backbones and training recipes remains computationally expensive and is left for future work.
\ours{} adds no inference-time computation because the MVP head is removed after training, but it incurs a small training-time overhead for the auxiliary MVP loss.
Our diagnostic analyses also require additional probe training, which is used only for analysis and not for deployment.



\section{Broader Impacts}
\label{appen:broader_impacts}
This work identifies \emph{directional motion blindness} in Video-LLMs: motion direction is linearly decodable from internal representations, yet the model fails to reliably bind it to the correct verbal response.
Our diagnosis reveals that this failure stems from a direction binding gap rather than from missing visual perception.
We show that \ours{} improves motion direction understanding on both synthetic and real-world videos, which may benefit applications that require fine-grained motion understanding, \eg robotics, autonomous navigation, and assistive technologies.
Our results suggest that improving Video-LLMs may require diagnosis-driven supervision beyond scaling models or datasets.
However, \ours{} relies on synthetic motion supervision with simplified dynamics and controlled trajectories, which may not fully capture the complexity of real-world motion.
Improved motion understanding capabilities could also be applied to surveillance or tracking systems.
We encourage future work on responsible deployment and evaluation of such technologies.
\newpage

\newpage
\clearpage


\vspace{-1.5em}
\vspace{-1.5em}

\clearpage
\newpage
\section*{NeurIPS Paper Checklist}

\begin{enumerate}

\item {\bf Claims}
    \item[] Question: Do the main claims made in the abstract and introduction accurately reflect the paper's contributions and scope?
    \item[] Answer: \answerYes{} 
    \item[] Justification: The abstract and introduction clearly and accurately reflect the paper’s core contributions and scope. Specifically, the paper claims (i) the identification of \emph{directional motion blindness} as a systematic failure mode in Video-LLMs, (ii) a diagnosis that localizes this failure to a \emph{direction binding gap}, and (iii) a solution consisting of the \md{} dataset and the \ours{} objective. These claims are consistently supported by the analysis sections and the experimental evaluation sections of the paper.

\item {\bf Limitations}
    \item[] Question: Does the paper discuss the limitations of the work performed by the authors?
    \item[] Answer: \answerYes{} 
    \item[] Justification: We have mentioned limitations in \apref{limit}

\item {\bf Theory assumptions and proofs}
    \item[] Question: For each theoretical result, does the paper provide the full set of assumptions and a complete (and correct) proof?
    \item[] Answer: \answerNA{} 
    \item[] Justification: The paper does not include any theoretical results.

    \item {\bf Experimental result reproducibility}
    \item[] Question: Does the paper fully disclose all the information needed to reproduce the main experimental results of the paper to the extent that it affects the main claims and/or conclusions of the paper (regardless of whether the code and data are provided or not)?
    \item[] Answer: \answerYes{} 
    \item[] Justification: The paper provides sufficient details to reproduce the main experimental results, including dataset construction, model architecture, training configurations, analysis procedures, and evaluation protocols. Additional implementation details are provided in \apref{imple_details}, analysis protocols are described in \apref{anal_details}, instruction-data analysis details are provided in \apref{videollms_instruction_dataset_anal}, and additional experimental setups and analyses are included in \apref{additional_exp}.

\item {\bf Open access to data and code}
    \item[] Question: Does the paper provide open access to the data and code, with sufficient instructions to faithfully reproduce the main experimental results, as described in supplemental material?
    \item[] Answer: \answerYes{} 
    \item[] Justification: An anonymized subset of the dataset, together with instructions for reproducing the main experiments, will be provided in the supplemental material during the review process. The paper and appendix additionally describe the dataset construction, generation procedures, training configurations, analysis protocols, and evaluation settings in \apref{imple_details}, \apref{datasets}. The full dataset and codebase will be publicly released with documentation upon acceptance.

\item {\bf Experimental setting/details}
    \item[] Question: Does the paper specify all the training and test details (e.g., data splits, hyperparameters, how they were chosen, type of optimizer) necessary to understand the results?
    \item[] Answer: \answerYes{} 
    \item[] Justification: The paper specifies the training and evaluation settings in sufficient detail to understand the reported results. The main text describes the model architecture, training procedures, and evaluation protocols, while detailed implementation settings, hyperparameters, optimizer configurations, and fine-tuning setups are provided in \apref{imple_details}. Dataset construction, benchmark configurations, and data splits are also clearly defined in \apref{datasets}.

\item {\bf Experiment statistical significance}
    \item[] Question: Does the paper report error bars suitably and correctly defined or other appropriate information about the statistical significance of the experiments?
    \item[] Answer: \answerNo{} 
    \item[] Justification: The paper does not report error bars or statistical significance measures such as confidence intervals or results over multiple runs. Instead, we support the main conclusions through consistent trends across multiple datasets, experimental settings, and controlled model comparisons.

\item {\bf Experiments compute resources}
    \item[] Question: For each experiment, does the paper provide sufficient information on the computer resources (type of compute workers, memory, time of execution) needed to reproduce the experiments?
    \item[] Answer: \answerYes{} 
    \item[] Justification: We provide detailed information about the compute resources used for the experiments, such as GPU types, memory, or training time in Appendix~\ref{appen:training_from_scratch} and Appendix~\ref{appen:lora_finetune_setup}.
    
\item {\bf Code of ethics}
    \item[] Question: Does the research conducted in the paper conform, in every respect, with the NeurIPS Code of Ethics \url{https://neurips.cc/public/EthicsGuidelines}?
    \item[] Answer: \answerYes{} 
    \item[] Justification: The research conducted in this paper fully conforms with the NeurIPS Code of Ethics.

\item {\bf Broader impacts}
    \item[] Question: Does the paper discuss both potential positive societal impacts and negative societal impacts of the work performed?
    \item[] Answer: \answerYes{} 
    \item[] Justification: The broader impacts are discussed in \apref{broader_impacts}
    
\item {\bf Safeguards}
    \item[] Question: Does the paper describe safeguards that have been put in place for responsible release of data or models that have a high risk for misuse (e.g., pre-trained language models, image generators, or scraped datasets)?
    \item[] Answer: \answerNA{} 
    \item[] Justification: The paper poses no such risks.

\item {\bf Licenses for existing assets}
    \item[] Question: Are the creators or original owners of assets (e.g., code, data, models), used in the paper, properly credited and are the license and terms of use explicitly mentioned and properly respected?
    \item[] Answer: \answerYes{} 
    \item[] Justification: Our experiments use third-party datasets under their original licenses or terms of use and do not relicense those data. Specifically, our synthetic data are constructed using object cutouts derived from MS COCO and scene backgrounds from Places365. For MS COCO, the annotations are released under CC BY 4.0, while the copyrights of the underlying images remain with the original image owners and associated source-image licenses.   For Places365, the official documentation states that the copyrights of all images belong to the image owners.   For the real-video benchmark, we use curated subsets from Something-Something-V2, KTH, and TOMATO. Something-Something-V2 is distributed under a research-use data license agreement,   the KTH Action Dataset is publicly available for academic research, and the TOMATO dataset card specifies a CC BY-SA 4.0 license. In training, we additionally use publicly released VideoChat2-IT annotations. Its documentation states that only the annotation files are directly distributed and that the original videos must be obtained from the provided source links. The dataset card further states that video copyrights belong to the original creators or hosting platforms and that additional source-dataset agreements may apply. 

\item {\bf New assets}
    \item[] Question: Are new assets introduced in the paper well documented and is the documentation provided alongside the assets?
    \item[] Answer: \answerYes{} 
    \item[] Justification: We introduce a new dataset (\md{}) and benchmark, and describe their construction, structure, and usage in detail in the paper and supplementary material. We will release the dataset and code with documentation, usage instructions, and license information upon acceptance.

\item {\bf Crowdsourcing and research with human subjects}
    \item[] Question: For crowdsourcing experiments and research with human subjects, does the paper include the full text of instructions given to participants and screenshots, if applicable, as well as details about compensation (if any)? 
    \item[] Answer: \answerNA{} 
    \item[] Justification: The paper does not involve crowdsourcing or research with human subjects.

\item {\bf Institutional review board (IRB) approvals or equivalent for research with human subjects}
    \item[] Question: Does the paper describe potential risks incurred by study participants, whether such risks were disclosed to the subjects, and whether Institutional Review Board (IRB) approvals (or an equivalent approval/review based on the requirements of your country or institution) were obtained?
    \item[] Answer: \answerNA{} 
    \item[] Justification: This paper does not involve research with human participants, and therefore IRB approvals or equivalent institutional review procedures are not applicable.

\item {\bf Declaration of LLM usage}
    \item[] Question: Does the paper describe the usage of LLMs if it is an important, original, or non-standard component of the core methods in this research? Note that if the LLM is used only for writing, editing, or formatting purposes and does \emph{not} impact the core methodology, scientific rigor, or originality of the research, declaration is not required.
    \item[] Answer: \answerYes{}
    \item[] Justification: The paper uses existing open-source Video-LLMs as experimental backbones for analysis and evaluation. In addition, GPT-based semantic filtering is used in the instruction-data analysis pipeline to identify direction-dependent QA pairs (\apref{videollms_instruction_dataset_anal}). The roles of these models, together with the associated analysis procedures and experimental settings, are described in the main paper and supplementary material.

\end{enumerate}
\end{document}